%% file: paper.tex
\documentclass{article}

\usepackage{microtype}
\usepackage{graphicx}
\usepackage{subcaption}
\usepackage{booktabs} 

\usepackage{hyperref}

\usepackage[preprint]{icml2026}

\usepackage{amsmath}
\usepackage{amssymb}
\usepackage{mathtools}
\usepackage{amsthm}

\usepackage[capitalize,noabbrev]{cleveref}

\theoremstyle{plain}
\newtheorem{theorem}{Theorem}[section]

\theoremstyle{definition}
\newtheorem{definition}[theorem]{Definition}

\theoremstyle{remark}

\usepackage[textsize=tiny]{todonotes}

\usepackage{tikz}

\usepackage[utf8]{inputenc} 
\usepackage[T1]{fontenc}    
\usepackage{booktabs}       
\usepackage{amsfonts}       
\usepackage{nicefrac}       
\usepackage{microtype}      
\usepackage{xcolor}         
\usepackage{multirow}
\usepackage{placeins}

\ificmlshowauthors
    \newcommand{\codeLink}{Code: \url{https://github.com/xXCoolinXx/SMIXAE}}
    \newcommand{\modelLink}{Models: \url{https://huggingface.co/xXCoolinXx/SMIXAE}}
\else
    \newcommand{\codeLink}{Code: \url{https://anonymous.4open.science/r/SMIXAE-49EB/README.md}}
    \newcommand{\modelLink}{Models: \url{https://anonymous-hf.up.railway.app/a/izi3d89jvizr/}}
\fi

\newcommand{\visualLink}{Visualization: \url{https://dainty-sawine-dc149c.netlify.app/}}

\begin{document}

\twocolumn[
\icmltitle{SMIXAE: Towards Unsupervised Manifold Discovery in Language Models}
\icmlsetsymbol{equal}{*}

\begin{icmlauthorlist}

\icmlauthor{Collin Francel}{ua}

\end{icmlauthorlist}

\icmlaffiliation{ua}{Department of Arts and Sciences, University of Alabama, Tuscaloosa, AL, United States}
\icmlcorrespondingauthor{Collin Francel}{ctfrancel@crimson.ua.edu}

  \icmlkeywords{Mechanistic Interpretability, Representation Geometry, Sparse Autoencoders, Machine Learning}

\vskip 0.3in
]

\printAffiliationsAndNotice{}

\begin{abstract}

    Sparse autoencoders (SAEs) have been used widely to decompose and interpret neural network activations, especially those of transformer language models. One key issue with SAEs is their inability to directly model multidimensional features. Instead, SAEs may tile such features by a set of independent directions that must be grouped together after the SAE training phase, impeding discoverability and interpretation of learned feature representations. We begin to address this issue by introducing the Sparse MIXture of Autoencoders (SMIXAE) architecture. Empirically, we provide evidence that SMIXAE models have success both in directly learning previously identified manifold structures, as well as finding novel structures, within the open source Gemma 2 2B and 9B models. Finally, we discuss several limitations and point towards areas for future work.
\end{abstract}

\section{Introduction}

\input{paper/newline_combined}

Understanding the representation geometry of neural networks, especially transformer language models, is an important step towards building safe and interpretable artificial intelligence. A key paradigm for understanding these representations is activation decomposition, which seeks to train an additional model on the activations of the underlying neural network to learn a more interpretable basis for the activation space. 

A major activation decomposition technique is the sparse autoencoder (SAE) \citep{bricken-sae}. SAEs are a dictionary learning technique that attempts to learn a dictionary of linear directions that correspond to features in the activation space. While SAEs have seen success for interpreting models, there remain several critical issues, such as feature splitting \citep{featuresplitting} and lack of composability of latent representations \citep{matryoshka}. Further, in the presence of multidimensional features, SAEs may learn arbitrarily many directions to tile the feature, yet do not explicitly group those directions together, impeding discoverability and automated interpretability \citep{anthropicnewline, modell, engels, scaling-manifolds, sae-manifold}.

This work introduces the Sparse MIXture of Autoencoders (SMIXAE) architecture in an attempt to better capture multidimensional features. SMIXAE is a multi-layer autoencoder partitioned into a set of sparsely-activating expert autoencoders. Compared to SAEs, we add additional encoder and decoder layers to allow for greater expressivity. We combine this design with a low bottleneck dimension to structurally ensure that additional capacity is used to model complex features. We demonstrate evidence that SMIXAE can directly model multidimensional features, finding similar structures as in \citet{modell, anthropicnewline, sinii, sae-manifold}, as well as novel features. Finally, we discuss several limitations of SMIXAE and elucidate pathways for improvement. \footnote{\codeLink, \modelLink, \visualLink}

\section{Related Work}

\paragraph{Supervised Manifold Discovery.} A common technique to extract multidimensional features from LLMs is Principal Component Analysis (PCA). PCA has been used on LLMs to find manifolds for months, weekdays, time, temperature, newline counting, and more \citep{engels, anthropicnewline, modell, sae-manifold}. However, PCA is a supervised technique, and as such it is unable to find multidimensional features at scale. 

\paragraph{Unsupervised Manifold Discovery.} One previous work that has tackled the problem of unsupervised discovery is \citet{bilinear}. The authors use a bilinear autoencoder to yield improved expressivity through the addition of quadratic terms. While they show preliminary success in unsupervised manifold discovery, their approach is fundamentally limited in expressivity. In concurrent work, \citet{sae-manifold} construct an Ising coupling-based method that finds communities of latents in trained sparse autoencoders. They show success in unsupervised discovery, but they note that their technique is fundamentally limited by the featurization of SAEs themselves. 

\paragraph{Manifold Learning.} Manifold learning is an existing and deeply related research area. The Sparse Manifold Transform, introduced in \citet{smt}, uses classical machine learning methods to model the input distribution as a single manifold. They first learn a sparse nonlinear dictionary over the dataset, then optimize a linear transformation of the dictionary into a space of lower dimension. Chart Autoencoders \citep{chartae} are another approach, which learns a set of local charts over a single manifold embedded in a high dimensional space, and then linearly projects into a lower dimension. This projection is then linearly embedded into the higher dimensional space with another series of charts. Finally, MIXAE \citep{mixae} learns a set of autoencoders, each with the goal of learning a local manifold for a particular cluster in the data, and aggregates their outputs using a mixing network. SMIXAE applies ideas from these works and adapts them to the setting where many different manifolds may be present within the same activation. 

\input{paper/table_newline}
\section{Background}

\subsection{Representation Hypotheses}
Understanding representation geometries in LLMs is crucial for model verification and alignment with human values. In this work, we seek to study such representations and discover them in an unsupervised fashion. Concretely, we study activations in several layers in LLMs and attempt to model the representation geometries present in a given activation. Throughout this work, we refer to such geometries present as (possibly multidimensional) \textbf{features}, and refer to representations of those features found by activation decomposition models as \textbf{latents}, \textbf{latent representations} or \textbf{latent feature representations}. We also occasionally refer to multidimensional features as \textbf{manifolds}.

In order to model features in LLMs, it is necessary to first develop a hypothesis about how the model represents such features. We base our investigation around the multidimensional Linear Representation Hypothesis (mLRH), defined below. mLRH models activations as a sum of features, where features are structured subspaces that the model uses to represent some aspect of the input. Each feature has an associated function, parameterized by the input $x$, which may be nonlinear.

\begin{definition}[Multidimensional Linear Representation Hypothesis]

There exists a collection of features labeled $f \in F$, with dimensions $d_f \ll n$, and associated embedding matrices $E_f \in \mathbb{R}^{n \times d_f}$ ($\text{Rank}(E_f) = d_f$), such that the functional relationship between an input $x \in \mathcal{X}$ and its representation $\Psi(x) \in \mathbb{R}^n$ is 
\begin{equation}
\Psi(x) = \sum_{f \in F} \mathbb{I}_{f}(x) E_f v_f(x) \label{mlrh},
\end{equation}
where $v_f(x) : \mathcal{X} \mapsto  \mathbb{R}^{d_f}$ is a function, and $\mathbb{I}_f(x)$ is the indicator function denoting whether feature $f$ is present in $x$.

Further, assuming the superposition hypothesis \citep{toymodels}, we have that the sum of feature dimensions is far greater than $n$ and that there are sparsely many features active for a given input $x$, i.e.
\begin{equation}
    \sum_{f\in F} d_f \mathbb{I}_f(x) \ll n \ll \sum_{f\in F}d_f \label{superposition}.
\end{equation}

\end{definition}

The statement above is inspired by \citet{modell}, with changes to improve generality. In transformer models, $\Psi(x)$ may be taken to be the residual stream activation at a given layer for the input prompt $x$. 

Notably, mLRH is a strong relaxation of the standard Linear Representation Hypothesis \citep{park-lrh}, which forces $v_f(x)$ to be one dimensional and constrains $||v_f(x)||_2$ to denote how 'present' the feature $f$ is in $x$. Standard LRH has directly inspired the design of SAEs for interpreting LLMs, as SAEs model activations with linear directions, but lacks the expressivity of mLRH and does not hold in the presence of more complex representation structures.

We consider $v_f(x)$ to denote the structure of the semantic information in $f$. For example, $v_f(x)$ can span a circular or helical manifold, such as those found in \citet{nanda} and \citet{anthropicnewline}, to represent addition and counting operations, respectively. Another possible structure is the fractal, as demonstrated in \citet{shai}, to represent belief state about a hidden Markov model.

\subsection{Sparse Autoencoders}

A common architecture for learning these representations is the Sparse Autoencoder. SAEs are defined as follows. Let $n$ be the dimension of the input, and let $w$ be the width of the SAE, i.e. how many latent variables the SAE has. Let $W_e \in \mathbb{R}^{w\times n}$ be the encoder weight matrix, and $b_e \in \mathbb{R}^w$ be the encoder bias. Then the encoding step is represented by $f_e(x) = \sigma(W_e x + b_e)$ where $\sigma$ is some sparsity inducing activation, such as $\text{JumpReLU}$ \citep{google-jumprelu} or $\text{TopK}$ \citep{topksae}. The goal of SAEs is for the latent space spanned by $f_e(x)$ to represent a set of sparse, interpretable features, corresponding to linear directions (as predicted by the standard LRH). Let $W_d \in \mathbb{R}^{n \times w}$ and $b_d \in \mathbb{R}^{n}$. Then the decoding step is $\hat{x} = W_d f_e(x) + b_d  $. To train the SAE, a standard $\text{MSE}$ reconstruction loss is applied: $\mathcal{L} (x,\hat{x})= \text{MSE}(x, \hat{x})$.

Sparse autoencoders have shown remarkable ability to recover representations in model activations \citep{bricken-sae}, even in the presence of nonlinear manifolds \citep{modell, anthropicnewline, notallmen}. However, in the latter case, SAEs are thought to present a difficult methodological challenge through feature splitting along manifolds \citep{modell, scaling-manifolds, sae-manifold}. Instead of learning multidimensional geometries directly, SAEs may instead tile manifolds using many different directions, making it difficult to group together all latent variables for a particular concept. Ideally, we would like an architecture that can directly learn such manifolds in an unsupervised setting, so that they can be more readily analyzed and interpreted at scale.

\input{paper/probe_gemma_2_9b_l11}

\section{Methods}
\subsection{SMIXAE}

In this work, we introduce the Sparse MIXture of Autoencoders (SMIXAE) architecture. Conceptually, we move towards flexibly modeling both independent, orthogonal directions and more complex multidimensional structures. SMIXAE accomplishes this by using a four layer architecture, which is partitioned into a set of individual autoencoders. Each autoencoder has a small bottleneck dimension - in this work, fixed at $3$ dimensions - such that the extra parameter capacity must either be left unused or devoted towards modeling nonlinear structures. We refer to individual autoencoders in the full SMIXAE architecture as experts.

\subsubsection{Architecture}
SMIXAE is mathematically described as follows: Suppose $p$ is the size of the first latent dimension, and $b$ is the size of the second latent dimension (we refer to this as the bottleneck, with $b=3$ throughout this work). Additionally, suppose $j$ is the number of expert autoencoders. Let $W_e \in\mathbb{R}^{p \times j \times n}$, $W_b \in\mathbb{R}^{b\times j \times p}$, $b_e \in \mathbb{R}^p$. Then the encoding step is modeled by
\begin{equation}
f_e(x) = \sigma(W_b(\text{ReLU}(W_e x + b_e) )) \label{smixae_enc},
\end{equation}

where $\sigma$ is a sparsity inducing group activation function. In this work, we use the $\text{BatchTopK}$ function over the $L^2$ norms of expert activations \citep{batchtopk}, i.e. $\sigma(x) = \mathbb{I} (\text{BatchTopK}(||x||_2) > 0) \cdot x$. We also track the exponential moving average of the minimum admitted activation as the threshold $t$. At inference time, the sparsity function then becomes $\sigma(x) = \mathbb{I}(||x||_2 > t) \cdot x$. Additionally, we use $\text{LeakyReLU}$ with a small negative slope ($\alpha=10^{-4})$ in place of $\text{ReLU}$ to avoid implementing multilevel dead neuron tracking.  

We follow the encoder with a linear decoder. Let $W_{ub} \in \mathbb{R}^{p \times j \times b}$ and let $W_d \in \mathbb{R}^{p \times n}$ . Let $b_d \in\mathbb{R}^n$. Then the decoder is written as
\begin{equation}
\hat{x} = W_d(W_{ub}f(x)) + b_d.\label{smixae_dec}
\end{equation}

Our loss objective is as follows
\[
\mathcal{L}(x, \hat{x}) = \text{MSE}(x, \hat{x}) + \lambda \mathcal{L}_\text{aux} (f_e(x)) .
\]
Where $\mathcal{L}_{\text{aux}}$ is an auxiliary loss term that pushes dead experts to grow in norm. We find empirically that the most commonly used auxiliary loss function for \texttt{TopK} and \texttt{BatchTopK} SAEs \citep{topksae}, which applies a gradient to dead latents for explaining the residual error, induces feature splitting \citep{featuresplitting} in SMIXAE. As such, we formulate the following auxiliary loss:
\[
\mathcal{L}_{aux}(f_e(x)) = \sum_i \text{ReLU} (t-||f_{e_i}(x)||_2) ||[W_{ub}W_d]_i ||_F,
\]
where the gradients of the decoder norm are detached to avoid the degenerate solution where the decoder matrices approach the $0$ matrix. This loss is styled after the JumpReLU pre-activation loss from \citet{anthropic-jumprelu} but applied to BatchTopK. We find that this loss effectively avoids dead experts in SMIXAE without inducing observable feature splitting. 

We also rescale encoder activations by the Frobenius norm of the decoder before $\text{BatchTopK}$, so that our activations become $f_i(x) ||[W_{ub} W_d]_i||_F$, and do not downscale back to the original activation after $\text{BatchTopK}$ is applied. We find that this architectural choice provides stability for the encoder activation norm, avoiding continuous growth during training and leading to higher quality final models.

We visualize the performance of a single autoencoder in SMIXAE when modeling a torus or a helix, randomly embedded in $\mathbb{R}^{100}$, in Figure~\ref{sad_toymodel} (Appendix~\ref{app:sasm}), demonstrating that each expert is capable of learning a particular geometry when provided sufficient training data in a noise free setting and illustrating that this autoencoder design is a useful building block for SMIXAE.

\input{paper/table_core_eval}

\subsubsection{Motivation}

In this section, we seek to make explicit the connection between our statement of the mLRH and SMIXAE. In particular, we wish to demonstrate that approximations to the mLRH exist in the hypothesis space of SMIXAE. Of course, whether SMIXAE learns the mLRH decomposition in practice depends on the veracity of mLRH and real-world training dynamics.  

Suppose we are examining the activations of a model at a particular layer from input $x \in \mathcal{X}$. Let the activation be denoted by $\Psi(x)$ and assume that $\Psi(x) \in \mathbb{R}^n$ is structured as in \eqref{mlrh}. Using an autoencoder architecture, we then seek to recover the set of functions $v_f(x)$ and their presence $\mathbb{I}_f(x)$, for these are the individual, interpretable units of the representation. Finally, we seek to reconstruct the representation $\Psi(x)$ in order to provide optimization pressure on the learned latent representations.  

By assumption, $\Psi(x)$ is a sum of all feature representations $\Psi_f(x)$. As such, we may effectively partition our autoencoder into a set of smaller autoencoders, each with the goal of modeling a single $\Psi_f(x)$. Let the number of autoencoders, $j$, be greater than $|F|$. 

The first objective for each autoencoder is to learn an approximation to $v_f(x)$. Assume that there exists some function $g_f : \mathbb{R}^n \to \mathbb{R}^{b}$, $b \geq \max_{f \in F} d_f$ such that $g_f(\Psi(x)) \approx \iota_f v_f(x)$, where $\iota_f \in \mathbb{R}^{b \times d_f}$ is an isometry.\footnote{This assumption is reasonable when assuming the superposition hypothesis, though we do not formally show it here.} We thus seek to approximate the function $g_f(x)$. Indeed, by a classical result due to \citet{hornik}, we may approximate $g_f(x)$ arbitrarily well (dependent on width $p$) with a 2-layer ReLU network, provided $g_f(x) \in L^q (\mathbb{R}^n)$ for some fixed $q \geq 1$. As such, we construct the encoder of each autoencoder of SMIXAE to be a 2-layer ReLU network in \eqref{smixae_enc}. Denote this approximation by $\hat{g}_f(x)$. We thus obtain an approximation for $v_f(x)$ by $\hat{v}_f(x) = \hat{g}_f(\Psi(x))$.

Next, we need some way to approximate $\mathbb{I}_f(x)$. A reasonable assumption is that there exists a threshold $t \in \mathbb{R}^+$ such that if the Euclidean norm $||\hat{v}_f(x)||_2$ is above the threshold, the feature is present in the representation $\Psi(x)$. This approximation acts as a noise filter, as we assume that lower activation norm implies that the activation of $\hat{v}_f(x)$ is more likely to be noise. As such, let $\hat{\mathbb{I}}_f(x) = \mathbb{I}(||\hat{v}_f(x)||_2 > t)$. \footnote{During training, we are approximating $t$ with $\text{BatchTopK}$.}

Combined, we obtain our latent representation of $\Psi_f(x)$ as $\hat{\mathbb{I}}_f(x) \hat{v}_f(x)$. Finally, we simply need to learn a linear projection of this latent representation into $\mathbb{R}^n$, and sum up all such representations to obtain the reconstructed representation $\hat{\Psi}(x)$. In SMIXAE, we accomplish this with our linear decoder, as in \eqref{smixae_dec}, which implicitly sums together the outputs of each autoencoder. This concludes our motivation of SMIXAE.

\subsection{Training}

We implement SMIXAE within \texttt{SAELens} \citep{saelens} and make extensive use of the library's training infrastructure to train SMIXAE. We exclusively load and evaluate models with \texttt{transformers} \citep{transformers}.

We train SMIXAE on activations from Gemma 2 2B, on layer 12, and Gemma 2 9B \citep{gemma2}, on layers 11 and 20. We train on layers 12 of Gemma 2 2B and 20 of Gemma 2 9B for they are common targets for SAE training \citep{neuronpedia}. We choose layer 11 of Gemma 2 9B to ascertain whether our architecture can learn the newline counting manifold identified by \citet{sinii}. All training runs are performed on a single NVIDIA A100 GPU. Hyperparameter configurations are detailed in Table \ref{tab:hpo} (Appendix \ref{sec:hpo}).

\subsection{Probing}

In order to examine SMIXAE outputs on downstream tasks which may utilize manifold representations, we construct several labeled datasets. In particular, we construct datasets for examining the following properties: hours, weekdays, months, time units, temperatures, colors, emotions, and living things. The first six datasets are chosen due to having relational structure that would be well modeled by manifolds, and since such representations have been found in LLMs by \citet{modell, engels}, among others. We choose living things to determine whether the SMIXAE can also represent hierarchical and tree-like structures, as previously found in \citet{hierarchy}, though modeling these structures is not the primary goal of SMIXAE. 

For each dataset, we construct several hypotheses about ways the model may represent the data. For example, for hours we hypothesize that the model may represent the 24 hours in a day using a 24 hour ring structure, two ring structures for AM/PM, or by clustering AM and PM times separately. 

For all ring hypotheses, including hours, we transform the labelled data into $(\cos(k/n \cdot 2\pi), \sin(k/n \cdot 2\pi)$, where $k$ is the position in the number of classes (e.g. $k=13$ for time 1PM) and $n$ (e.g. $n=24$ for the 24 hour ring) is the number of classes. 

Another transformation we apply is $\log(1+x)$ or $\log_{10}$ for temperature and time units tasks, respectively. We apply $\log$ transformations as it is thought that humans perceive such numerical quantities logarithmically rather than linearly \citep{log_numbers}, and as such it is interesting to determine if LLMs represent these same quantities in kind. 

We discover experts that are relevant for the given task by first using multivariate Fisher score as a heuristic filter (in cases where the data is continuous/high label count, we first bucket the data and then use the same filter step) to obtain the top $50$ experts. We then perform regression using $5$-fold cross validation for each expert on a given task, using one of linear, multinomial, or logistic regression, depending on the task. 

The full list of hypotheses examined and details on the regressions used for each dataset are found in Table \ref{tab:probing}. For more information about the datasets used, please see Appendix \ref{app:datasets}.

\subsubsection{Newline Counting Manifold}
For probing for the newline counting manifold, we directly adapt the approach and layers examined from \citet{sinii}. We find SMIXAE activations on \texttt{monology/pile-uncopyrighted} \citep{monology_pile, thepile}, and format samples such that each line is $80$ or $150$ characters long, followed by a newline. We rank experts by $\Delta R^2$ between periodic and linear regression models applied to each expert, with labels being the number of characters since the previous newline, to find experts which most exhibit the expected curvature. 

\subsubsection{Random Sampling}
To evaluate SMIXAE for fully unsupervised discovery, we collect samples of at most $1,000$ points of SMIXAE expert activations on a sample of $10,000$ prompts from \texttt{monology/pile-uncopyrighted} \citep{monology_pile, thepile}. We then take a random sample of $10$ experts that activated at least $100$ times in the sample. We manually examine each expert to attain a qualitative perspective about its structure and to assess whether the expert fires on a particular concept. We then manually label experts with the concept that they appear to fire on, or label the expert as noise if it doesn't appear to fire on anything in particular. 

\subsubsection{Core Evaluation}

Finally, we re-implement the core evaluation suite from SAEBench \citep{saebench} to exclusively use the \texttt{transformers} library, to match the rest of our training and evaluation setup. We evaluate $L_0$, fraction alive, explained variance, mean squared error, and cosine similarity on a sample of approximately $1,000,000$ tokens, and evaluate the cross entropy score on a sample of approximately $400,000$ tokens.
\input{paper/table_probing}

\section{Results}

\paragraph{Probing. } On the probing results, detailed in Table \ref{tab:probing}, we find high regression scores across all tasks. This provides preliminary evidence that SMIXAE captures the probed concepts within single experts. These results do not, however, certainly imply that the learned structures are causally relevant, nor do they guarantee that the learned structures of top experts correspond to the expected structure. 

As such, we performed a qualitative assessment of the top experts for each probing task. While we frequently find experts that do have alignment with the expected structure, they are not always the highest ranked expert. For example, for the 12-Month Ring hypothesis on Gemma 2 9B, layer 11, we find that the top expert has a higher correlation with the periodic labels but no apparent structure, while the 2nd ranked expert has a defined ring structure.

Another issue we saw with regression scores was for logistic and multinomial probing. Many experts, possibly by chance, learn discrete clusters in the bottleneck space that do not have an apparent structure or ordering. This was the case on the Living Things task, as the probes were able to achieve very high accuracy, but were, in fact, simply exploiting the discrete clusters structure to find a separating plane, and there was no evidence of an orthogonal clustering between plants and animals, as previously observed in \citet{hierarchy}, or of a tree like structure for taxons. 

Because of these issues, we reviewed the regression scores, and then visually examined expert activation graphs to determine whether their score results from structure or from noise. We perform this process over the top 10 experts for each hypothesis of each task, and we plot the experts with the highest visual alignment to the expected structure in Figures \ref{fig:probe_gemma_2_9b_l11},  \ref{fig:probe_gemma_2_2b_l12}, and \ref{fig:probe_gemma_2_9b_l20}.\footnote{Visualizations for all of the top 10 SMIXAE experts are available on \url{https://dainty-sawine-dc149c.netlify.app/}}

We find that Gemma 2 9B, Layer 11, has the greatest number of structurally aligned experts. In particular, for 6 out of 7 tasks, we were able to find experts with both high structural alignment to the label as well as high regression score. Gemma 2 9B, layer 20 had less such experts, having so for only 4/7 tasks. Finally, Gemma 2 2B had the least such structured experts, having so for only 3/7 tasks. 

One particularly noteworthy expert is Expert 485 in the Gemma 2 9B, Layer 11 SMIXAE (Figure \ref{fig:probe_gemma_2_9b_l11}, (f)), on the temperature task. We expected to see a linear or curvilinear structure, and did find in Expert 892. However, Expert 485 appears to take such a linear structure and pinch the middle range of temperatures, such that more moderate temperatures are spatially co-located, and more extreme temperatures are spatially co-located. We speculate that the model uses this representation to make sense of extreme versus mild temperatures, though we caution that this suggestion needs to be confirmed through causal analysis.

\paragraph{Newline Counting Manifold. } On Gemma 2 9B, we search for the newline counting manifold, following the methodology of \citet{sinii}. In Table \ref{tab:newline}, we report $\Delta R^2_{\text{per}}$ for the top expert and the mean for the top $5$ experts, and provide further breakdown per expert in Table \ref{tab:newline_appendix_gemma_2_9b_l11}. We find that the top $2$ experts have values near $0.5$, followed by a sharp fall off in value. Given this gap, we plot only the top two experts on Layer 11 for comparison against \citet{sinii}. We note that this implies the newline counting manifold has been split over multiple experts.

In Figure \ref{fig:newline_gemma_2_9b_l11}, we find high visual alignment with the structure discovered by SMIXAE and that found by \citet{sinii} through PCA analysis. We note that PC4-6 in their figure contains significantly more coiling than either of our experts. We are unsure why this has occurred, but speculate it could be due to our SMIXAE models possibly being undertrained, or due to the encoder of SMIXAE lacking sufficient expressivity to model such a tightly coiled helix. 

\paragraph{Random Sampling. } We evaluate our random sample of $10$ experts for each trained SMIXAE model. Since our sample size is quite small, it is difficult to draw broad conclusions about these results. However, it provides some preliminary information about what geometries are commonly learned, and whether most SMIXAE experts are interpretable. In Table \ref{tab:random_sample}, we find that most experts in each sample are interpretable,  with only $0-3$ per sample either being uninterpretable to the human grader or purely noise. We additionally find that a sizable number of experts in Gemma 2 9B are approximately linear, hinting that, while standard LRH doesn't hold in general, it describes a reasonable fraction of representation geometries. Interestingly, we find that only one expert in Gemma 2 2B appears to consist of entirely linear features. We suspect this is due to sample noise. 

We further find that lower-rank features often share an expert. For example, three approximately orthogonal linear directions often share a single expert. This complicates interpretability, as it cannot be assumed that each expert of SMIXAE captures a single manifold. See Section 6 for further discussion.

We provide visualizations for each random sample in Figures \ref{fig:random_gemma_2_2b_l12}, \ref{fig:random_gemma_2_9b_l11} and \ref{fig:random_gemma_2_9b_l20} (Appendix \ref{appendix:random_sample}). 
\paragraph{Core Evaluation.} We examine the core evaluation results for SMIXAE, with Gemma Scope SAEs as a reference baseline. We find that SMIXAE attains reasonable numbers on all such metrics, indicating that all SMIXAE models were successfully trained. We also see that the fraction of alive latents in SMIXAE is quite high, demonstrating that our newly introduced auxiliary loss was beneficial in preventing dead experts. 

We note that, despite significantly increasing the parameter count, SMIXAE does not exceed the performance of Gemma Scope on reconstruction metrics. This indicates that reconstruction performance is determined by the width of the autoencoder, rather than by raw parameter count, and verifies our bottleneck design choice. As found in the above results, SMIXAE successfully devotes this additional parameter capacity to nonlinear structures in the bottleneck space, rather than over-optimizing for reconstruction. Alternatively, a densely connected autoencoder design would be expected to over-optimize reconstruction and be comparable with Gemma Scope despite having more limited representational capacity. 

\begin{table}
\begin{tabular}{c c|c c c}
    Model & Layer & Noise & Linear & Nonlinear \\
    \midrule
    Gemma 2 2B & $12$ & $3/10$ & $1/10$ & $6/10$ \\
    Gemma 2 9B & $11$ & $2/10$ & $6/10$ & $2/10$\\
    Gemma 2 9B & $20$ & $0/10$ & $4/10$ & $6/10$ \\
\end{tabular}

\caption{Per model+layer counts of experts in the random sample that (1) didn't appear to represent any particular concept/were entirely noise and (2) that was made up of approximately linear, orthogonal directions.  }
\label{tab:random_sample}
\end{table}

\section{Limitations and Future Directions}
While this work introduces an interesting new paradigm for interpretability, several limitations still remain. 

\paragraph{Toy Models with Multidimensional Features.} We found that SMIXAE failed on toy models with multidimensional features that we tried, despite promising results in LLMs on both supervised probing and unsupervised discovery. As such, we suspect that the simplest construction, which isometrically embeds origin-centered manifolds into a higher dimensional space and sums them together (see \citet{sae-manifold}, Appendix E for more details), fails to meaningfully capture how multidimensional features are embedded in LLMs. In particular, we believe that including affine shifts and clustering of multidimensional structure may be necessary to construct a useful toy model. We invite ideas for new toy model designs that more closely match the topology of multidimensional features in LLM activations.

\paragraph{Learnability of Bottleneck Dimension.} In this work, we fix $d_{\text{bottleneck}}=3$ for simplicity of visualization, and because we expect that most features in language models can be represented in fewer than $3$ dimensions. However, we demonstrate both that higher dimensional features may be split among multiple experts and that $1$-dimensional linear features may share an expert, complicating interpretation of expert activations. Ideally, we would like to learn the correct $d_{\text{bottleneck}}$ per expert. Future work may explore a heterogenous approach between SAEs and SMIXAEs, or learning $d_{\text{bottleneck}}$ during training.

\paragraph{Causal Verification.} While we demonstrate that SMIXAE experts learn representations that appear to match the intended structure for a given task, we do not yet provide claims as to whether these representations are causal. Future work will explore activation patching and causal steering to address this gap. See \citet{manifold-steering} for concurrent analysis in this area. 

\paragraph{Complexity-Interpretability Tradeoff.} SMIXAE adds significant additional complexity, compared to SAEs, by adding additional layers and moving beyond modeling activations with linear directions. Future work should explore ways to incorporate interpretable nonlinear functions, such as splines, in place of the nonlinearities in the first encoder layer. 

\section{Conclusion}

This work introduces the Sparse Mixture of Autoencoders architecture (SMIXAE) as an alternative method to SAEs for interpreting LLM activations. We demonstrate promising results in multidimensional structure recovery, but also find key challenges that obstruct its use for interpretability in its current form. We hope that SMIXAE inspires other beyond-SAE architectures and further reassessment of the hypotheses that underpin our understanding of representation geometries.

\FloatBarrier
\section*{Impact Statement}

This paper presents work whose goal is to advance the field of Machine
Learning. There are many potential societal consequences of our work, none
which we feel must be specifically highlighted here.

\bibliographystyle{icml2026}
\bibliography{refs}



\appendix

\onecolumn

\section{Training Hyperparameters}\label{sec:hpo}

We provide all hyperparameters used in Table \ref{tab:hpo}. We hold all hyperparameters constant regardless of model or layer. The only difference is input dimension, which is $2304$ for Gemma 2 2B and $3584$ for Gemma 2 9B.

\begin{table*}

\centering

\begin{tabular}{c|c}
     Hyperparameter & Value  \\
     Training Tokens & $500,000,000$ \\
     Number of Experts & $2048$ \\
     Dimension of Expert & $16$ \\
     K Experts & $64$ \\
     Bottleneck Dimension & $3$ \\
     Auxiliary Loss coefficient & $9 \cdot 10^{-6}$ \\
     Threshold Learning Rate & $1 \cdot 10^{-1}$ \\
     Decoder Initialization Norm & $0.1$ \\
     Learning Rate & $5\cdot 10^{-4}$ \\
     Warmup Steps & $500$ \\
     Decay \% Training & $20\%$ \\
     Scheduler & WSD \\
     Adam $\beta_1$ & 0.9 \\
     Adam $\beta_2$ & 0.999 \\
     Training \texttt{dtype} & \texttt{bfloat16} \\ 
     Context Size (LLM) & 128 \\
     Batch Size & 8192
     
\end{tabular}

\caption{We keep hyperparameters, except for input dimension, fixed for all models. }
\label{tab:hpo}
\end{table*}

\section{Full Table results} 

In Tables \ref{tab:probing_appendix_gemma_2_2b_l12}, \ref{tab:probing_appendix_gemma_2_9b_l11}, and \ref{tab:probing_appendix_gemma_2_9b_l20} we provide the complete probing scores for the top $10$ SMIXAE experts on each task. 

In Tables \ref{tab:newline_appendix_gemma_2_9b_l11} and \ref{tab:newline_appendix_gemma_2_9b_l20}, we provide the full $\Delta R^2_\text{per}$ scores for the top 10 experts on the newline counting task. 

\input{paper/table_probing_appendix}
\input{paper/table_newline_appendix}

\section{Visualizations for Other Models and Layers}

In Figure \ref{fig:probe_gemma_2_2b_l12}, we provide visualizations on the probing task for Gemma 2 2B, layer 12. 

In Figure \ref{fig:probe_gemma_2_9b_l20}, we provide visualizations on the probing tasks for Gemma 2 9B, layer 20.

\input{paper/probe_gemma_2_2b_l12}
\input{paper/probe_gemma_2_9b_l20}

\section{Random Sample of Expert Activations} \label{appendix:random_sample}

In order to provide a preliminary account of true unsupervised discovery of features with SMIXAE, we collect samples of at most $1$ thousand points of SMIXAE expert activations on a sample of $10$ thousand prompts from \texttt{monology/pile-uncopyrighted} \citep{monology_pile, thepile}. We then take a random sample of $10$ experts that activated at least $100$ times across all tokens, and plot the results in the following figures. We additionally include manually determined labels, which reflect the human grader's inference about the meaning of the activations. These results may be found in Figures \ref{fig:random_gemma_2_2b_l12}, \ref{fig:random_gemma_2_9b_l11} and \ref{fig:random_gemma_2_9b_l20}. 

Occasionally, certain structures may be marked as "Noise" when there is no apparent structure, or suffixed with a question mark to reflect greater uncertainty about the label compared to higher confidence labels. 

\input{paper/random_gemma_2_2b_l12}
\input{paper/random_gemma_2_9b_l11}
\input{paper/random_gemma_2_9b_l20}

\section{Additional Newline Manifold Visualizations
}

We mainly examine the newline counting manifold in Gemma 2 9B, layer 11. However, we also produce results for Gemma 2 9B, layer 20 (Figure \ref{fig:newline_gemma_2_9b_l20}) and for Gemma 2 2B, layer 12 (Figure \ref{fig:newline_gemma_2_2b_l12}).

\input{paper/newline_gemma_2_2b_l12}
\input{paper/newline_gemma_2_9b_l20}

\section{Probing Datasets} \label{app:datasets}

In order to elicit the desired concepts for each probing task, we construct datasets of sentences that reference the desired concept. Specifically, we create a set of templates for each task, parameterized by a randomly chosen person name (to provide variance) and the target concept. 

Prior work \citep{poobing} finds that SAEs are not performant at aggregating representations over the entire input prompt. This failure is due to a lack of temporal consistency enforced during SAE training, as examined in works like \citet{priorsintime}. As such, it is necessary either for post-hoc aggregation, such as mean or max pooling, to be applied, or for the target concept to be expressed in the same location in the prompt as where the probe is trained. SMIXAE shares the same training environment as standard SAEs, and so inherits the same issue with single-token probing. However, pooling is also problematic for SMIXAE, as neither mean nor max pooling necessarily respects the manifold geometry.

Therefore, to avoid the issues of both single token probing and pooling, we fix the location of the target concept at the end of the prompt, train the probe on the last token of the prompt, and add variability to probing through having a set of $30$ templates available for each task with additional variability added by choice of the name of the human actor in the sentence. Examples of these templates are found in Table \ref{tab:templates}. 

We randomly generate $1,000$ samples for each task using these templates.

\begin{table}[h]
  \centering
  \begin{tabular}{ll}
  \hline
  \textbf{Task} & \textbf{Example Template} \\
  \hline
  Weekdays & ``\{name\} will finalize the PCR results on \{day\}'' \\
  Hours & ``\{name\} set the alarm for \{hour\}'' \\
  Temperatures & ``\{name\} noted the sample chamber registered \{temp\}°F'' \\
  Time Units & ``\{name\} set the experiment timer for one \{unit\}'' \\
  Living Things & ``The researcher identified the specimen as a \{organism\}'' \\
  Colors & ``\{name\} labeled the sample vial with tape coded \{color\}'' \\
  Months & ``\{name\} will submit the annual report in \{month\}'' \\
  \hline
  \end{tabular}
  \caption{Dataset generation templates for probing tasks}
  \label{tab:templates}
\end{table}

\section{Single Autoencoder on Single Manifold} \label{app:sasm}

\begin{figure}
\centering
\includegraphics[]{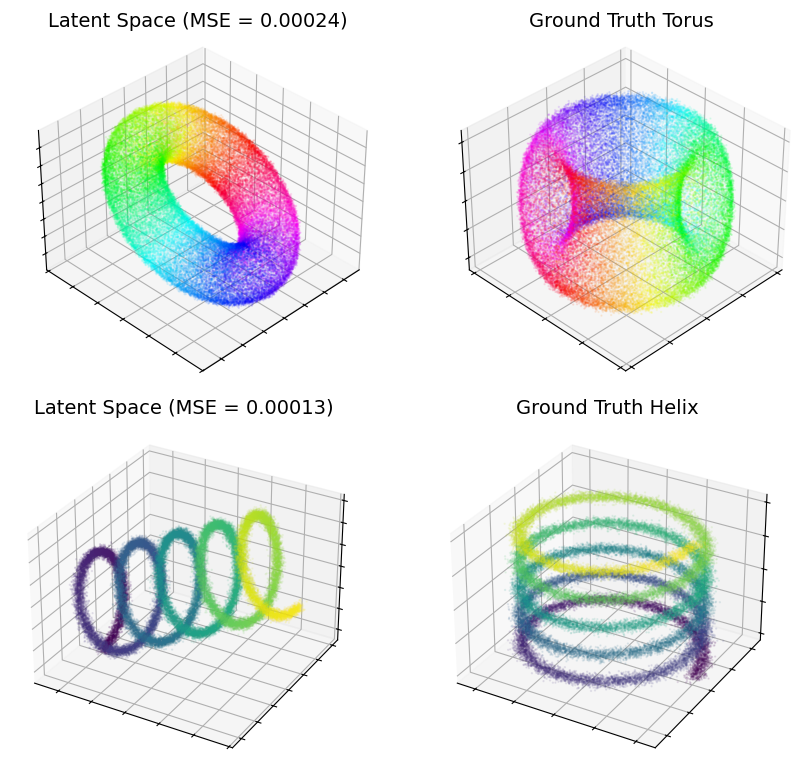}

\caption{Visualization of a single autoencoder trained on a torus or helix. We observe low MSE loss, showcasing that each autoencoder in SMIXAE is capable of learning manifolds when provided sufficient training data in a noise free setting. }
\label{sad_toymodel}
\end{figure}

In Figure \ref{sad_toymodel}, we visualize a single autoencoder of SMIXAE on a single manifold toy model. We first generate a set of noisy points on each manifold, and then embed them into a $100$ dimensional space using a randomly generated linear embedding. We train a single autoencoder on each set of points, and plot both the latent space of the autoencoder and the ground truth manifold. We color the torii by the toroidal angle, and color the helix by the parameter of the helix. We observe low MSE loss, showcasing that each autoencoder in SMIXAE is capable of learning manifolds when provided sufficient training data. 

\end{document}

%% file: paper/newline_combined.tex
\begin{figure}[t]
\centering

\noindent\makebox[\linewidth][c]{%
\begin{minipage}[c]{0.43\linewidth}%
\centering%
\begin{tikzpicture}[baseline=(img.base)]
  \node[inner sep=0pt] (img) {\includegraphics[width=\linewidth,keepaspectratio]{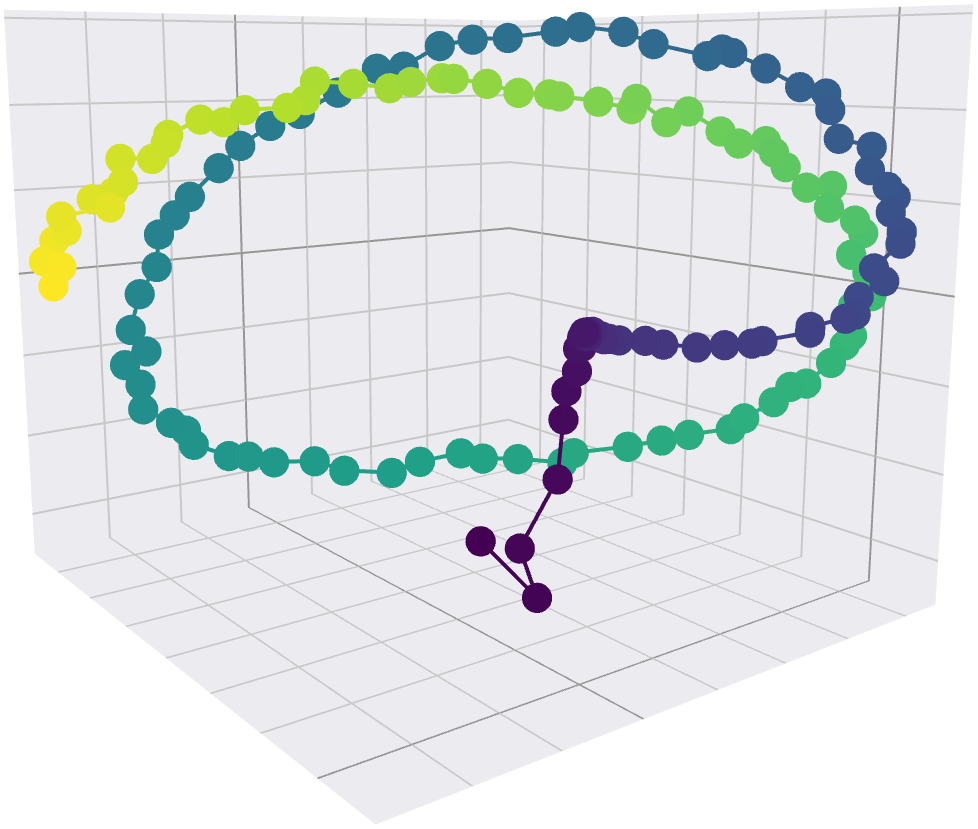}};
  \node[anchor=north west, inner sep=2pt, overlay] at (img.north west) {\small\textbf{(a)}};
\end{tikzpicture}%
\end{minipage}%
\hfill%
\begin{minipage}[c]{0.43\linewidth}%
\centering%
\begin{tikzpicture}[baseline=(img.base)]
  \node[inner sep=0pt] (img) {\includegraphics[width=\linewidth,height=4.0cm,keepaspectratio]{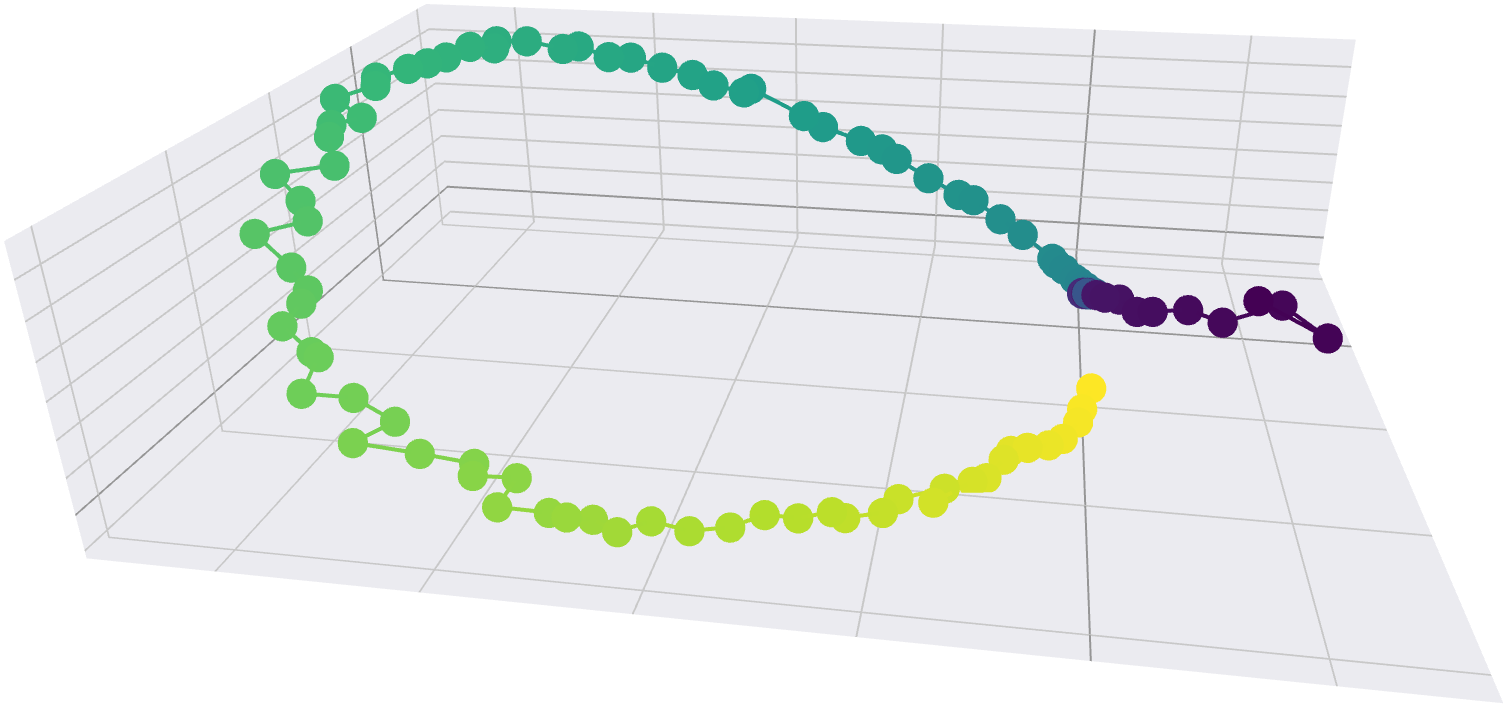}};
  \node[anchor=north west, inner sep=2pt, overlay] at (img.north west) {\small\textbf{(b)}};
\end{tikzpicture}%
\end{minipage}%
\hfill%
\begin{minipage}[c]{0.12\linewidth}%
\centering%
\includegraphics[width=\linewidth,keepaspectratio]{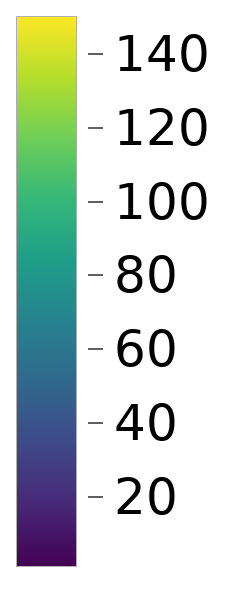}%
\end{minipage}%
}

\vspace{4pt}

\noindent\begin{tikzpicture}[baseline=(img.base)]
  \node[inner sep=0pt] (img) {\includegraphics[width=\linewidth]{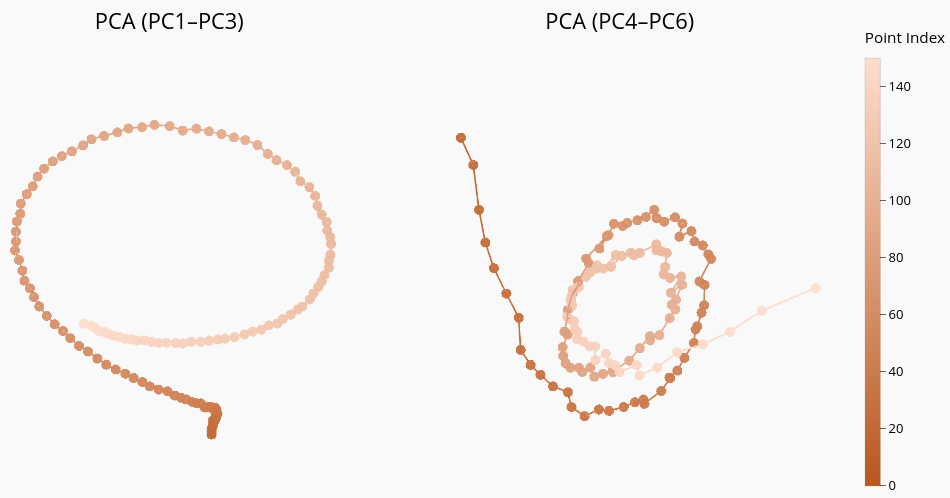}};
  \node[anchor=north west, inner sep=2pt, overlay] at (img.north west) {\small\textbf{(c)}};
\end{tikzpicture}

\caption{Newline counting manifold in Gemma 2 9B, Layer 11, on 150-char wrapped text. \textbf{(a, b)} SMIXAE bottleneck activations for the top two experts by $\Delta R^2_{\text{per}}$: (a) Expert 541, rank 1 (Score\,=\,0.548); (b) Expert 1521, rank 2 (Score\,=\,0.500). Points are per-class mean activations colored by number of characters since the previous newline. \textbf{(c)} Visualization from \citet{sinii}: PCA over the same layer's activations, showing the top 6 principal components with class means for the same target. The SMIXAE model directly learns the structures in two experts, while \citet{sinii} surface it through PCA analysis on the layer activations.}
\label{fig:newline_gemma_2_9b_l11}
\end{figure}

%% file: paper/table_newline.tex
\begin{table}[htbp]
\centering
\small
\caption{Characters since previous newline encoding in SMIXAE experts at layers 11 and 20 of Gemma 2 9B, evaluated at two different line lengths. The metric $\Delta R^2_{\text{per}} = R^2_{\text{periodic}} - R^2_{\text{linear}}$ measures whether the expert explains periodic labels better than it does linear labels in standard linear regression; larger positive values indicate geometry consistent with previously discovered newline counting manifolds. Top-1 is the score of the single best expert; Top-5$_{\mu}$ is the mean across the top-5 experts.}
\label{tab:newline}
\begin{tabular}{l cc cc}
\toprule
\multicolumn{1}{l}{Gemma 2 9B} & \multicolumn{2}{c}{Layer 11} & \multicolumn{2}{c}{Layer 20} \\
\cmidrule(lr){2-3} \cmidrule(lr){4-5}
Line Length ($\Delta R^2_{\text{per.}}$) & Top-1  & Top-5$_{\mu}$ & Top-1 & Top-5$_{\mu}$  \\
\midrule
80 chars & 0.558 & 0.290 & 0.507 & 0.178 \\
150 chars & 0.548 & 0.287 & 0.320 & 0.135 \\
\bottomrule
\end{tabular}
\end{table}

%% file: paper/probe_gemma_2_9b_l11.tex

\begin{figure*}[t]
\centering
\noindent\makebox[\linewidth][c]{%
\begin{minipage}[t]{0.3576\linewidth}%
\vspace{0pt}\centering%
\begin{minipage}[c][3.3cm][c]{0.7407\linewidth}%
\centering%
\begin{tikzpicture}[baseline=(img.base)]
  \node[inner sep=0pt] (img) {\includegraphics[width=\linewidth,height=3.3cm,keepaspectratio]{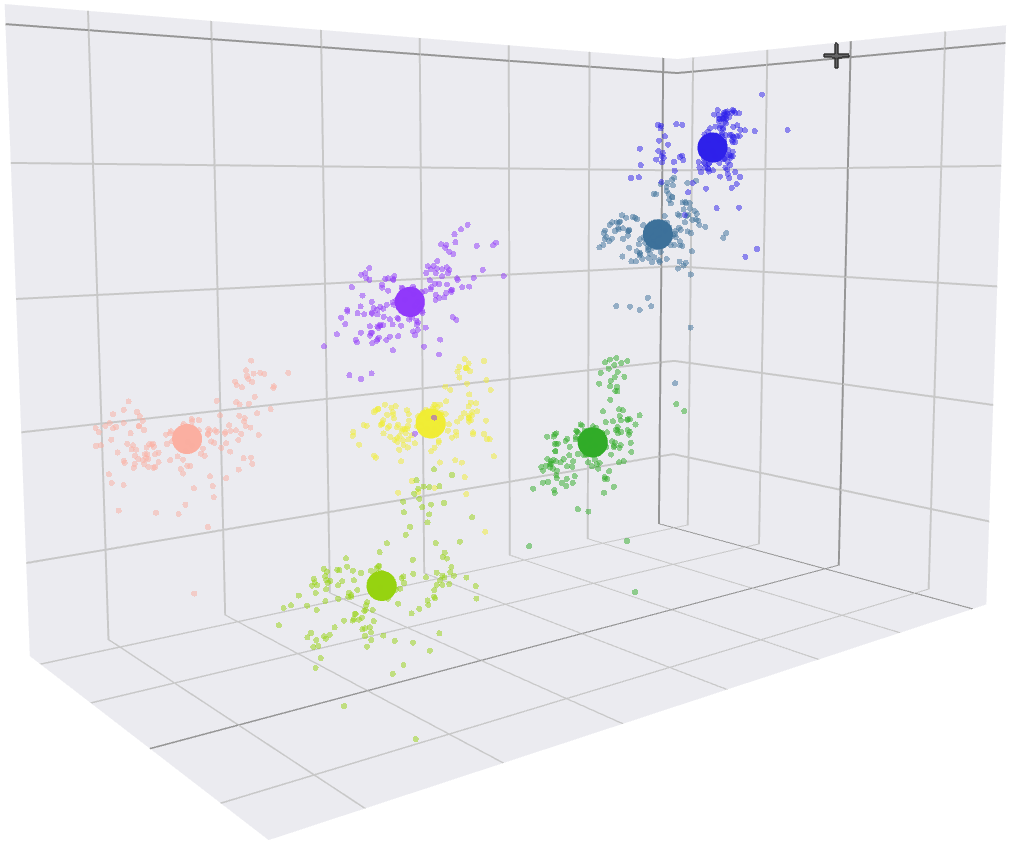}};
  \node[anchor=north west, inner sep=2pt, overlay] at (img.north west) {\small\textbf{(a)}};
\end{tikzpicture}%
\end{minipage}%
\begin{minipage}[c][3.3cm][c]{0.2593\linewidth}%
\centering%
\includegraphics[width=\linewidth,height=3.3cm,keepaspectratio]{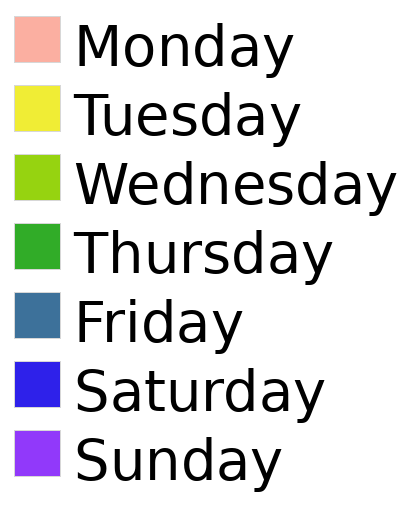}%
\end{minipage}%
\par\vspace{1pt}%
{\footnotesize\textit{Weekdays}}%
\end{minipage}%
\hspace{2.5mm}%
\begin{minipage}[t]{0.6224\linewidth}%
\vspace{0pt}\centering%
\begin{minipage}[c][3.3cm][c]{0.4255\linewidth}%
\centering%
\begin{tikzpicture}[baseline=(img.base)]
  \node[inner sep=0pt] (img) {\includegraphics[width=\linewidth,height=3.3cm,keepaspectratio]{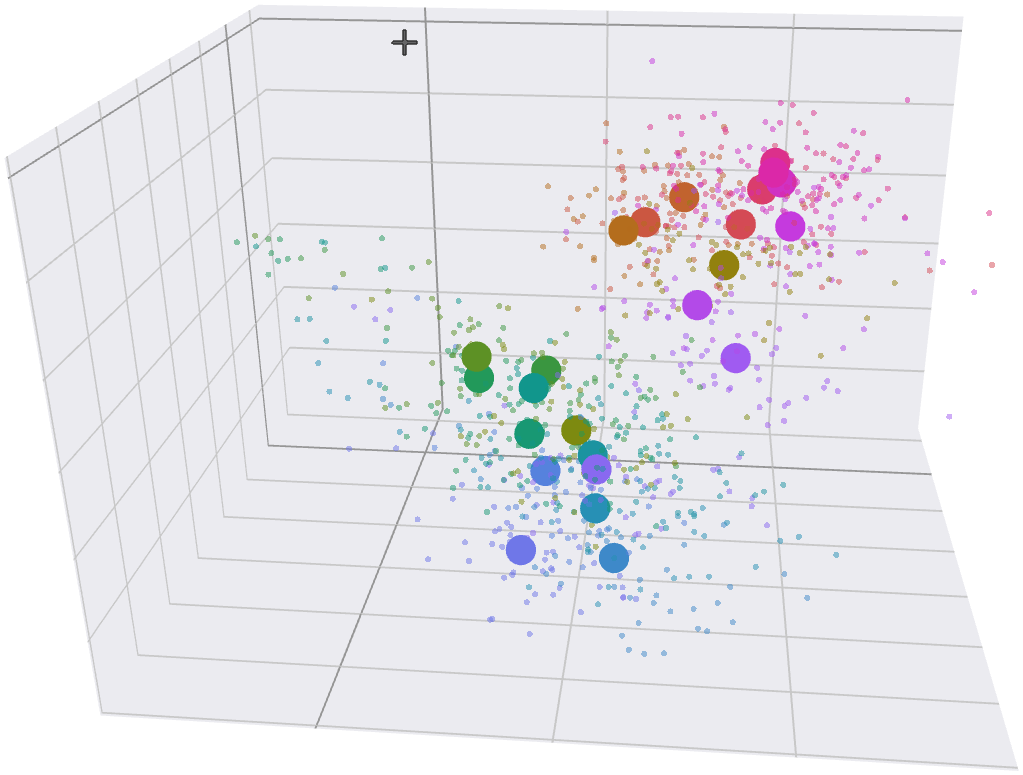}};
  \node[anchor=north west, inner sep=2pt, overlay] at (img.north west) {\small\textbf{(b)}};
\end{tikzpicture}%
\end{minipage}%
\begin{minipage}[c][3.3cm][c]{0.4255\linewidth}%
\centering%
\begin{tikzpicture}[baseline=(img.base)]
  \node[inner sep=0pt] (img) {\includegraphics[width=\linewidth,height=3.3cm,keepaspectratio]{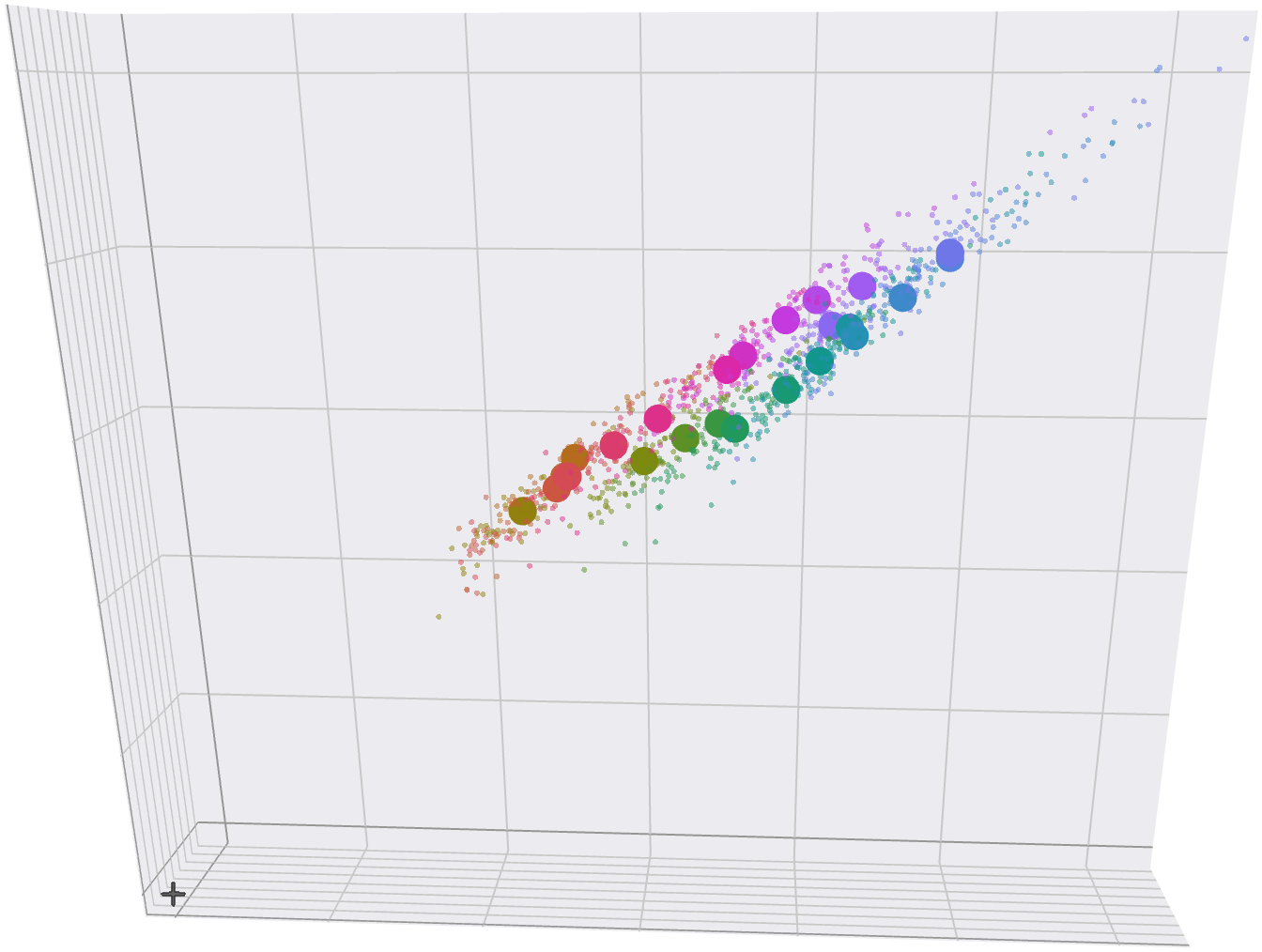}};
  \node[anchor=north west, inner sep=2pt, overlay] at (img.north west) {\small\textbf{(c)}};
\end{tikzpicture}%
\end{minipage}%
\begin{minipage}[c][3.3cm][c]{0.1489\linewidth}%
\centering%
\includegraphics[width=\linewidth,height=3.3cm,keepaspectratio]{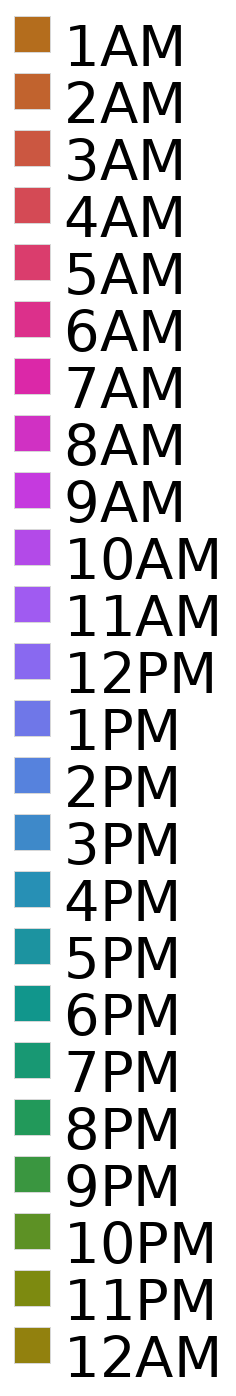}%
\end{minipage}%
\par\vspace{1pt}%
{\footnotesize\textit{Hours}}%
\end{minipage}%
}
\vspace{5pt}

\noindent\makebox[\linewidth][c]{%
\begin{minipage}[t]{0.3576\linewidth}%
\vspace{0pt}\centering%
\begin{minipage}[c][3.3cm][c]{0.7407\linewidth}%
\centering%
\begin{tikzpicture}[baseline=(img.base)]
  \node[inner sep=0pt] (img) {\includegraphics[width=\linewidth,height=3.3cm,keepaspectratio]{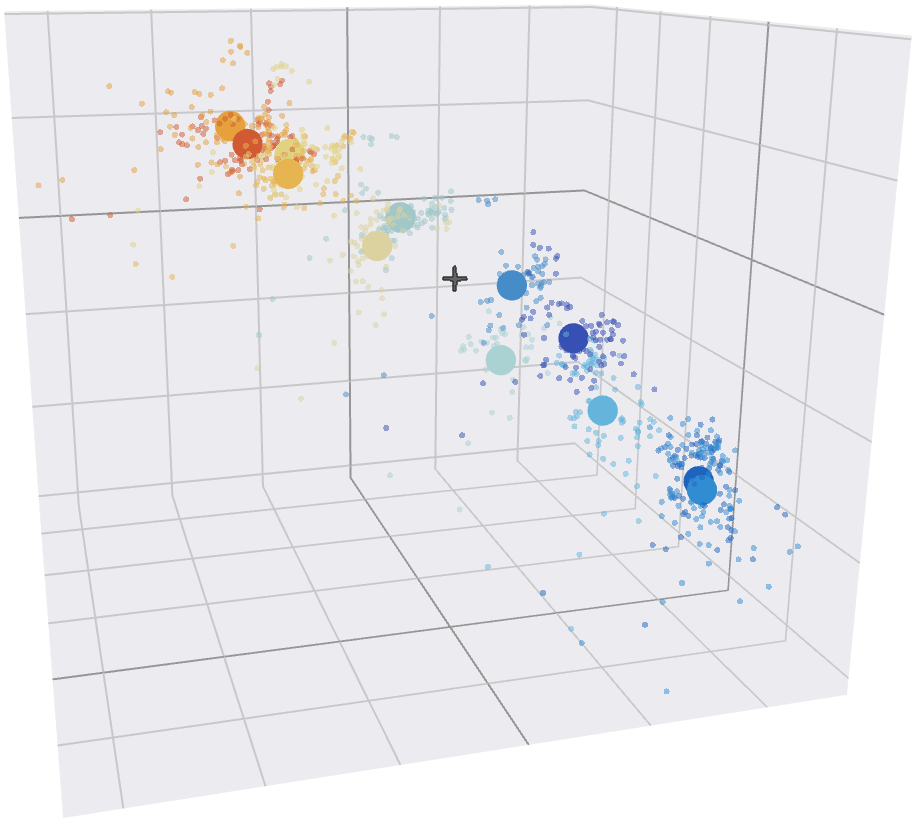}};
  \node[anchor=north west, inner sep=2pt, overlay] at (img.north west) {\small\textbf{(d)}};
\end{tikzpicture}%
\end{minipage}%
\begin{minipage}[c][3.3cm][c]{0.2593\linewidth}%
\centering%
\includegraphics[width=\linewidth,height=3.3cm,keepaspectratio]{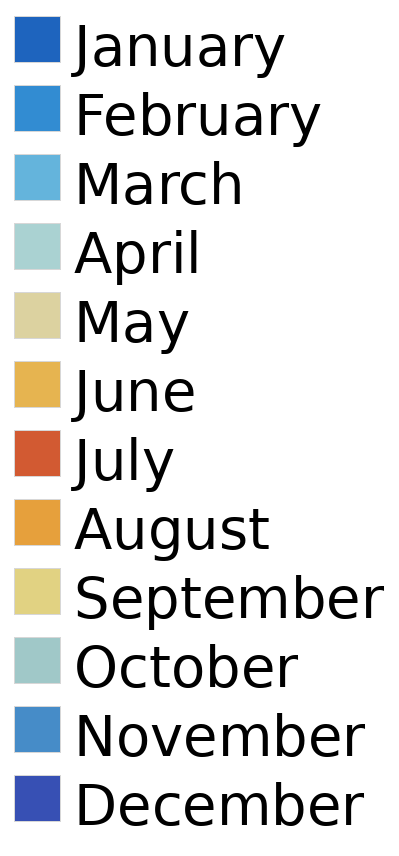}%
\end{minipage}%
\par\vspace{1pt}%
{\footnotesize\textit{Months}}%
\end{minipage}%
\hspace{2.5mm}%
\begin{minipage}[t]{0.6224\linewidth}%
\vspace{0pt}\centering%
\begin{minipage}[c][3.3cm][c]{0.4255\linewidth}%
\centering%
\begin{tikzpicture}[baseline=(img.base)]
  \node[inner sep=0pt] (img) {\includegraphics[width=\linewidth,height=3.3cm,keepaspectratio]{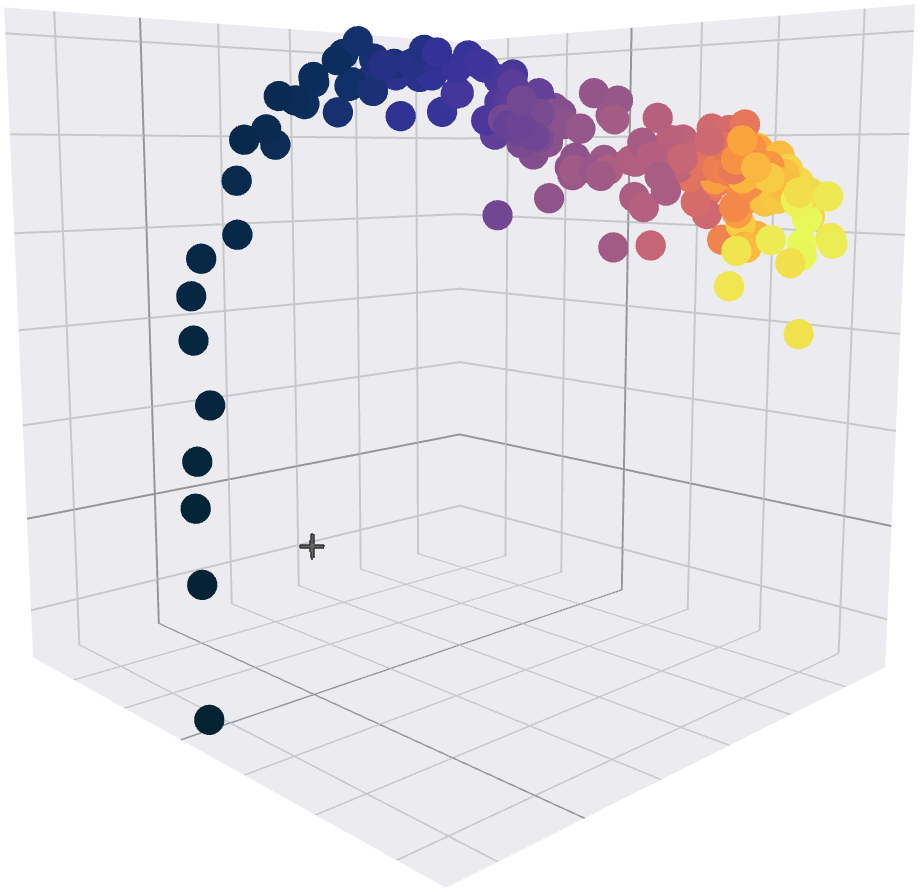}};
  \node[anchor=north west, inner sep=2pt, overlay] at (img.north west) {\small\textbf{(e)}};
\end{tikzpicture}%
\end{minipage}%
\begin{minipage}[c][3.3cm][c]{0.4255\linewidth}%
\centering%
\begin{tikzpicture}[baseline=(img.base)]
  \node[inner sep=0pt] (img) {\includegraphics[width=\linewidth,height=3.3cm,keepaspectratio]{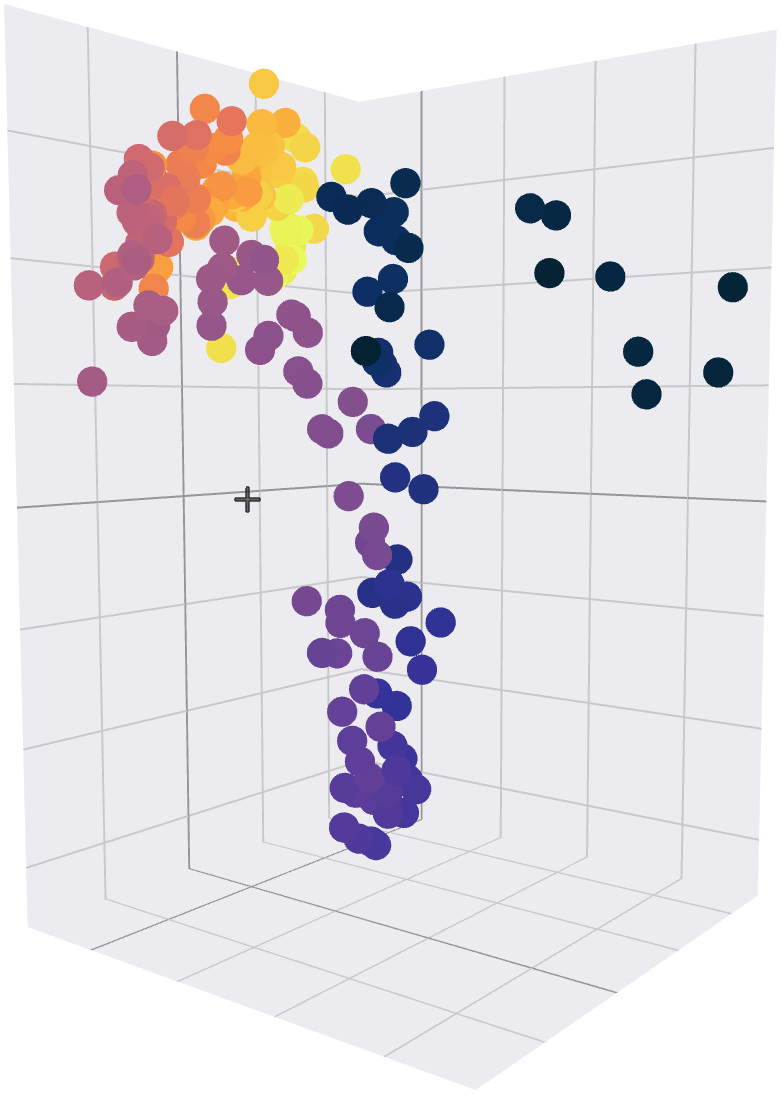}};
  \node[anchor=north west, inner sep=2pt, overlay] at (img.north west) {\small\textbf{(f)}};
\end{tikzpicture}%
\end{minipage}%
\begin{minipage}[c][3.3cm][c]{0.1489\linewidth}%
\centering%
\includegraphics[width=\linewidth,height=3.3cm,keepaspectratio]{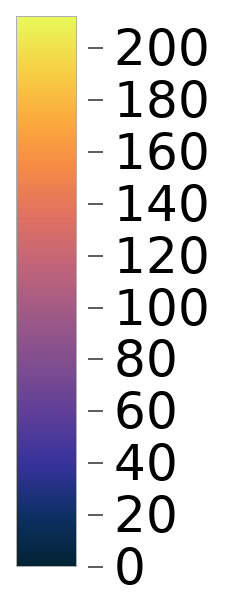}%
\end{minipage}%
\par\vspace{1pt}%
{\footnotesize\textit{Temperature}}%
\end{minipage}%
}
\vspace{5pt}

\noindent\makebox[\linewidth][c]{%
\begin{minipage}[t]{0.3949\linewidth}%
\vspace{0pt}\centering%
\begin{minipage}[c][3.3cm][c]{0.7407\linewidth}%
\centering%
\begin{tikzpicture}[baseline=(img.base)]
  \node[inner sep=0pt] (img) {\includegraphics[width=\linewidth,height=3.3cm,keepaspectratio]{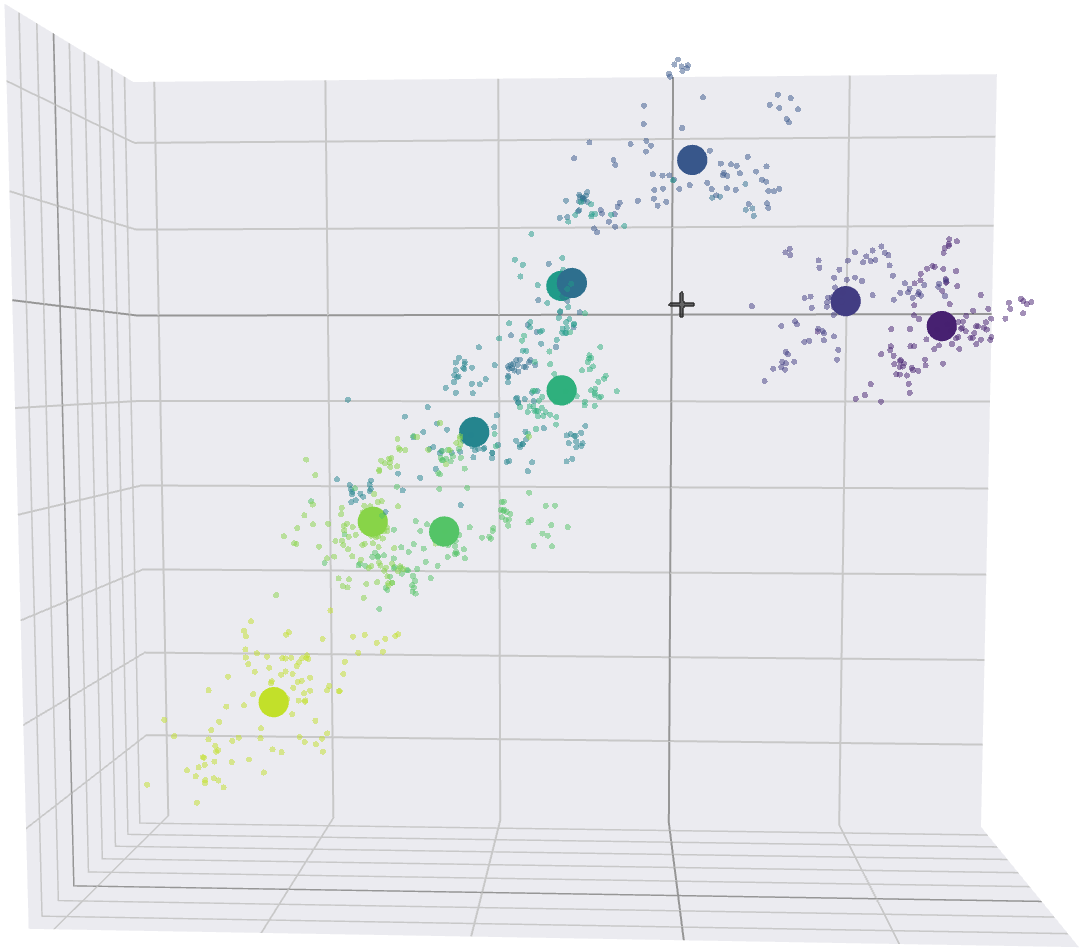}};
  \node[anchor=north west, inner sep=2pt, overlay] at (img.north west) {\small\textbf{(g)}};
\end{tikzpicture}%
\end{minipage}%
\begin{minipage}[c][3.3cm][c]{0.2593\linewidth}%
\centering%
\includegraphics[width=\linewidth,height=3.3cm,keepaspectratio]{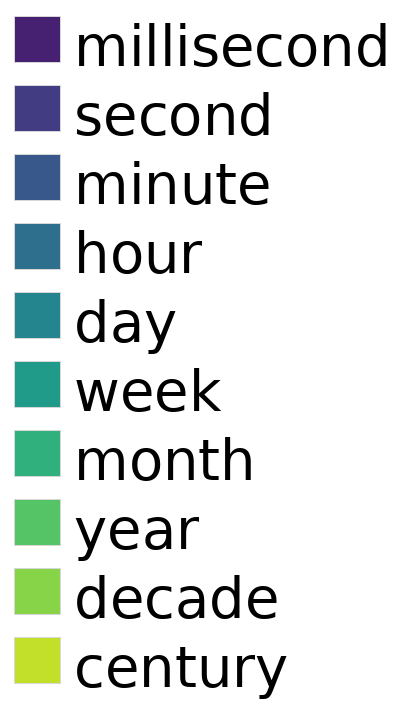}%
\end{minipage}%
\par\vspace{1pt}%
{\footnotesize\textit{Time Units}}%
\end{minipage}%
\hspace{2.5mm}%
\begin{minipage}[t]{0.3949\linewidth}%
\vspace{0pt}\centering%
\begin{minipage}[c][3.3cm][c]{0.7407\linewidth}%
\centering%
\begin{tikzpicture}[baseline=(img.base)]
  \node[inner sep=0pt] (img) {\includegraphics[width=\linewidth,height=3.3cm,keepaspectratio]{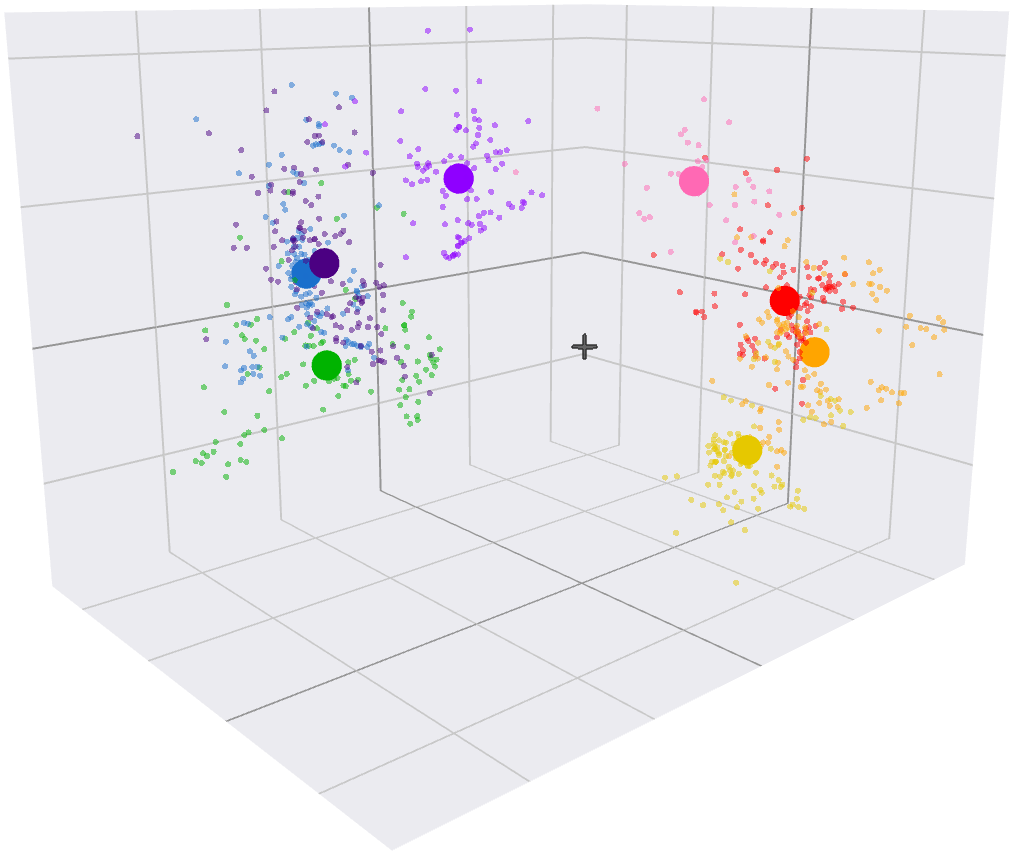}};
  \node[anchor=north west, inner sep=2pt, overlay] at (img.north west) {\small\textbf{(h)}};
\end{tikzpicture}%
\end{minipage}%
\begin{minipage}[c][3.3cm][c]{0.2593\linewidth}%
\centering%
\includegraphics[width=\linewidth,height=3.3cm,keepaspectratio]{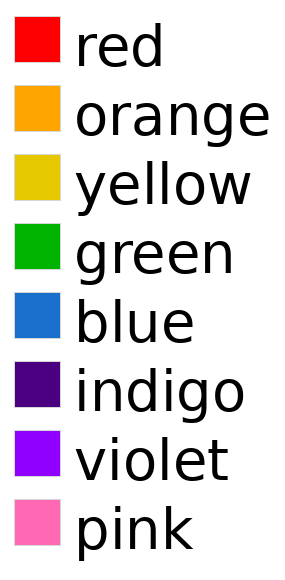}%
\end{minipage}%
\par\vspace{1pt}%
{\footnotesize\textit{Colors}}%
\end{minipage}%
}
\caption{SMIXAE Bottleneck Activations on Gemma 2 9B, Layer 11. Each plot shows the 3-D bottleneck activations of a single SMIXAE expert; small points are individual token activations colored by ground-truth label, and larger points mark per-class means. The $+$ symbol marks the origin of the plot. \textbf{Weekdays}: (a) Expert 76, rank 1, 7-Day Ring ($R^2$\,=\,0.855).  \textbf{Hours}: (b) Expert 1884, rank 2, AM vs PM (Accuracy\,=\,0.997). (c) Expert 1965, rank 2, 24-Hour Ring ($R^2$\,=\,0.653).  \textbf{Months}: (d) Expert 1942, rank 2, 12-Month Ring ($R^2$\,=\,0.602).  \textbf{Temperature}: (e) Expert 892, rank 1, Fahrenheit ($R^2$\,=\,0.916). (f) Expert 485, rank 2, Fahrenheit ($R^2$\,=\,0.729).  \textbf{Time Units}: (g) Expert 786, rank 1, $\log\_{10}$ Duration ($R^2$\,=\,0.894).  \textbf{Colors}: (h) Expert 500, rank 1, Hue Ring ($R^2$\,=\,0.827).}
\label{fig:probe_gemma_2_9b_l11}
\end{figure*}

%% file: paper/table_core_eval.tex
\begin{table*}[htbp]
\centering
\small
\caption{Core evaluation metrics on SMIXAE, with GemmaScope SAE baselines as reference, evaluated on OpenWebText (128-token context windows). L0, width, and fraction alive are calculated on the unflattened latent space of SMIXAE. SMIXAE performs reasonably well on all evaluation metrics, and, importantly, has not over-optimized reconstruction. This shows that, while parameter capacity drastically increased, the bottleneck design effectively mitigates over-optimization. }
\label{tab:saebench}
\resizebox{\textwidth}{!}{%
\begin{tabular}{lll c c c c c c c c}
\toprule
Model & Layer & Autoencoder & Width & Parameters & L0 & Frac. Alive & Expl. Var. & CE Score & MSE & Cos. Sim. \\
\midrule
\multirow{2}{*}{Gemma 2 2B} & \multirow{2}{*}{12} & SMIXAE & 6,144 & 151,226,624 & 230.673 & 0.970 & 0.752 & 0.984 & 0.189 & 0.900 \\
 &  & GemmaScope 2B 16k & 16,384 & 75,532,544 & 184.550 & 0.999 & 0.841 & 0.994 & 0.121 & 0.938 \\
\midrule
\multirow{2}{*}{Gemma 2 9B} & \multirow{2}{*}{11} & SMIXAE & 6,144 & 235,113,984 & 229.361 & 0.988 & 0.655 & 0.985 & 0.231 & 0.876 \\
 &  & GemmaScope 9B 16k & 16,384 & 117,476,864 & 130.270 & 0.999 & 0.774 & 0.995 & 0.150 & 0.922 \\
 \midrule
 \multirow{2}{*}{Gemma 2 9B} & \multirow{2}{*}{20} & SMIXAE & 6,144 & 235,113,984 & 211.989 & 0.967 & 0.735 & 0.978 & 0.202 & 0.894 \\
 &  & GemmaScope 9B 16k & 16,384 & 117,476,864 & 139.525 & 0.999 & 0.819 & 0.991 & 0.138 & 0.928 \\
\bottomrule
\end{tabular}
}
\end{table*}

%% file: paper/table_probing.tex
\begin{table*}[htbp]
\centering
\small
\caption{Probing results for SMIXAE experts across several tasks. Each hypothesis targets a structured property that may be geometrically encoded in a 3-D expert bottleneck. For each hypothesis we fit a regression or classifier directly to the bottleneck activations of the top-performing experts and report the score of the single best expert (Top-1) and the mean over the top-5 experts (Top-5$_{\mu}$). $R^2$ is the coefficient of determination (linear and ridge regression); Acc.\ is classification accuracy (logistic and multinomial regression).}
\label{tab:probing}

\begin{tabular}{ll ll rr rr rr}
\toprule
\multicolumn{4}{l}{Gemma 2} & \multicolumn{2}{c}{9B, Layer 11} & \multicolumn{2}{c}{9B, Layer 20} & \multicolumn{2}{c}{2B, Layer 12} \\
\cmidrule(lr){5-6} \cmidrule(lr){7-8} \cmidrule(lr){9-10}
Task & Hypothesis & Regression & Score & Top-1 & Top-5$_{\mu}$ & Top-1 & Top-5$_{\mu}$ & Top-1 & Top-5$_{\mu}$ \\
\midrule
\multirow{2}{*}{Weekdays} & 7-Day Ring & Linear & $R^2$ & 0.855 & 0.751 & 0.724 & 0.594 & 0.579 & 0.487 \\
 & Mo-Fr vs Sa-Su & Logistic & Acc. & 1.000 & 0.994 & 0.998 & 0.944 & 0.973 & 0.935 \\
\midrule
\multirow{3}{*}{Hours} & 24-Hour Ring & Linear & $R^2$ & 0.671 & 0.593 & 0.590 & 0.494 & 0.681 & 0.491 \\
 & 12-Hour Ring & Linear & $R^2$ & 0.657 & 0.539 & 0.397 & 0.266 & 0.425 & 0.333 \\
 & AM vs PM & Logistic & Acc. & 1.000 & 0.984 & 1.000 & 0.911 & 1.000 & 0.939 \\
\midrule
\multirow{2}{*}{Temperature} & Linear Fahrenheit & Linear & $R^2$ & 0.916 & 0.610 & 0.875 & 0.531 & 0.663 & 0.559 \\
 & $\log(\text{Fahrenheit})$ & Linear & $R^2$ & 0.950 & 0.629 & 0.887 & 0.510 & 0.682 & 0.509 \\
\midrule
Time Units & $\log_{10}(\text{Duration})$ & Linear & $R^2$ & 0.894 & 0.737 & 0.856 & 0.672 & 0.695 & 0.617 \\
\midrule
\multirow{2}{*}{Living Things} & Plant vs Animal & Logistic & Acc. & 0.963 & 0.946 & 0.914 & 0.886 & 0.998 & 0.965 \\
 & Taxonomic Group & Multinomial & Acc. & 0.926 & 0.822 & 0.780 & 0.666 & 0.911 & 0.781 \\
\midrule
\multirow{2}{*}{Months} & 12-Month Ring & Linear & $R^2$ & 0.659 & 0.411 & 0.469 & 0.215 & 0.487 & 0.201 \\
 & Season & Multinomial & Acc. & 0.838 & 0.704 & 0.872 & 0.531 & 0.674 & 0.464 \\
\midrule
\multirow{3}{*}{Colors} & Hue Ring & Linear & $R^2$ & 0.827 & 0.775 & 0.575 & 0.441 & 0.655 & 0.546 \\
 & Normalized RGB & Linear & $R^2$ & 0.789 & 0.689 & 0.452 & 0.400 & 0.596 & 0.491 \\
 & Warm/Natural/Cool & Multinomial & Acc. & 0.995 & 0.947 & 0.888 & 0.806 & 0.865 & 0.844 \\
\bottomrule
\end{tabular}
\end{table*}

%% file: paper/table_probing_appendix.tex
\begin{table*}[htbp]
\centering
\small
\caption{Complete probing scores for all top-10 experts in 9B, Layer 11, listed per task and hypothesis. Columns 1--10 are rank positions; each cell shows the regression score. The same expert may appear under multiple hypotheses if it encodes more than one concept.}
\label{tab:probing_appendix_gemma_2_9b_l11}
\resizebox{\linewidth}{!}{%
\begin{tabular}{ll ll rrrrrrrrrr}
\toprule
 & & & & \multicolumn{10}{c}{Expert rank} \\
\cmidrule(lr){5-14}
Task & Hypothesis & Regression & Score & 1 & 2 & 3 & 4 & 5 & 6 & 7 & 8 & 9 & 10 \\
\midrule
\multirow{2}{*}{Weekdays} & 7-Day Ring & Linear & $R^2$ & 0.855 & 0.742 & 0.742 & 0.709 & 0.708 & 0.698 & 0.671 & 0.619 & 0.567 & 0.523 \\
 & Mo-Fr vs Sa-Su & Logistic & Acc. & 1.000 & 0.999 & 0.999 & 0.988 & 0.982 & 0.981 & 0.976 & 0.964 & 0.960 & 0.921 \\
\midrule
\multirow{3}{*}{Hours} & 24-Hour Ring & Linear & $R^2$ & 0.671 & 0.653 & 0.582 & 0.543 & 0.517 & 0.509 & 0.504 & 0.489 & 0.476 & 0.448 \\
 & 12-Hour Ring & Linear & $R^2$ & 0.657 & 0.652 & 0.552 & 0.435 & 0.399 & 0.384 & 0.353 & 0.298 & 0.255 & 0.224 \\
 & AM vs PM & Logistic & Acc. & 1.000 & 0.997 & 0.992 & 0.982 & 0.949 & 0.949 & 0.943 & 0.939 & 0.937 & 0.894 \\
\midrule
\multirow{2}{*}{Temperature} & Linear Fahrenheit & Linear & $R^2$ & 0.916 & 0.729 & 0.496 & 0.454 & 0.453 & 0.403 & 0.392 & 0.386 & 0.377 & 0.361 \\
 & $\log(\text{Fahrenheit})$ & Linear & $R^2$ & 0.950 & 0.750 & 0.522 & 0.495 & 0.430 & 0.378 & 0.328 & 0.290 & 0.288 & 0.284 \\
\midrule
Time Units & $\log_{10}(\text{Duration})$ & Linear & $R^2$ & 0.894 & 0.764 & 0.692 & 0.679 & 0.658 & 0.648 & 0.641 & 0.630 & 0.605 & 0.573 \\
\midrule
\multirow{2}{*}{Living Things} & Plant vs Animal & Logistic & Acc. & 0.963 & 0.945 & 0.944 & 0.941 & 0.935 & 0.935 & 0.929 & 0.927 & 0.921 & 0.908 \\
 & Taxonomic Group & Multinomial & Acc. & 0.926 & 0.808 & 0.800 & 0.793 & 0.783 & 0.780 & 0.766 & 0.765 & 0.751 & 0.734 \\
\midrule
\multirow{2}{*}{Months} & 12-Month Ring & Linear & $R^2$ & 0.659 & 0.602 & 0.485 & 0.161 & 0.148 & 0.145 & 0.144 & 0.133 & 0.077 & 0.077 \\
 & Season & Multinomial & Acc. & 0.838 & 0.835 & 0.757 & 0.547 & 0.546 & 0.447 & 0.436 & 0.433 & 0.431 & 0.423 \\
\midrule
\multirow{3}{*}{Colors} & Hue Ring & Linear & $R^2$ & 0.827 & 0.791 & 0.784 & 0.758 & 0.712 & 0.583 & 0.578 & 0.506 & 0.487 & 0.486 \\
 & Normalized RGB & Linear & $R^2$ & 0.789 & 0.683 & 0.682 & 0.658 & 0.631 & 0.627 & 0.615 & 0.509 & 0.492 & 0.477 \\
 & Warm/Natural/Cool & Multinomial & Acc. & 0.995 & 0.971 & 0.948 & 0.939 & 0.881 & 0.881 & 0.852 & 0.833 & 0.829 & 0.788 \\
\bottomrule
\end{tabular}}
\end{table*}

\begin{table*}[htbp]
\centering
\small
\caption{Complete probing scores for all top-10 experts in 9B, Layer 20, listed per task and hypothesis. Columns 1--10 are rank positions; each cell shows the regression score. The same expert may appear under multiple hypotheses if it encodes more than one concept.}
\label{tab:probing_appendix_gemma_2_9b_l20}
\resizebox{\linewidth}{!}{%
\begin{tabular}{ll ll rrrrrrrrrr}
\toprule
 & & & & \multicolumn{10}{c}{Expert rank} \\
\cmidrule(lr){5-14}
Task & Hypothesis & Regression & Score & 1 & 2 & 3 & 4 & 5 & 6 & 7 & 8 & 9 & 10 \\
\midrule
\multirow{2}{*}{Weekdays} & 7-Day Ring & Linear & $R^2$ & 0.724 & 0.690 & 0.595 & 0.481 & 0.478 & 0.399 & 0.135 & 0.133 & 0.116 & 0.085 \\
 & Mo-Fr vs Sa-Su & Logistic & Acc. & 0.998 & 0.975 & 0.933 & 0.910 & 0.903 & 0.840 & 0.748 & 0.745 & 0.730 & 0.723 \\
\midrule
\multirow{3}{*}{Hours} & 24-Hour Ring & Linear & $R^2$ & 0.590 & 0.545 & 0.468 & 0.437 & 0.430 & 0.421 & 0.395 & 0.362 & 0.355 & 0.340 \\
 & 12-Hour Ring & Linear & $R^2$ & 0.397 & 0.262 & 0.233 & 0.225 & 0.214 & 0.174 & 0.166 & 0.143 & 0.142 & 0.138 \\
 & AM vs PM & Logistic & Acc. & 1.000 & 0.997 & 0.877 & 0.858 & 0.826 & 0.822 & 0.807 & 0.792 & 0.765 & 0.758 \\
\midrule
\multirow{2}{*}{Temperature} & Linear Fahrenheit & Linear & $R^2$ & 0.875 & 0.530 & 0.530 & 0.390 & 0.331 & 0.321 & 0.303 & 0.276 & 0.253 & 0.251 \\
 & $\log(\text{Fahrenheit})$ & Linear & $R^2$ & 0.887 & 0.468 & 0.433 & 0.387 & 0.376 & 0.271 & 0.265 & 0.263 & 0.247 & 0.222 \\
\midrule
Time Units & $\log_{10}(\text{Duration})$ & Linear & $R^2$ & 0.856 & 0.704 & 0.627 & 0.621 & 0.553 & 0.515 & 0.502 & 0.474 & 0.442 & 0.441 \\
\midrule
\multirow{2}{*}{Living Things} & Plant vs Animal & Logistic & Acc. & 0.914 & 0.904 & 0.879 & 0.870 & 0.865 & 0.865 & 0.852 & 0.848 & 0.830 & 0.827 \\
 & Taxonomic Group & Multinomial & Acc. & 0.780 & 0.717 & 0.675 & 0.610 & 0.547 & 0.488 & 0.468 & 0.463 & 0.437 & 0.428 \\
\midrule
\multirow{2}{*}{Months} & 12-Month Ring & Linear & $R^2$ & 0.469 & 0.206 & 0.175 & 0.171 & 0.056 & 0.051 & 0.041 & 0.033 & 0.032 & 0.028 \\
 & Season & Multinomial & Acc. & 0.872 & 0.513 & 0.462 & 0.442 & 0.369 & 0.330 & 0.326 & 0.315 & 0.311 & 0.306 \\
\midrule
\multirow{3}{*}{Colors} & Hue Ring & Linear & $R^2$ & 0.575 & 0.457 & 0.443 & 0.368 & 0.364 & 0.337 & 0.327 & 0.296 & 0.285 & 0.202 \\
 & Normalized RGB & Linear & $R^2$ & 0.452 & 0.452 & 0.413 & 0.365 & 0.316 & 0.313 & 0.311 & 0.249 & 0.245 & 0.193 \\
 & Warm/Natural/Cool & Multinomial & Acc. & 0.888 & 0.881 & 0.774 & 0.752 & 0.733 & 0.726 & 0.664 & 0.657 & 0.655 & 0.589 \\
\bottomrule
\end{tabular}}
\end{table*}

\begin{table*}[htbp]
\centering
\small
\caption{Complete probing scores for all top-10 experts in 2B, Layer 12, listed per task and hypothesis. Columns 1--10 are rank positions; each cell shows the regression score. The same expert may appear under multiple hypotheses if it encodes more than one concept.}
\label{tab:probing_appendix_gemma_2_2b_l12}
\resizebox{\linewidth}{!}{%
\begin{tabular}{ll ll rrrrrrrrrr}
\toprule
 & & & & \multicolumn{10}{c}{Expert rank} \\
\cmidrule(lr){5-14}
Task & Hypothesis & Regression & Score & 1 & 2 & 3 & 4 & 5 & 6 & 7 & 8 & 9 & 10 \\
\midrule
\multirow{2}{*}{Weekdays} & 7-Day Ring & Linear & $R^2$ & 0.579 & 0.531 & 0.485 & 0.456 & 0.383 & 0.356 & 0.338 & 0.222 & 0.181 & 0.090 \\
 & Mo-Fr vs Sa-Su & Logistic & Acc. & 0.973 & 0.962 & 0.951 & 0.913 & 0.876 & 0.846 & 0.818 & 0.811 & 0.737 & 0.695 \\
\midrule
\multirow{3}{*}{Hours} & 24-Hour Ring & Linear & $R^2$ & 0.681 & 0.527 & 0.462 & 0.395 & 0.389 & 0.363 & 0.342 & 0.341 & 0.338 & 0.297 \\
 & 12-Hour Ring & Linear & $R^2$ & 0.425 & 0.362 & 0.356 & 0.317 & 0.206 & 0.203 & 0.180 & 0.165 & 0.142 & 0.124 \\
 & AM vs PM & Logistic & Acc. & 1.000 & 0.986 & 0.933 & 0.930 & 0.845 & 0.839 & 0.758 & 0.744 & 0.741 & 0.739 \\
\midrule
\multirow{2}{*}{Temperature} & Linear Fahrenheit & Linear & $R^2$ & 0.663 & 0.597 & 0.539 & 0.509 & 0.487 & 0.413 & 0.396 & 0.371 & 0.360 & 0.339 \\
 & $\log(\text{Fahrenheit})$ & Linear & $R^2$ & 0.682 & 0.515 & 0.492 & 0.436 & 0.422 & 0.419 & 0.409 & 0.351 & 0.344 & 0.303 \\
\midrule
Time Units & $\log_{10}(\text{Duration})$ & Linear & $R^2$ & 0.695 & 0.686 & 0.611 & 0.562 & 0.532 & 0.347 & 0.308 & 0.305 & 0.248 & 0.239 \\
\midrule
\multirow{2}{*}{Living Things} & Plant vs Animal & Logistic & Acc. & 0.998 & 0.986 & 0.966 & 0.956 & 0.920 & 0.914 & 0.906 & 0.902 & 0.901 & 0.861 \\
 & Taxonomic Group & Multinomial & Acc. & 0.911 & 0.779 & 0.772 & 0.726 & 0.717 & 0.691 & 0.688 & 0.674 & 0.670 & 0.662 \\
\midrule
\multirow{2}{*}{Months} & 12-Month Ring & Linear & $R^2$ & 0.487 & 0.281 & 0.114 & 0.076 & 0.048 & 0.020 & 0.017 & 0.015 & 0.013 & 0.012 \\
 & Season & Multinomial & Acc. & 0.674 & 0.540 & 0.392 & 0.366 & 0.348 & 0.296 & 0.294 & 0.277 & 0.276 & 0.275 \\
\midrule
\multirow{3}{*}{Colors} & Hue Ring & Linear & $R^2$ & 0.655 & 0.654 & 0.600 & 0.421 & 0.401 & 0.337 & 0.331 & 0.316 & 0.306 & 0.282 \\
 & Normalized RGB & Linear & $R^2$ & 0.596 & 0.526 & 0.488 & 0.467 & 0.378 & 0.376 & 0.358 & 0.338 & 0.331 & 0.268 \\
 & Warm/Natural/Cool & Multinomial & Acc. & 0.865 & 0.858 & 0.852 & 0.835 & 0.808 & 0.763 & 0.756 & 0.710 & 0.690 & 0.678 \\
\bottomrule
\end{tabular}}
\end{table*}

%% file: paper/table_newline_appendix.tex
\begin{table}[htbp]
\centering
\small
\caption{Complete newline-position probing scores for all top-10 experts in 9B, Layer 11 at each line length, ranked by $\Delta R^2_{\text{periodic}}$. This table supports Table\ref{tab:newline}.}
\label{tab:newline_appendix_gemma_2_9b_l11}
\resizebox{\linewidth}{!}{%
\begin{tabular}{l r r r r r r r r r r}
\toprule
 & \multicolumn{10}{c}{Expert rank} \\
\cmidrule(lr){2-11}
Line length & 1 & 2 & 3 & 4 & 5 & 6 & 7 & 8 & 9 & 10 \\
\midrule
80 chars & 0.558 & 0.430 & 0.392 & 0.039 & 0.032 & 0.026 & 0.025 & 0.018 & 0.018 & 0.015 \\
\midrule
150 chars & 0.548 & 0.500 & 0.317 & 0.041 & 0.028 & 0.027 & 0.027 & 0.015 & 0.010 & 0.009 \\
\bottomrule
\end{tabular}}
\end{table}

\begin{table}[htbp]
\centering
\small
\caption{Complete newline-position probing scores for all top-10 experts in 9B, Layer 20 at each line length, ranked by $\Delta R^2_{\text{periodic}}$. This table supports Table\ref{tab:newline}.}
\label{tab:newline_appendix_gemma_2_9b_l20}
\resizebox{\linewidth}{!}{%
\begin{tabular}{l r r r r r r r r r r}
\toprule
 & \multicolumn{10}{c}{Expert rank} \\
\cmidrule(lr){2-11}
Line length & 1 & 2 & 3 & 4 & 5 & 6 & 7 & 8 & 9 & 10 \\
\midrule
80 chars & 0.507 & 0.303 & 0.031 & 0.025 & 0.024 & 0.023 & 0.022 & 0.022 & 0.020 & 0.015 \\
\midrule
150 chars & 0.320 & 0.280 & 0.026 & 0.025 & 0.024 & 0.024 & 0.021 & 0.020 & 0.017 & 0.011 \\
\bottomrule
\end{tabular}}
\end{table}

%% file: paper/probe_gemma_2_2b_l12.tex

\begin{figure*}[t]
\centering
\noindent\makebox[\linewidth][c]{%
\begin{minipage}[t]{0.3267\linewidth}%
\vspace{0pt}\centering%
\begin{minipage}[c][4.0cm][c]{0.7407\linewidth}%
\centering%
\begin{tikzpicture}[baseline=(img.base)]
  \node[inner sep=0pt] (img) {\includegraphics[width=\linewidth,height=4.0cm,keepaspectratio]{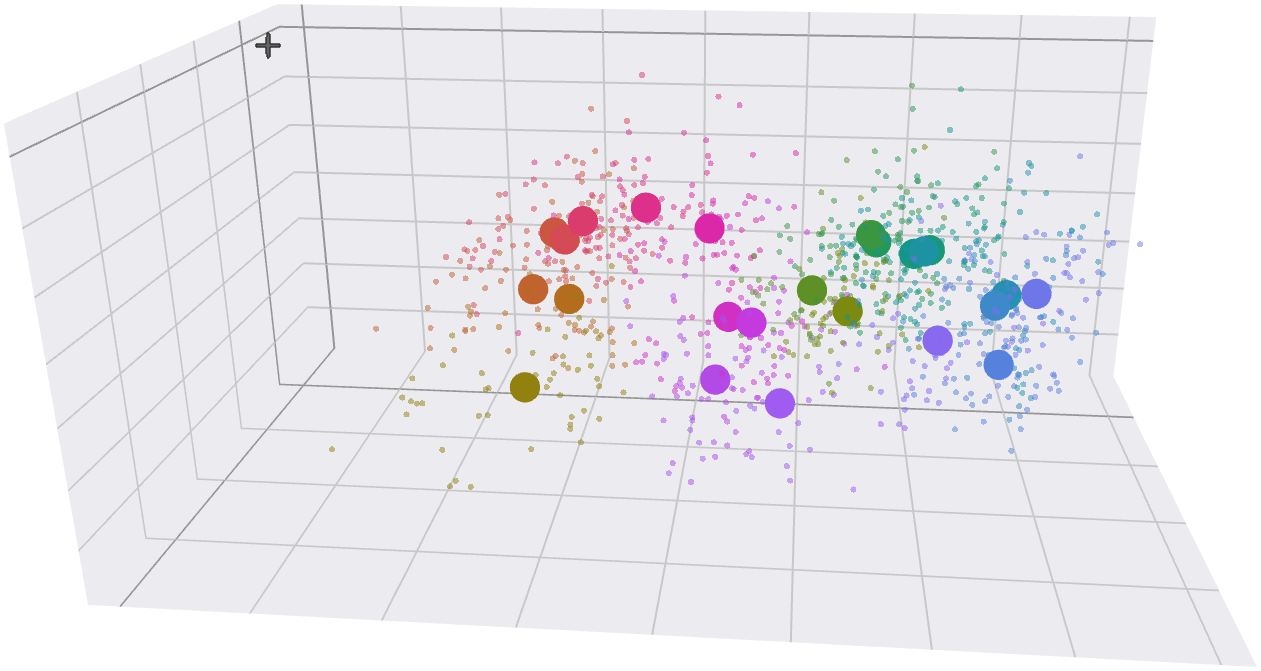}};
  \node[anchor=north west, inner sep=2pt, overlay] at (img.north west) {\small\textbf{(a)}};
\end{tikzpicture}%
\end{minipage}%
\begin{minipage}[c][4.0cm][c]{0.2593\linewidth}%
\centering%
\includegraphics[width=\linewidth,height=4.0cm,keepaspectratio]{paper/legends/legend_hours.png}%
\end{minipage}%
\par\vspace{1pt}%
{\footnotesize\textit{Hours}}%
\end{minipage}%
\hspace{2.5mm}%
\begin{minipage}[t]{0.3267\linewidth}%
\vspace{0pt}\centering%
\begin{minipage}[c][4.0cm][c]{0.7407\linewidth}%
\centering%
\begin{tikzpicture}[baseline=(img.base)]
  \node[inner sep=0pt] (img) {\includegraphics[width=\linewidth,height=4.0cm,keepaspectratio]{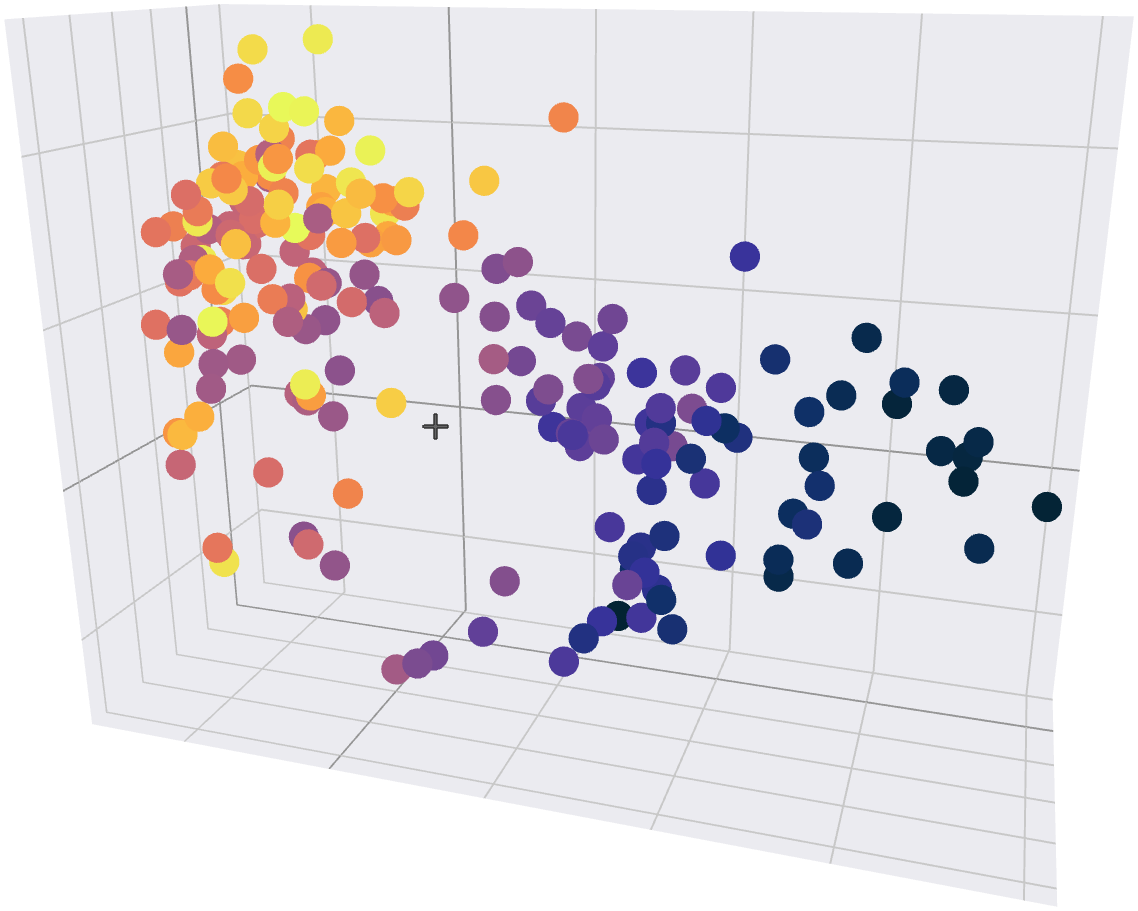}};
  \node[anchor=north west, inner sep=2pt, overlay] at (img.north west) {\small\textbf{(b)}};
\end{tikzpicture}%
\end{minipage}%
\begin{minipage}[c][4.0cm][c]{0.2593\linewidth}%
\centering%
\includegraphics[width=\linewidth,height=4.0cm,keepaspectratio]{paper/legends/legend_temperatures.png}%
\end{minipage}%
\par\vspace{1pt}%
{\footnotesize\textit{Temperature}}%
\end{minipage}%
\hspace{2.5mm}%
\begin{minipage}[t]{0.3267\linewidth}%
\vspace{0pt}\centering%
\begin{minipage}[c][4.0cm][c]{0.7407\linewidth}%
\centering%
\begin{tikzpicture}[baseline=(img.base)]
  \node[inner sep=0pt] (img) {\includegraphics[width=\linewidth,height=4.0cm,keepaspectratio]{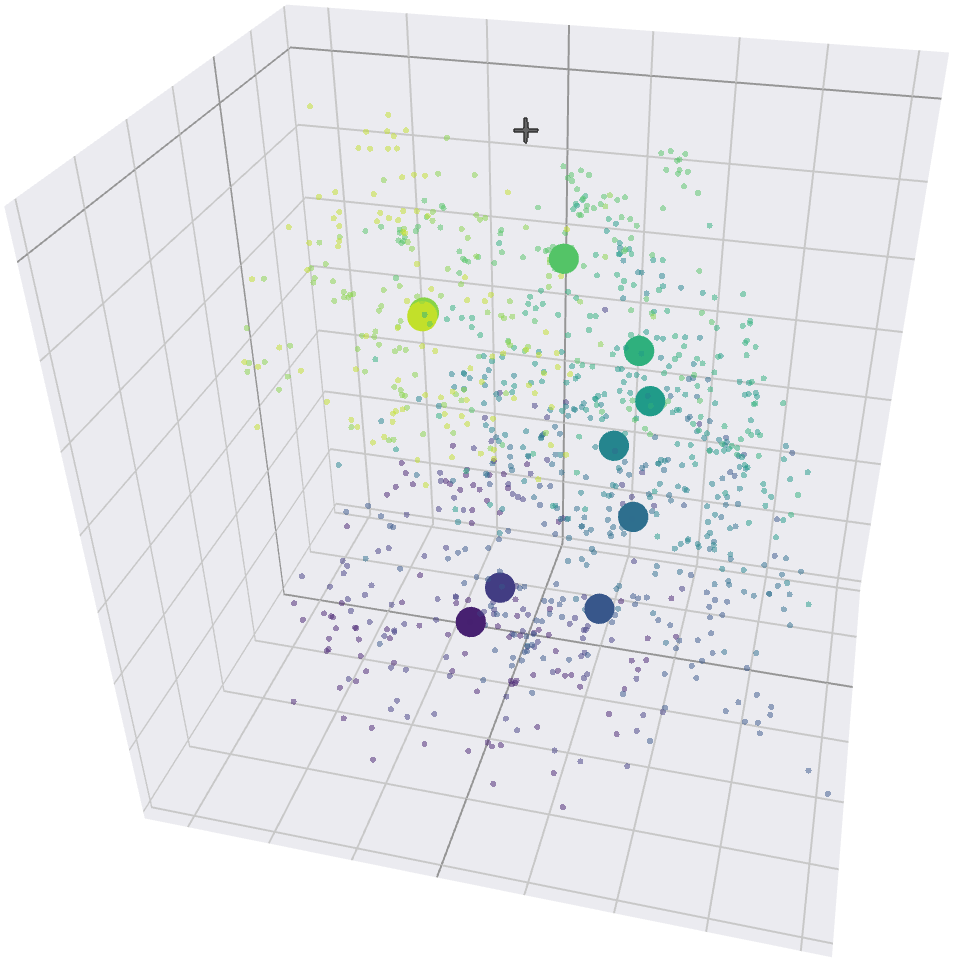}};
  \node[anchor=north west, inner sep=2pt, overlay] at (img.north west) {\small\textbf{(c)}};
\end{tikzpicture}%
\end{minipage}%
\begin{minipage}[c][4.0cm][c]{0.2593\linewidth}%
\centering%
\includegraphics[width=\linewidth,height=4.0cm,keepaspectratio]{paper/legends/legend_time_units.png}%
\end{minipage}%
\par\vspace{1pt}%
{\footnotesize\textit{Time Units}}%
\end{minipage}%
}
\caption{Gemma 2 2B, Layer 12. \textbf{Hours}: (a) Expert 589, rank 1, 24-Hour Ring ($R^2$\,=\,0.681).  \textbf{Temperature}: (b) Expert 789, rank 1, Fahrenheit ($R^2$\,=\,0.663).  \textbf{Time Units}: (c) Expert 1009, rank 2, $\log\_{10}$ Duration ($R^2$\,=\,0.686). Each plot shows the 3-D bottleneck activations of a single SMIXAE expert; small points are individual token activations colored by ground-truth label, and larger points mark per-class means.}
\label{fig:probe_gemma_2_2b_l12}
\end{figure*}

%% file: paper/probe_gemma_2_9b_l20.tex

\begin{figure*}[t]
\centering
\noindent\makebox[\linewidth][c]{%
\begin{minipage}[t]{0.3576\linewidth}%
\vspace{0pt}\centering%
\begin{minipage}[c][4.0cm][c]{0.7407\linewidth}%
\centering%
\begin{tikzpicture}[baseline=(img.base)]
  \node[inner sep=0pt] (img) {\includegraphics[width=\linewidth,height=4.0cm,keepaspectratio]{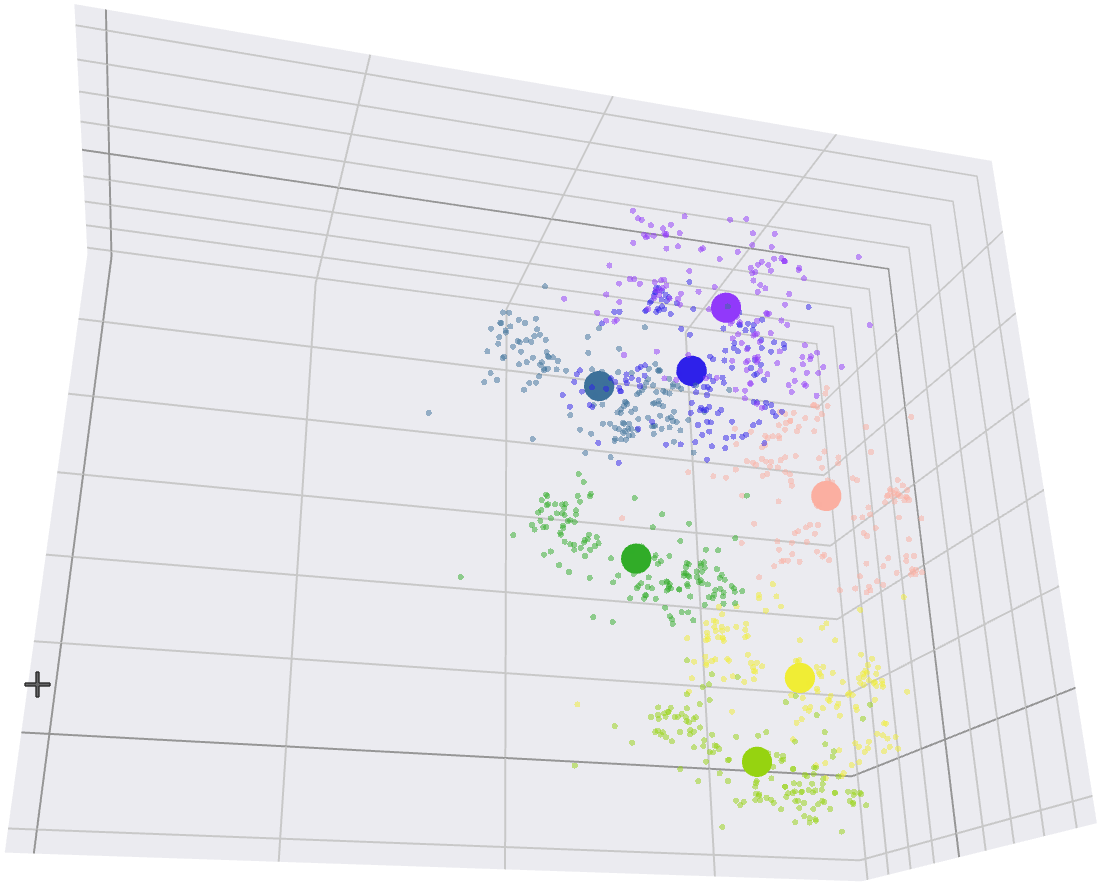}};
  \node[anchor=north west, inner sep=2pt, overlay] at (img.north west) {\small\textbf{(a)}};
\end{tikzpicture}%
\end{minipage}%
\begin{minipage}[c][4.0cm][c]{0.2593\linewidth}%
\centering%
\includegraphics[width=\linewidth,height=4.0cm,keepaspectratio]{paper/legends/legend_weekdays.png}%
\end{minipage}%
\par\vspace{1pt}%
{\footnotesize\textit{Weekdays}}%
\end{minipage}%
\hspace{2.5mm}%
\begin{minipage}[t]{0.6224\linewidth}%
\vspace{0pt}\centering%
\begin{minipage}[c][4.0cm][c]{0.4255\linewidth}%
\centering%
\begin{tikzpicture}[baseline=(img.base)]
  \node[inner sep=0pt] (img) {\includegraphics[width=\linewidth,height=4.0cm,keepaspectratio]{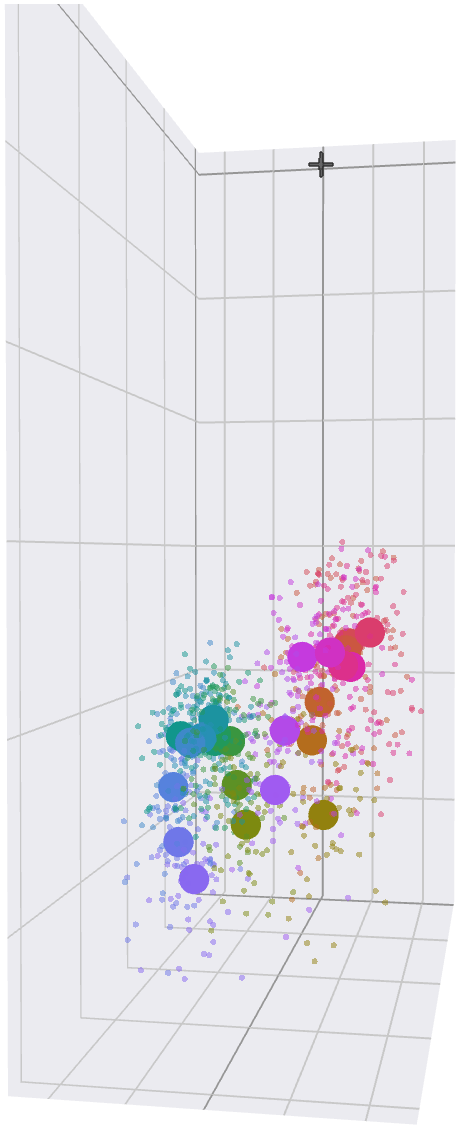}};
  \node[anchor=north west, inner sep=2pt, overlay] at (img.north west) {\small\textbf{(b)}};
\end{tikzpicture}%
\end{minipage}%
\begin{minipage}[c][4.0cm][c]{0.4255\linewidth}%
\centering%
\begin{tikzpicture}[baseline=(img.base)]
  \node[inner sep=0pt] (img) {\includegraphics[width=\linewidth,height=4.0cm,keepaspectratio]{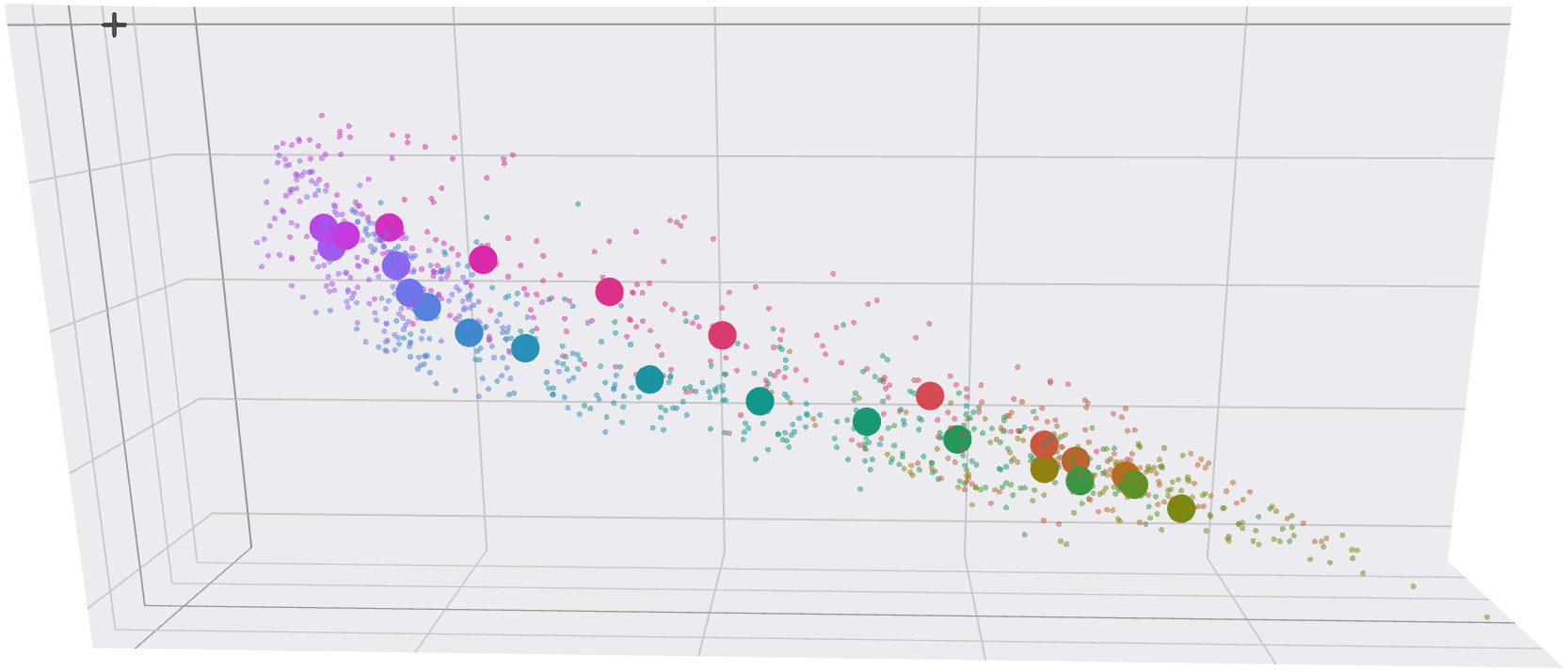}};
  \node[anchor=north west, inner sep=2pt, overlay] at (img.north west) {\small\textbf{(c)}};
\end{tikzpicture}%
\end{minipage}%
\begin{minipage}[c][4.0cm][c]{0.1489\linewidth}%
\centering%
\includegraphics[width=\linewidth,height=4.0cm,keepaspectratio]{paper/legends/legend_hours.png}%
\end{minipage}%
\par\vspace{1pt}%
{\footnotesize\textit{Hours}}%
\end{minipage}%
}
\vspace{5pt}

\noindent\makebox[\linewidth][c]{%
\begin{minipage}[t]{0.3949\linewidth}%
\vspace{0pt}\centering%
\begin{minipage}[c][4.0cm][c]{0.7407\linewidth}%
\centering%
\begin{tikzpicture}[baseline=(img.base)]
  \node[inner sep=0pt] (img) {\includegraphics[width=\linewidth,height=4.0cm,keepaspectratio]{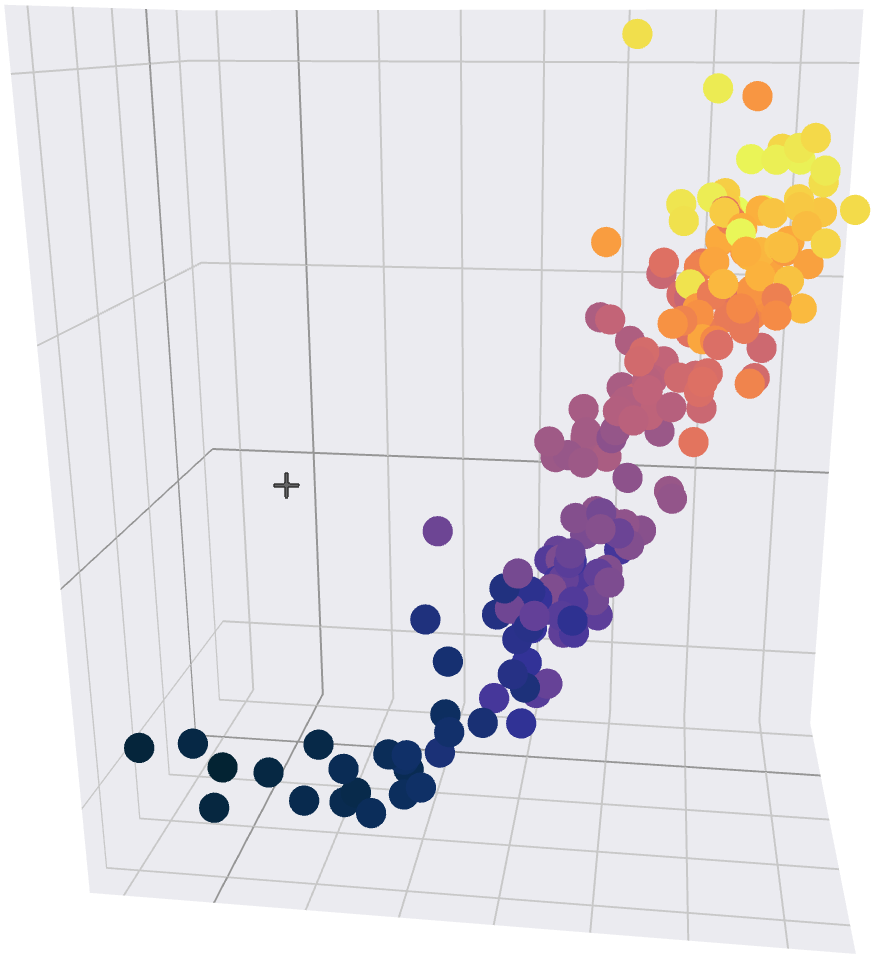}};
  \node[anchor=north west, inner sep=2pt, overlay] at (img.north west) {\small\textbf{(d)}};
\end{tikzpicture}%
\end{minipage}%
\begin{minipage}[c][4.0cm][c]{0.2593\linewidth}%
\centering%
\includegraphics[width=\linewidth,height=4.0cm,keepaspectratio]{paper/legends/legend_temperatures.png}%
\end{minipage}%
\par\vspace{1pt}%
{\footnotesize\textit{Temperature}}%
\end{minipage}%
\hspace{2.5mm}%
\begin{minipage}[t]{0.3949\linewidth}%
\vspace{0pt}\centering%
\begin{minipage}[c][4.0cm][c]{0.7407\linewidth}%
\centering%
\begin{tikzpicture}[baseline=(img.base)]
  \node[inner sep=0pt] (img) {\includegraphics[width=\linewidth,height=4.0cm,keepaspectratio]{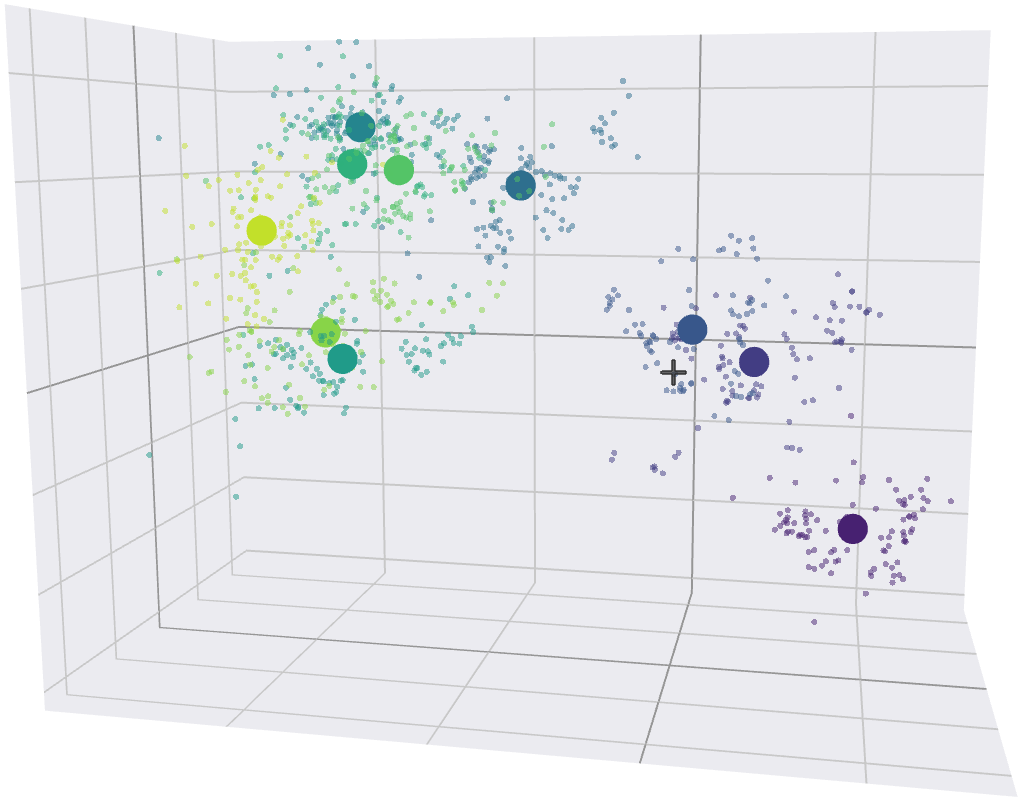}};
  \node[anchor=north west, inner sep=2pt, overlay] at (img.north west) {\small\textbf{(e)}};
\end{tikzpicture}%
\end{minipage}%
\begin{minipage}[c][4.0cm][c]{0.2593\linewidth}%
\centering%
\includegraphics[width=\linewidth,height=4.0cm,keepaspectratio]{paper/legends/legend_time_units.png}%
\end{minipage}%
\par\vspace{1pt}%
{\footnotesize\textit{Time Units}}%
\end{minipage}%
}
\caption{Gemma 2 9B, Layer 20. \textbf{Weekdays}: (a) Expert 391, rank 1, 7-Day Ring ($R^2$\,=\,0.724).  \textbf{Hours}: (b) Expert 1041, rank 2, AM vs PM (Accuracy\,=\,0.997). (c) Expert 1078, rank 1, 24-Hour Ring ($R^2$\,=\,0.590).  \textbf{Temperature}: (d) Expert 1716, rank 1, Fahrenheit ($R^2$\,=\,0.875).  \textbf{Time Units}: (e) Expert 923, rank 1, $\log\_{10}$ Duration ($R^2$\,=\,0.856). Each plot shows the 3-D bottleneck activations of a single SMIXAE expert; small points are individual token activations colored by ground-truth label, and larger points mark per-class means.}
\label{fig:probe_gemma_2_9b_l20}
\end{figure*}

%% file: paper/random_gemma_2_2b_l12.tex

\begin{figure*}[t]
\centering
\noindent\makebox[\linewidth][c]{%
\begin{minipage}[t]{\linewidth}%
\vspace{0pt}\centering%
\begin{minipage}[c][4.0cm][c]{0.3267\linewidth}%
\centering%
\begin{tikzpicture}[baseline=(img.base)]
  \node[inner sep=0pt] (img) {\includegraphics[width=\linewidth,height=4.0cm,keepaspectratio]{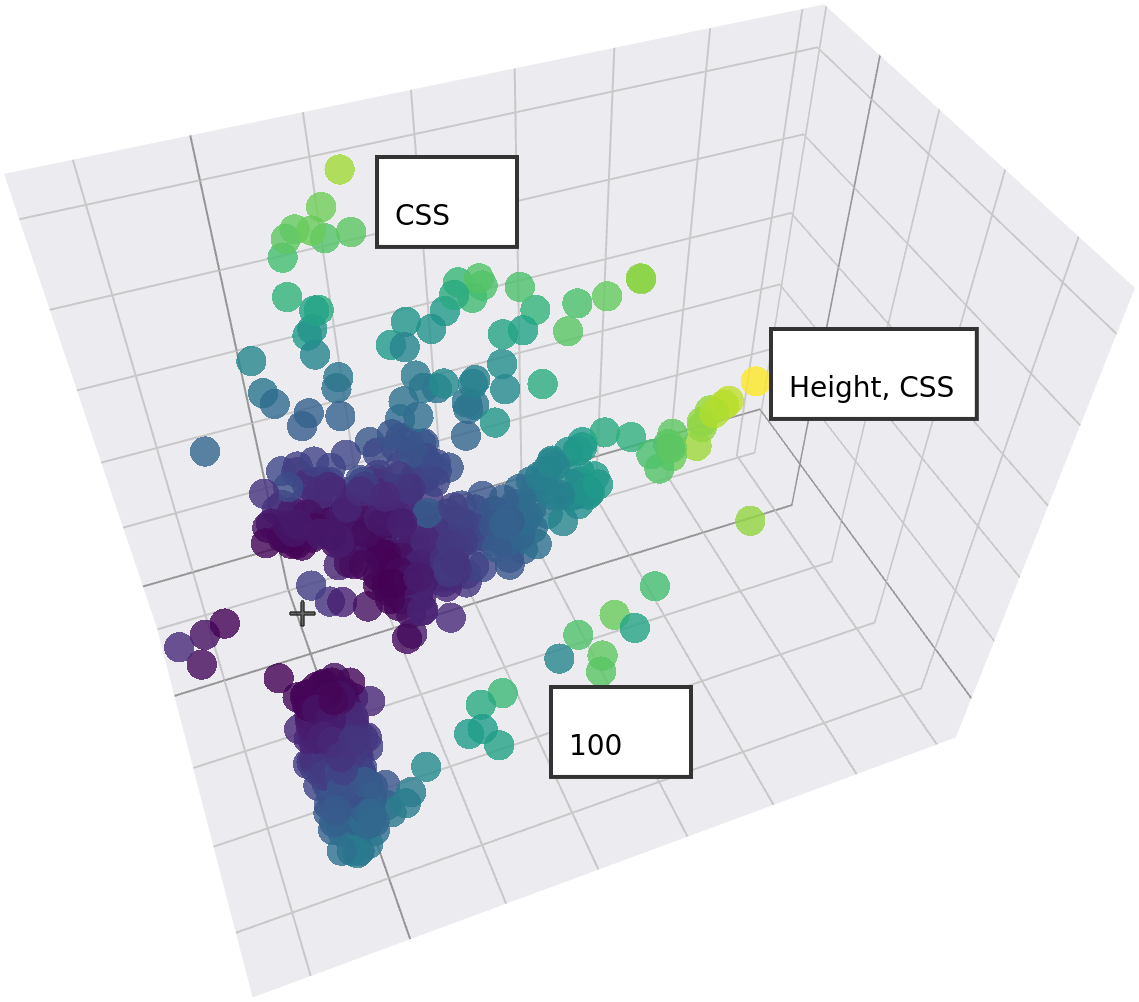}};
  \node[anchor=north west, inner sep=2pt, overlay] at (img.north west) {\small\textbf{(a)}};
\end{tikzpicture}%
\end{minipage}%
\begin{minipage}[c][4.0cm][c]{0.3267\linewidth}%
\centering%
\begin{tikzpicture}[baseline=(img.base)]
  \node[inner sep=0pt] (img) {\includegraphics[width=\linewidth,height=4.0cm,keepaspectratio]{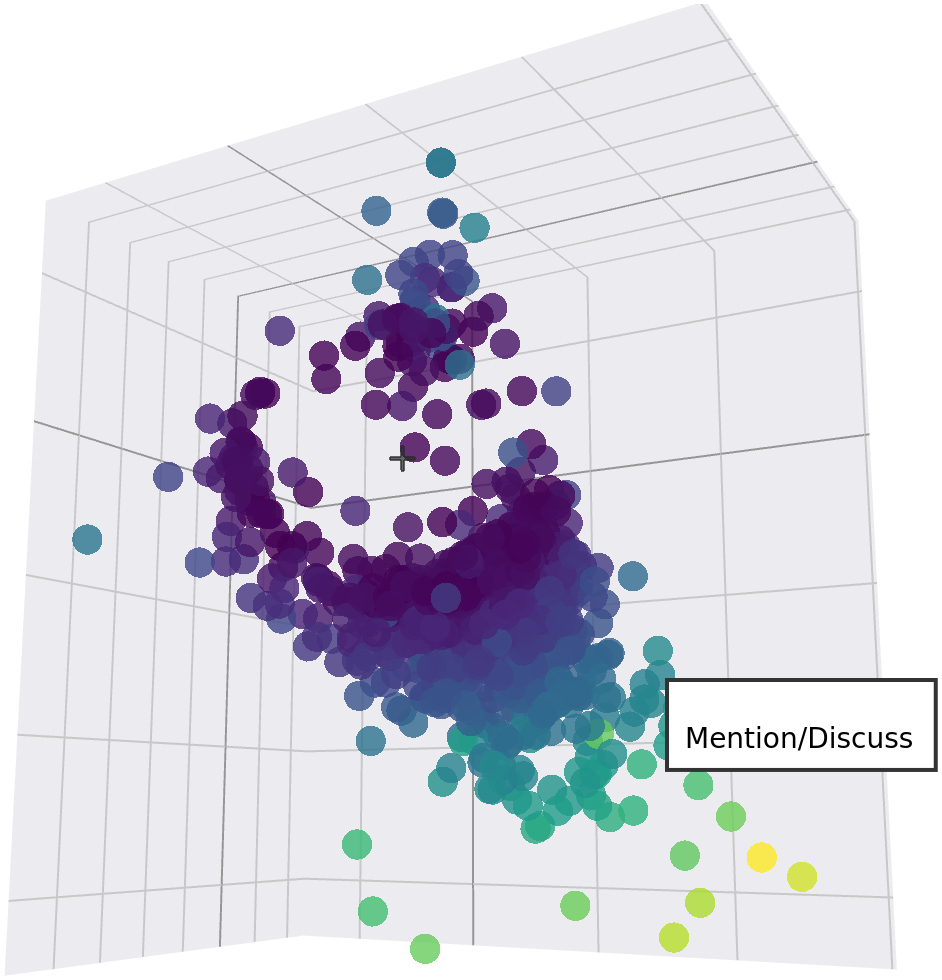}};
  \node[anchor=north west, inner sep=2pt, overlay] at (img.north west) {\small\textbf{(b)}};
\end{tikzpicture}%
\end{minipage}%
\begin{minipage}[c][4.0cm][c]{0.3267\linewidth}%
\centering%
\begin{tikzpicture}[baseline=(img.base)]
  \node[inner sep=0pt] (img) {\includegraphics[width=\linewidth,height=4.0cm,keepaspectratio]{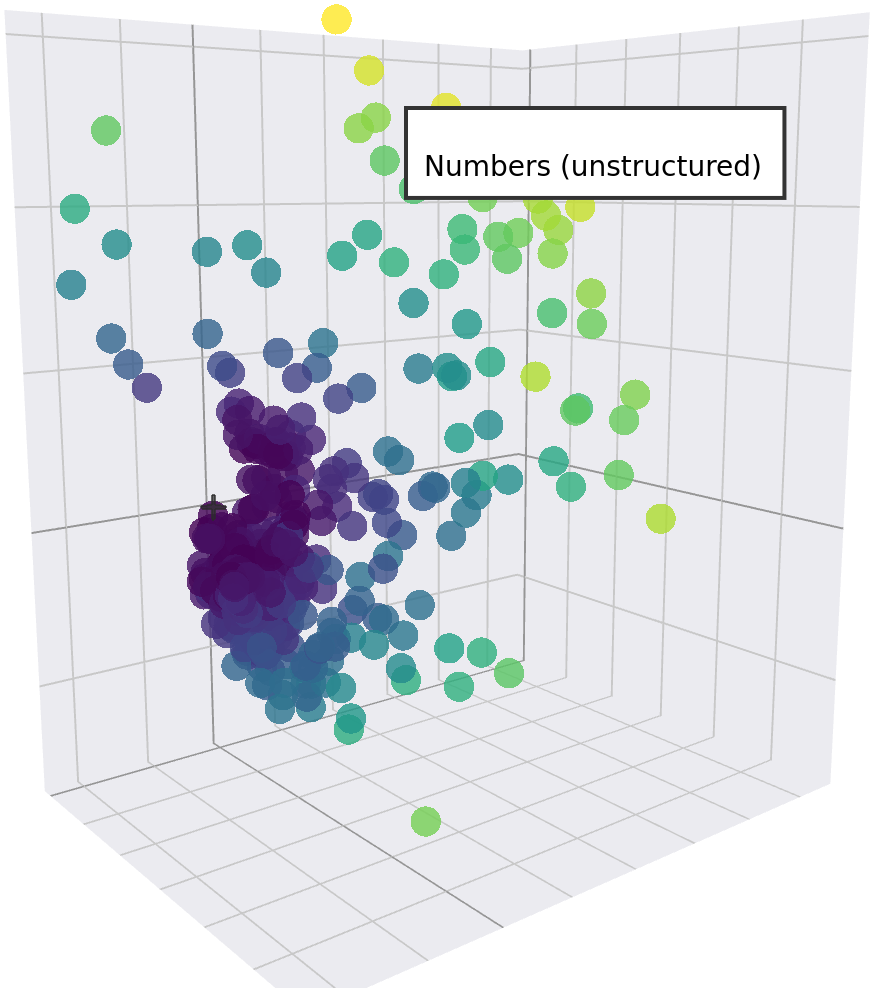}};
  \node[anchor=north west, inner sep=2pt, overlay] at (img.north west) {\small\textbf{(c)}};
\end{tikzpicture}%
\end{minipage}%
\par\vspace{3pt}%
\begin{minipage}[c][4.0cm][c]{0.3267\linewidth}%
\centering%
\begin{tikzpicture}[baseline=(img.base)]
  \node[inner sep=0pt] (img) {\includegraphics[width=\linewidth,height=4.0cm,keepaspectratio]{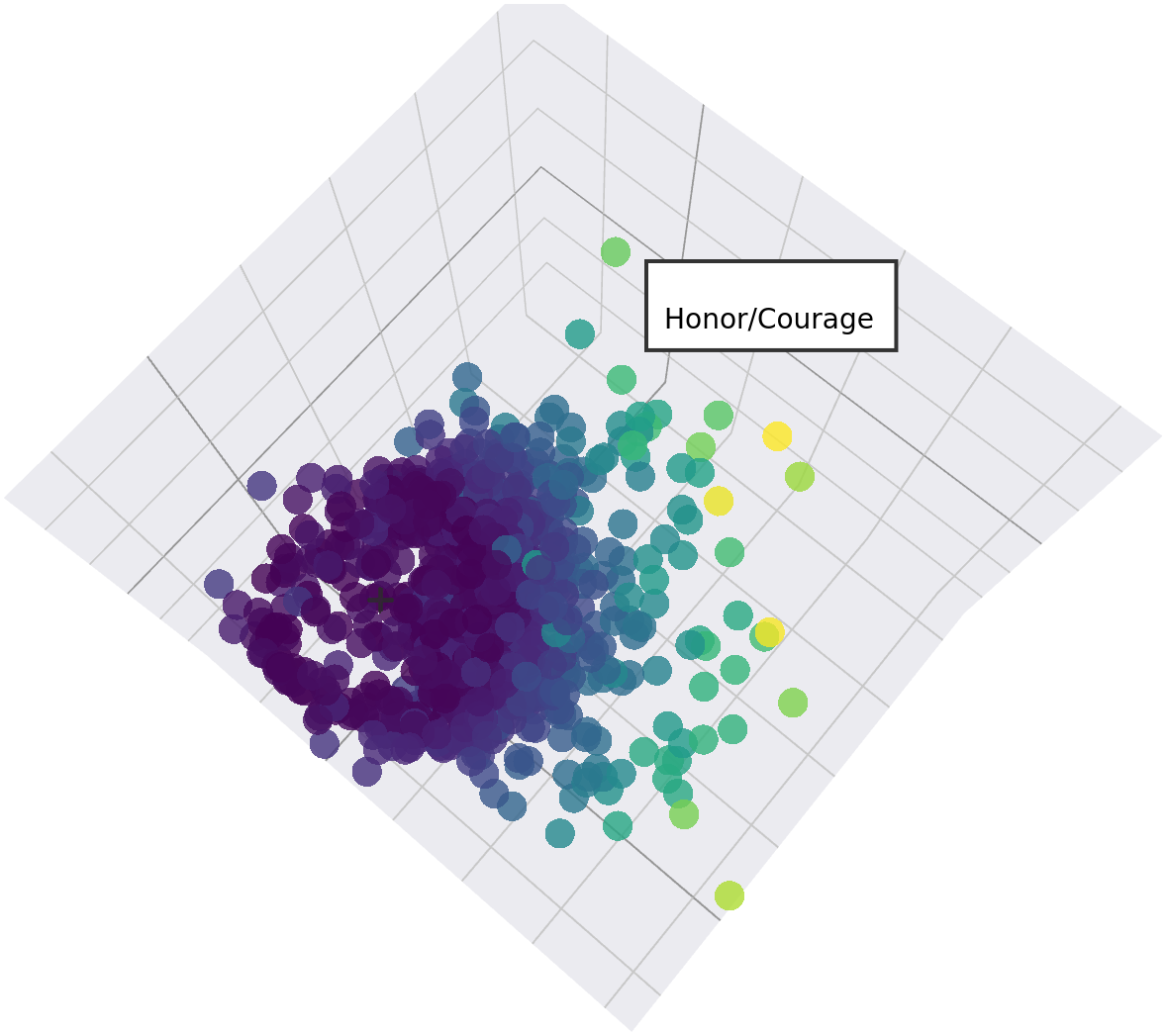}};
  \node[anchor=north west, inner sep=2pt, overlay] at (img.north west) {\small\textbf{(d)}};
\end{tikzpicture}%
\end{minipage}%
\begin{minipage}[c][4.0cm][c]{0.3267\linewidth}%
\centering%
\begin{tikzpicture}[baseline=(img.base)]
  \node[inner sep=0pt] (img) {\includegraphics[width=\linewidth,height=4.0cm,keepaspectratio]{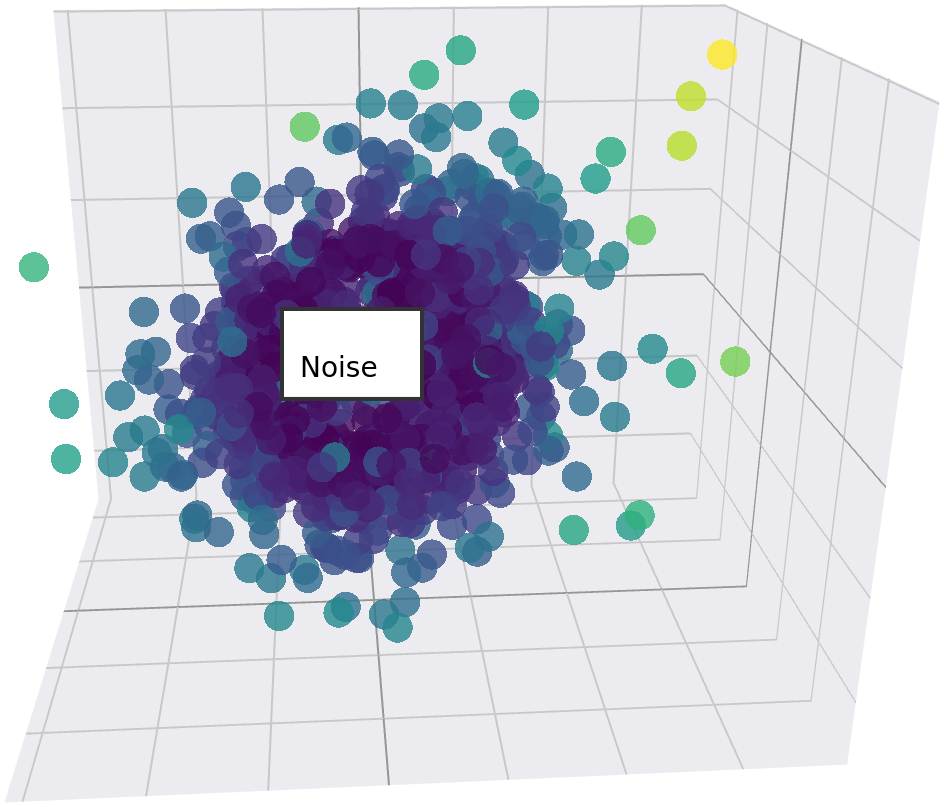}};
  \node[anchor=north west, inner sep=2pt, overlay] at (img.north west) {\small\textbf{(e)}};
\end{tikzpicture}%
\end{minipage}%
\begin{minipage}[c][4.0cm][c]{0.3267\linewidth}%
\centering%
\begin{tikzpicture}[baseline=(img.base)]
  \node[inner sep=0pt] (img) {\includegraphics[width=\linewidth,height=4.0cm,keepaspectratio]{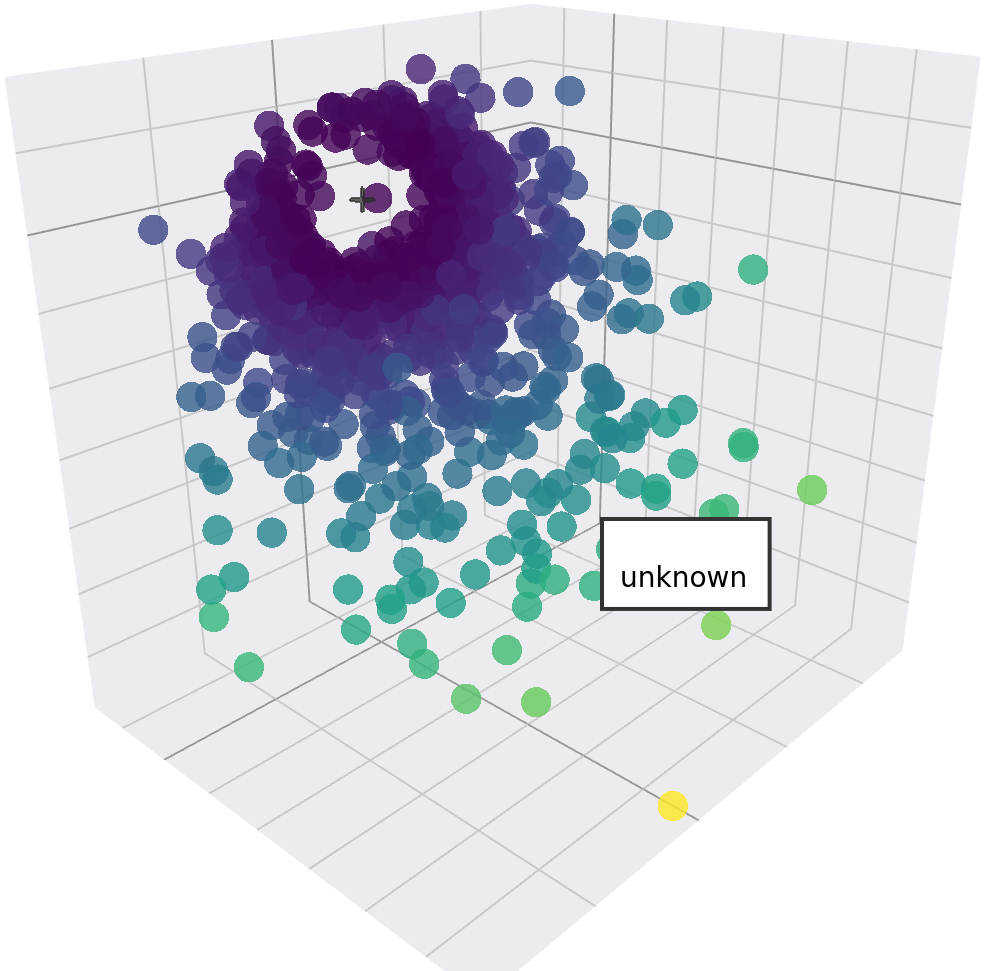}};
  \node[anchor=north west, inner sep=2pt, overlay] at (img.north west) {\small\textbf{(f)}};
\end{tikzpicture}%
\end{minipage}%
\par\vspace{3pt}%
\begin{minipage}[c][4.0cm][c]{0.3267\linewidth}%
\centering%
\begin{tikzpicture}[baseline=(img.base)]
  \node[inner sep=0pt] (img) {\includegraphics[width=\linewidth,height=4.0cm,keepaspectratio]{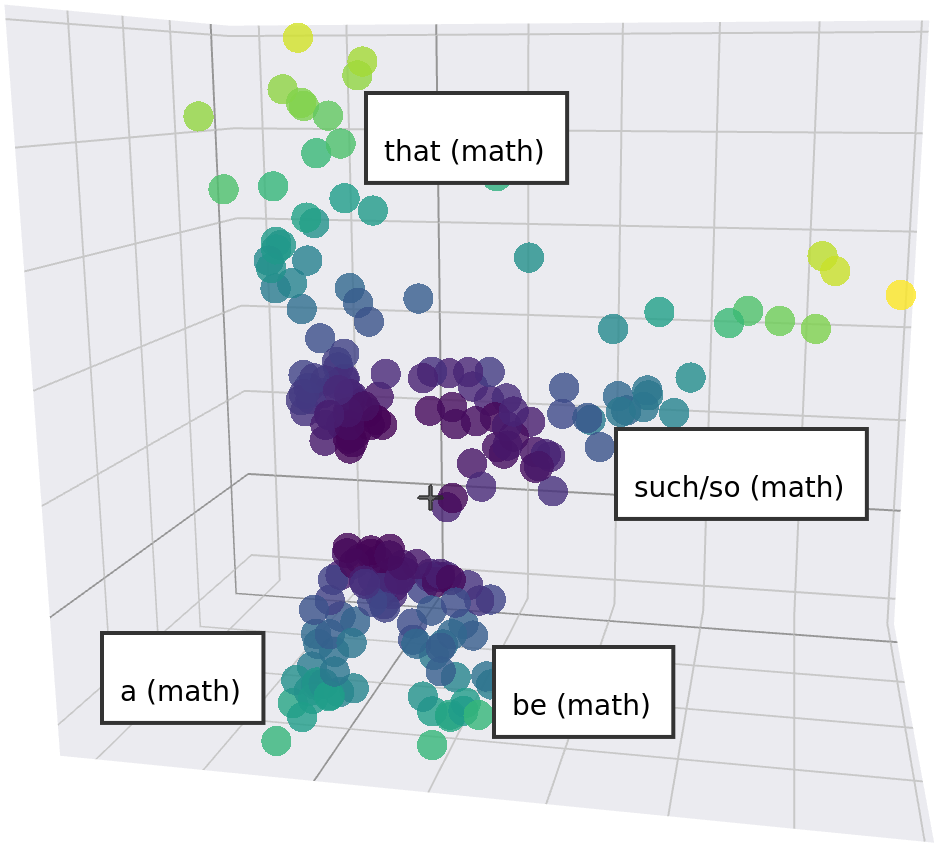}};
  \node[anchor=north west, inner sep=2pt, overlay] at (img.north west) {\small\textbf{(g)}};
\end{tikzpicture}%
\end{minipage}%
\begin{minipage}[c][4.0cm][c]{0.3267\linewidth}%
\centering%
\begin{tikzpicture}[baseline=(img.base)]
  \node[inner sep=0pt] (img) {\includegraphics[width=\linewidth,height=4.0cm,keepaspectratio]{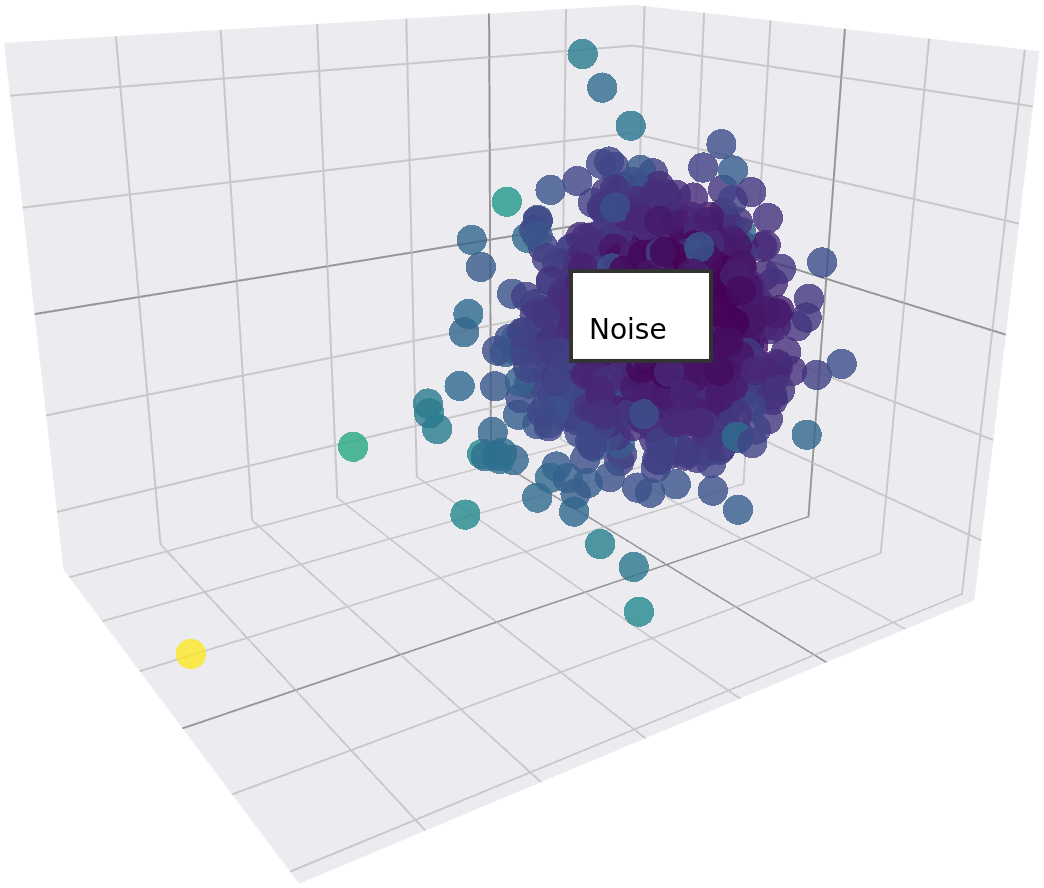}};
  \node[anchor=north west, inner sep=2pt, overlay] at (img.north west) {\small\textbf{(h)}};
\end{tikzpicture}%
\end{minipage}%
\begin{minipage}[c][4.0cm][c]{0.3267\linewidth}%
\centering%
\begin{tikzpicture}[baseline=(img.base)]
  \node[inner sep=0pt] (img) {\includegraphics[width=\linewidth,height=4.0cm,keepaspectratio]{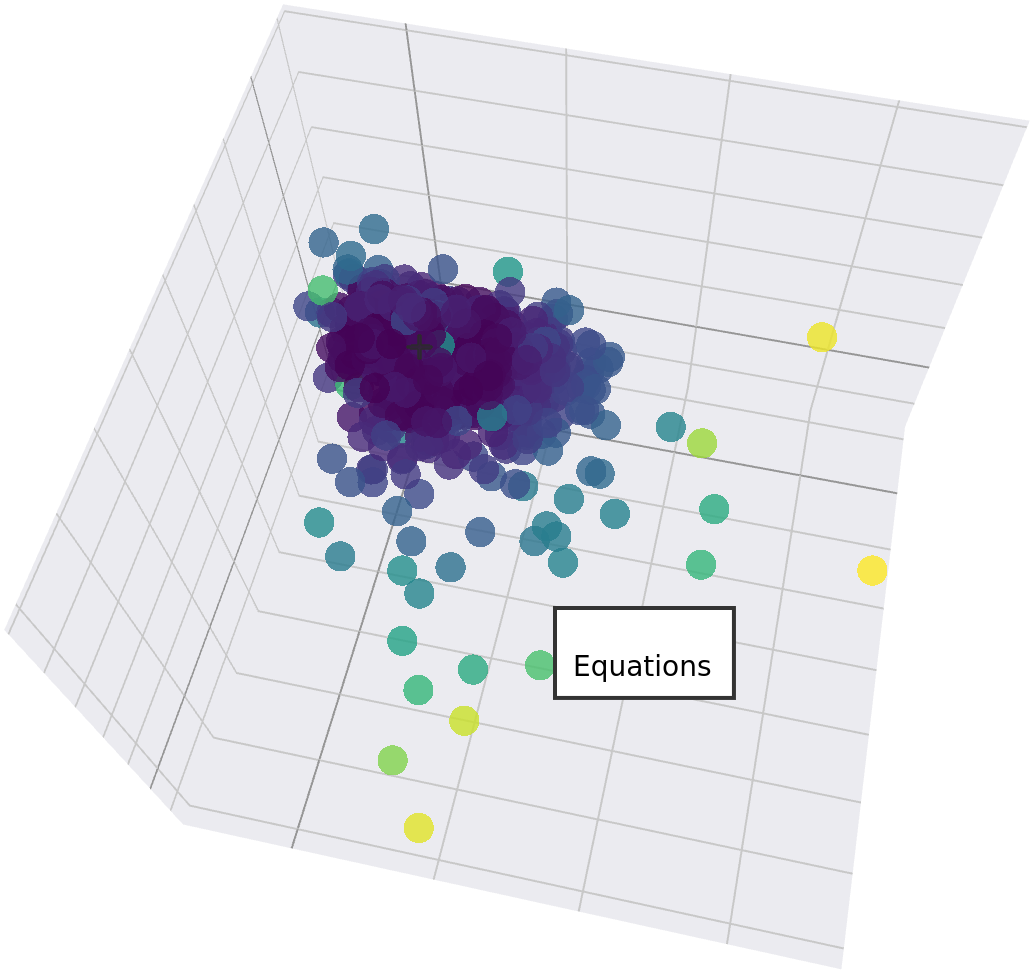}};
  \node[anchor=north west, inner sep=2pt, overlay] at (img.north west) {\small\textbf{(i)}};
\end{tikzpicture}%
\end{minipage}%
\par\vspace{3pt}%
\begin{minipage}[c][4.0cm][c]{0.3267\linewidth}%
\centering%
\begin{tikzpicture}[baseline=(img.base)]
  \node[inner sep=0pt] (img) {\includegraphics[width=\linewidth,height=4.0cm,keepaspectratio]{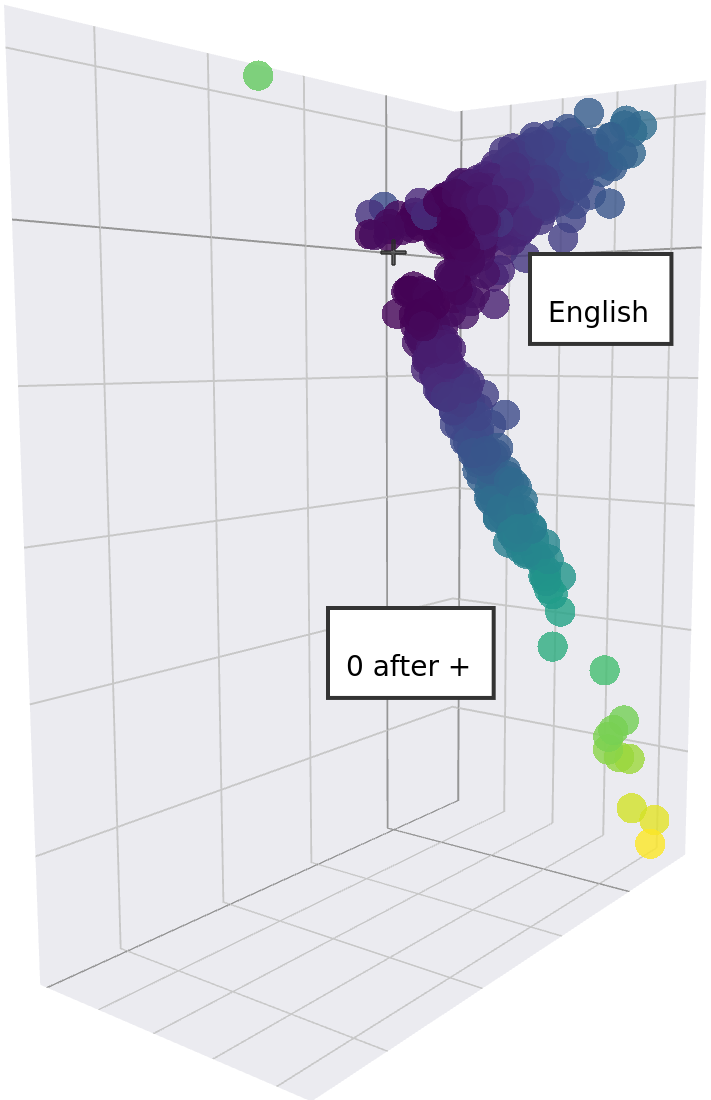}};
  \node[anchor=north west, inner sep=2pt, overlay] at (img.north west) {\small\textbf{(j)}};
\end{tikzpicture}%
\end{minipage}%
\begin{minipage}[c][4.0cm][c]{0.3267\linewidth}%
\centering%
\vspace{0pt}%
\end{minipage}%
\begin{minipage}[c][4.0cm][c]{0.3267\linewidth}%
\centering%
\vspace{0pt}%
\end{minipage}%
\par\vspace{1pt}%
{\footnotesize\textit{Random Experts (Unlabeled)}}%
\end{minipage}%
}
\caption{Gemma 2 2B, Layer 12. \textbf{Random Experts}: (a) Expert 1359. (b) Expert 1568. (c) Expert 1578. (d) Expert 213. (e) Expert 234. (f) Expert 293. (g) Expert 477. (h) Expert 51. (i) Expert 524. (j) Expert 587. Each plot shows the 3-D bottleneck activations of a single SMIXAE expert; points are individual token activations colored by distance from the origin.}
\label{fig:random_gemma_2_2b_l12}
\end{figure*}

%% file: paper/random_gemma_2_9b_l11.tex

\begin{figure*}[t]
\centering
\noindent\makebox[\linewidth][c]{%
\begin{minipage}[t]{\linewidth}%
\vspace{0pt}\centering%
\begin{minipage}[c][4.0cm][c]{0.3267\linewidth}%
\centering%
\begin{tikzpicture}[baseline=(img.base)]
  \node[inner sep=0pt] (img) {\includegraphics[width=\linewidth,height=4.0cm,keepaspectratio]{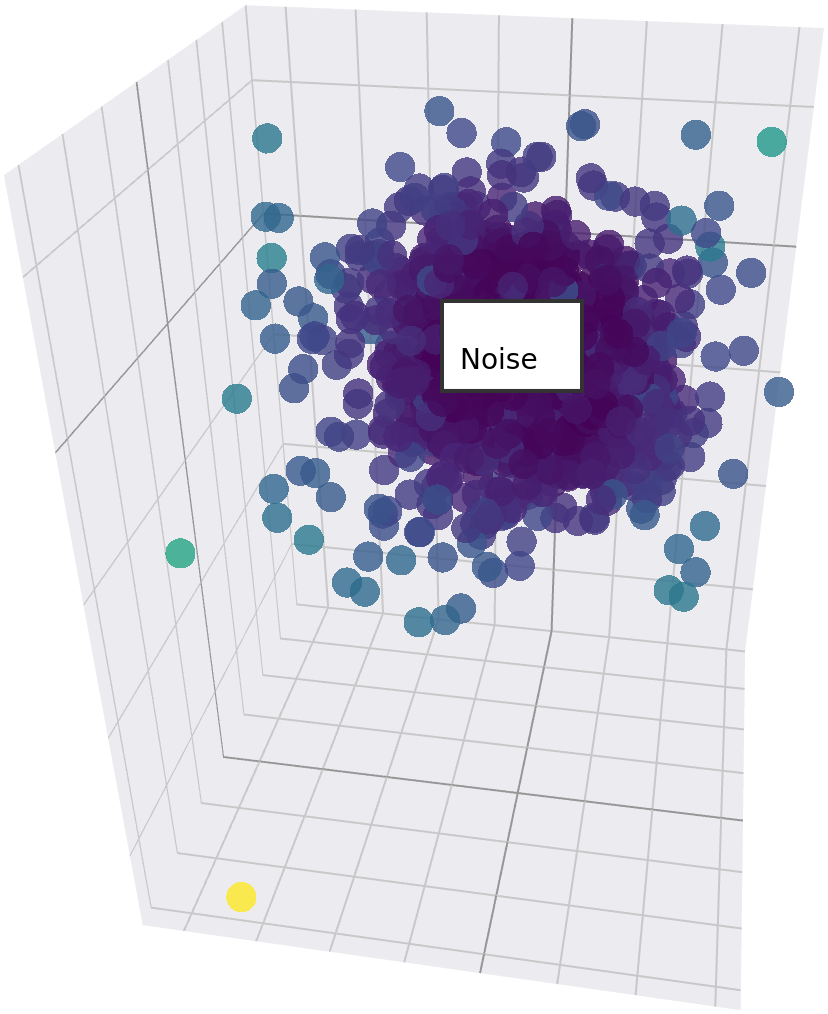}};
  \node[anchor=north west, inner sep=2pt, overlay] at (img.north west) {\small\textbf{(a)}};
\end{tikzpicture}%
\end{minipage}%
\begin{minipage}[c][4.0cm][c]{0.3267\linewidth}%
\centering%
\begin{tikzpicture}[baseline=(img.base)]
  \node[inner sep=0pt] (img) {\includegraphics[width=\linewidth,height=4.0cm,keepaspectratio]{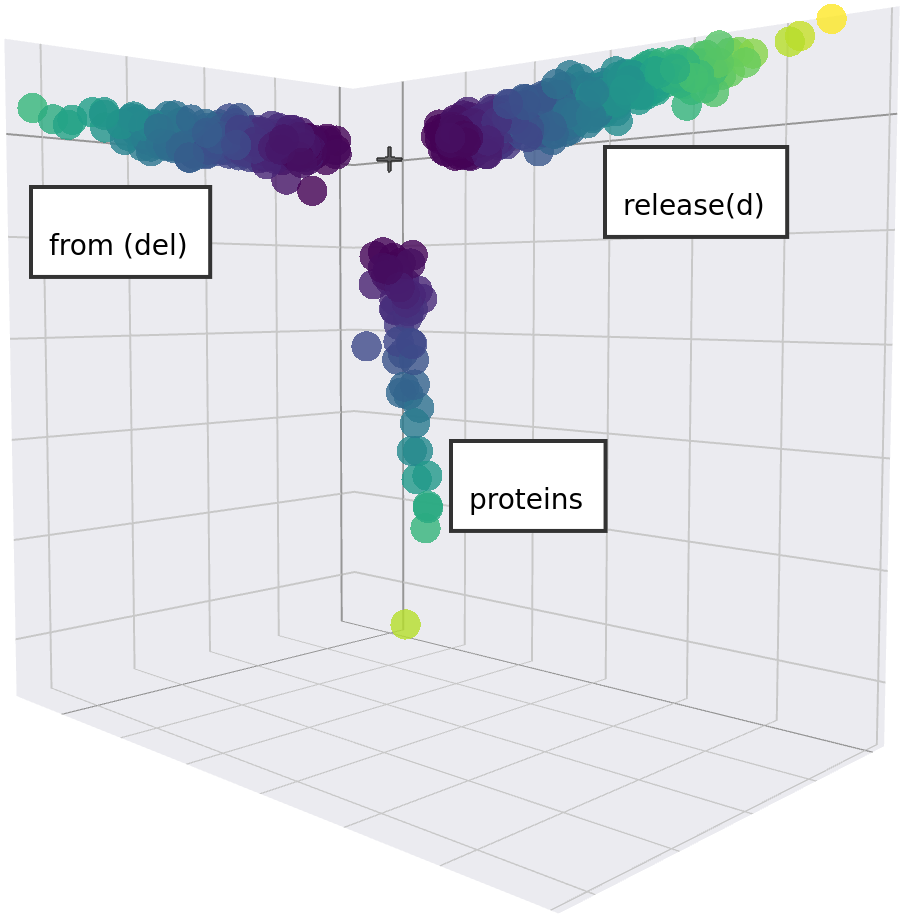}};
  \node[anchor=north west, inner sep=2pt, overlay] at (img.north west) {\small\textbf{(b)}};
\end{tikzpicture}%
\end{minipage}%
\begin{minipage}[c][4.0cm][c]{0.3267\linewidth}%
\centering%
\begin{tikzpicture}[baseline=(img.base)]
  \node[inner sep=0pt] (img) {\includegraphics[width=\linewidth,height=4.0cm,keepaspectratio]{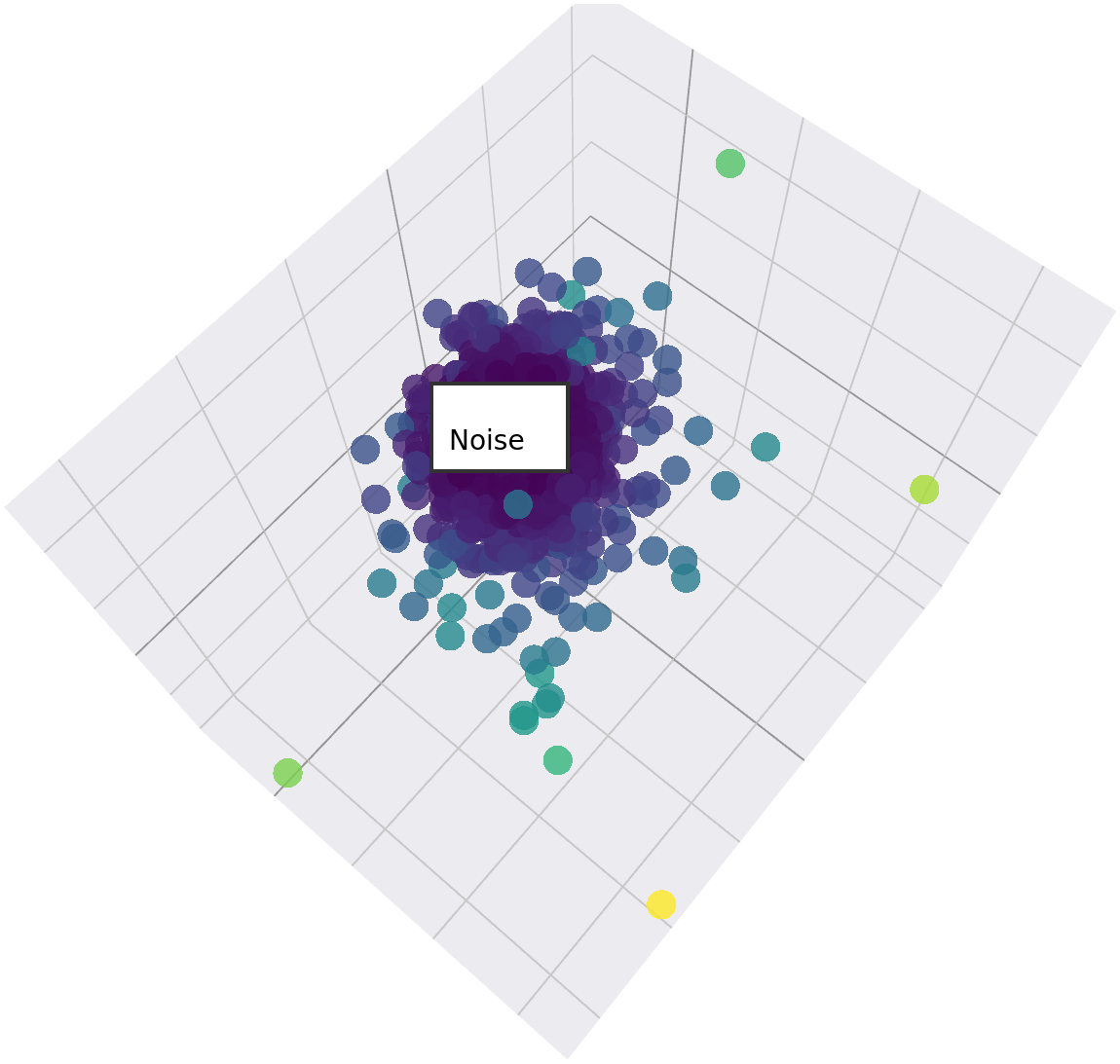}};
  \node[anchor=north west, inner sep=2pt, overlay] at (img.north west) {\small\textbf{(c)}};
\end{tikzpicture}%
\end{minipage}%
\par\vspace{3pt}%
\begin{minipage}[c][4.0cm][c]{0.3267\linewidth}%
\centering%
\begin{tikzpicture}[baseline=(img.base)]
  \node[inner sep=0pt] (img) {\includegraphics[width=\linewidth,height=4.0cm,keepaspectratio]{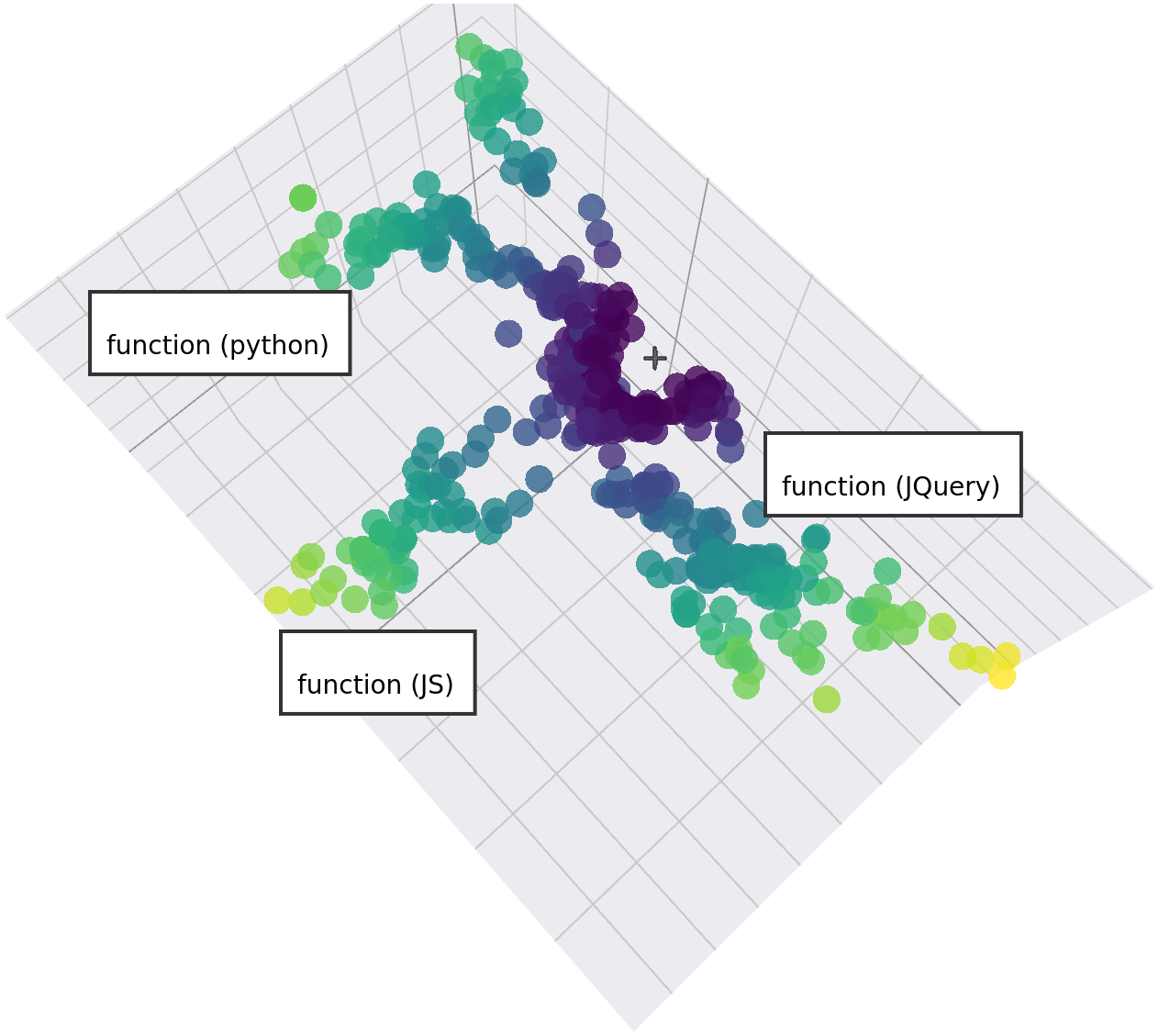}};
  \node[anchor=north west, inner sep=2pt, overlay] at (img.north west) {\small\textbf{(d)}};
\end{tikzpicture}%
\end{minipage}%
\begin{minipage}[c][4.0cm][c]{0.3267\linewidth}%
\centering%
\begin{tikzpicture}[baseline=(img.base)]
  \node[inner sep=0pt] (img) {\includegraphics[width=\linewidth,height=4.0cm,keepaspectratio]{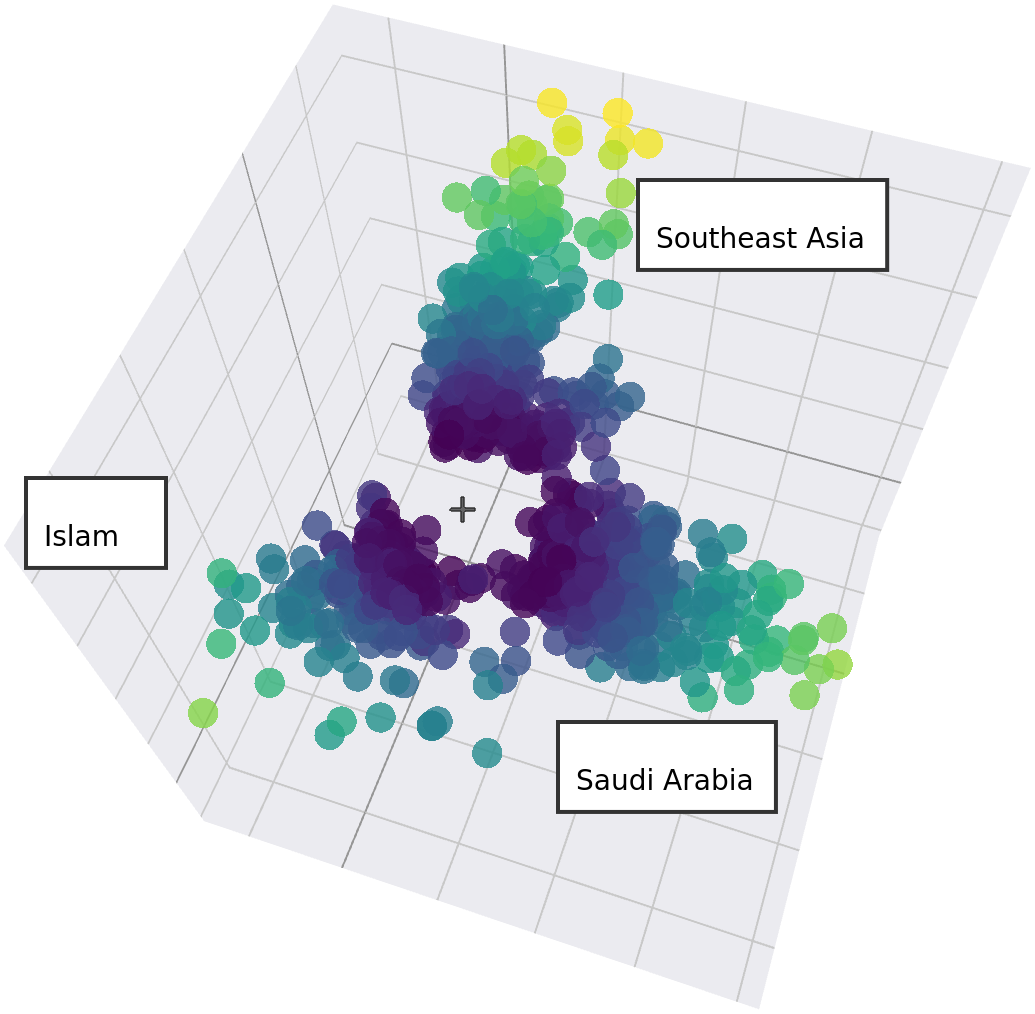}};
  \node[anchor=north west, inner sep=2pt, overlay] at (img.north west) {\small\textbf{(e)}};
\end{tikzpicture}%
\end{minipage}%
\begin{minipage}[c][4.0cm][c]{0.3267\linewidth}%
\centering%
\begin{tikzpicture}[baseline=(img.base)]
  \node[inner sep=0pt] (img) {\includegraphics[width=\linewidth,height=4.0cm,keepaspectratio]{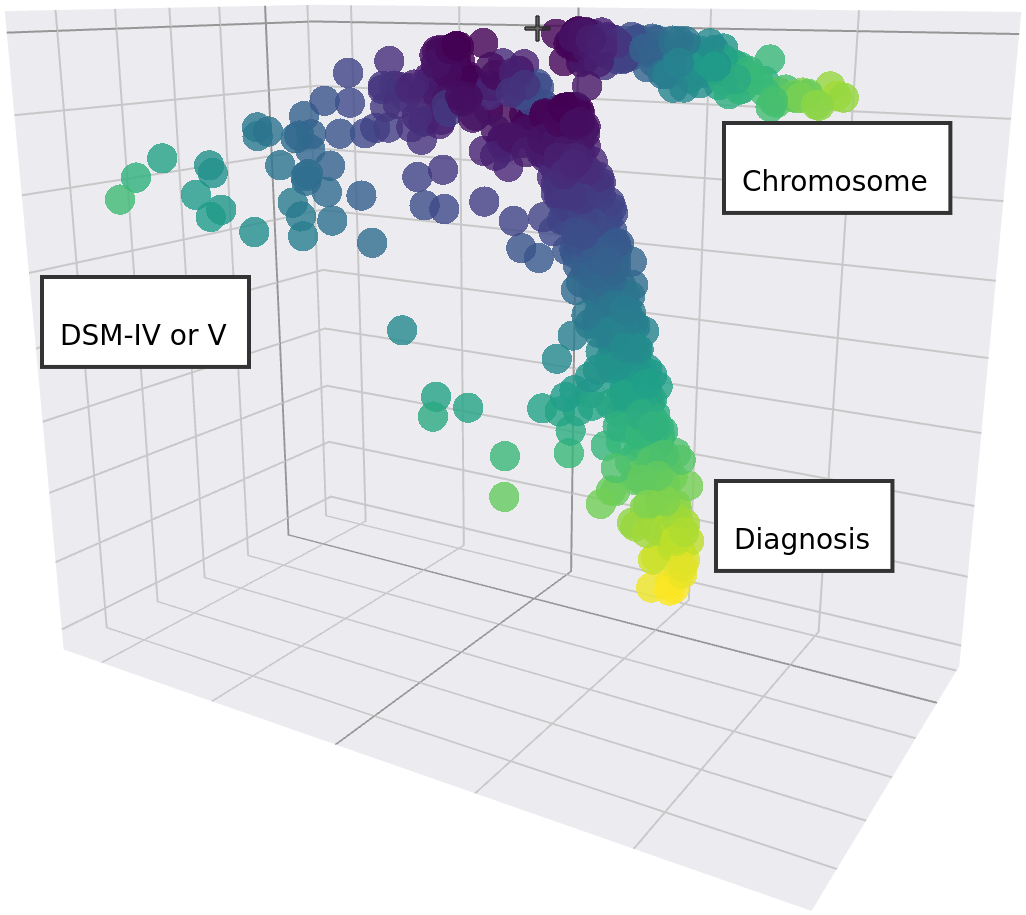}};
  \node[anchor=north west, inner sep=2pt, overlay] at (img.north west) {\small\textbf{(f)}};
\end{tikzpicture}%
\end{minipage}%
\par\vspace{3pt}%
\begin{minipage}[c][4.0cm][c]{0.3267\linewidth}%
\centering%
\begin{tikzpicture}[baseline=(img.base)]
  \node[inner sep=0pt] (img) {\includegraphics[width=\linewidth,height=4.0cm,keepaspectratio]{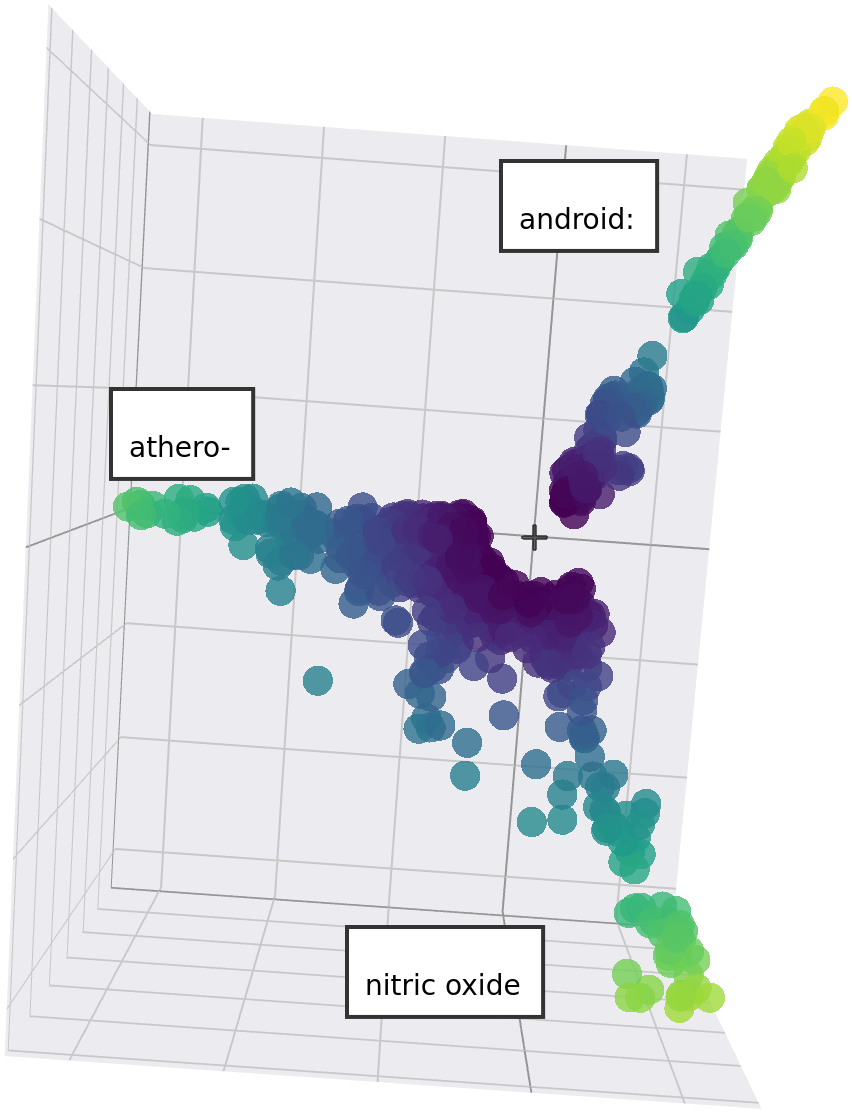}};
  \node[anchor=north west, inner sep=2pt, overlay] at (img.north west) {\small\textbf{(g)}};
\end{tikzpicture}%
\end{minipage}%
\begin{minipage}[c][4.0cm][c]{0.3267\linewidth}%
\centering%
\begin{tikzpicture}[baseline=(img.base)]
  \node[inner sep=0pt] (img) {\includegraphics[width=\linewidth,height=4.0cm,keepaspectratio]{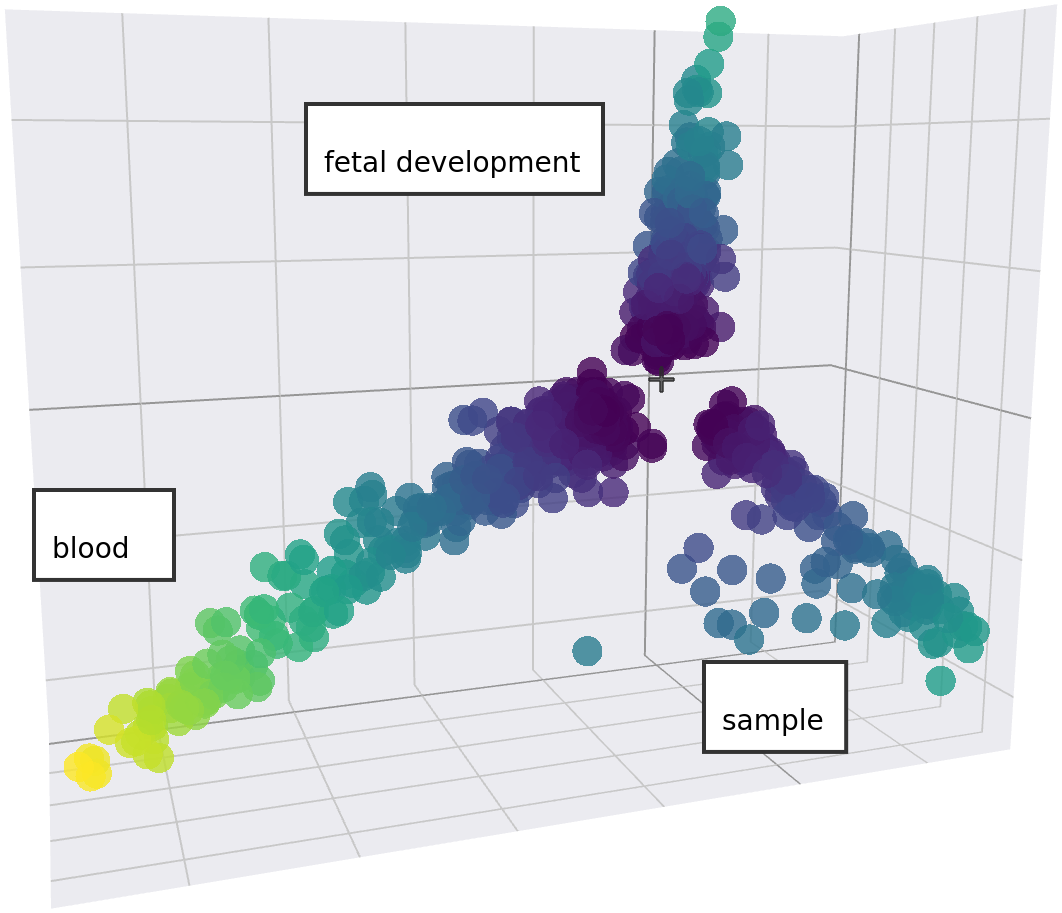}};
  \node[anchor=north west, inner sep=2pt, overlay] at (img.north west) {\small\textbf{(h)}};
\end{tikzpicture}%
\end{minipage}%
\begin{minipage}[c][4.0cm][c]{0.3267\linewidth}%
\centering%
\begin{tikzpicture}[baseline=(img.base)]
  \node[inner sep=0pt] (img) {\includegraphics[width=\linewidth,height=4.0cm,keepaspectratio]{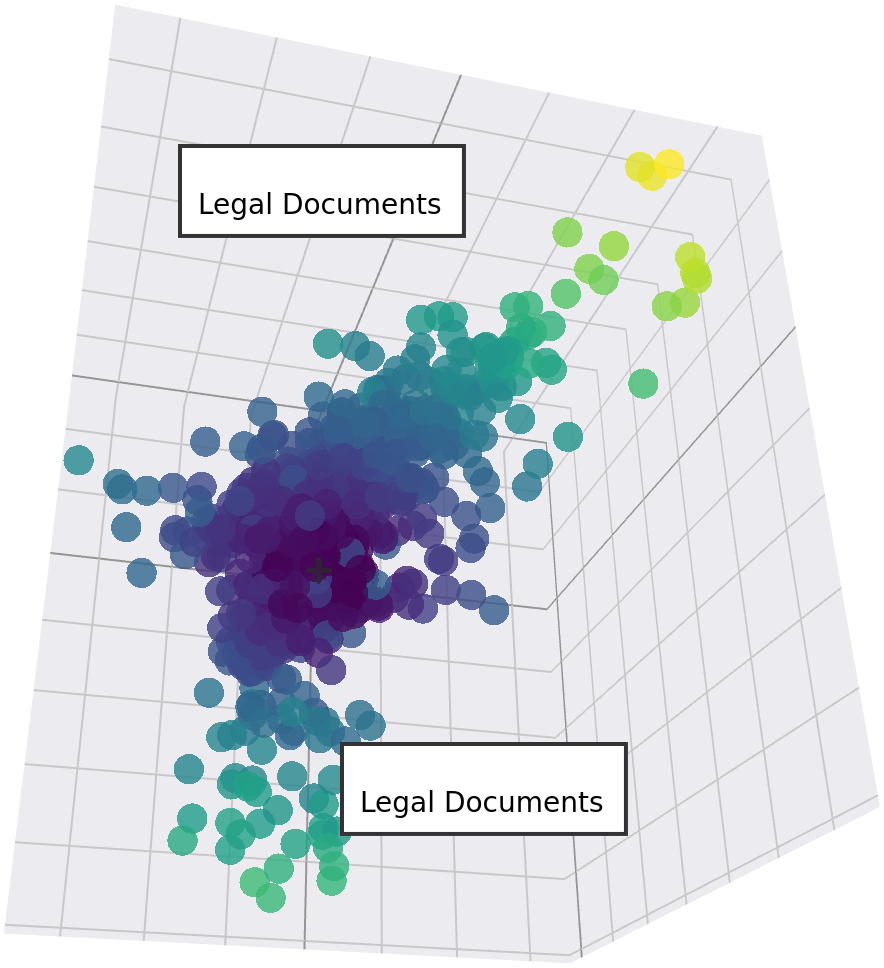}};
  \node[anchor=north west, inner sep=2pt, overlay] at (img.north west) {\small\textbf{(i)}};
\end{tikzpicture}%
\end{minipage}%
\par\vspace{3pt}%
\begin{minipage}[c][4.0cm][c]{0.3267\linewidth}%
\centering%
\begin{tikzpicture}[baseline=(img.base)]
  \node[inner sep=0pt] (img) {\includegraphics[width=\linewidth,height=4.0cm,keepaspectratio]{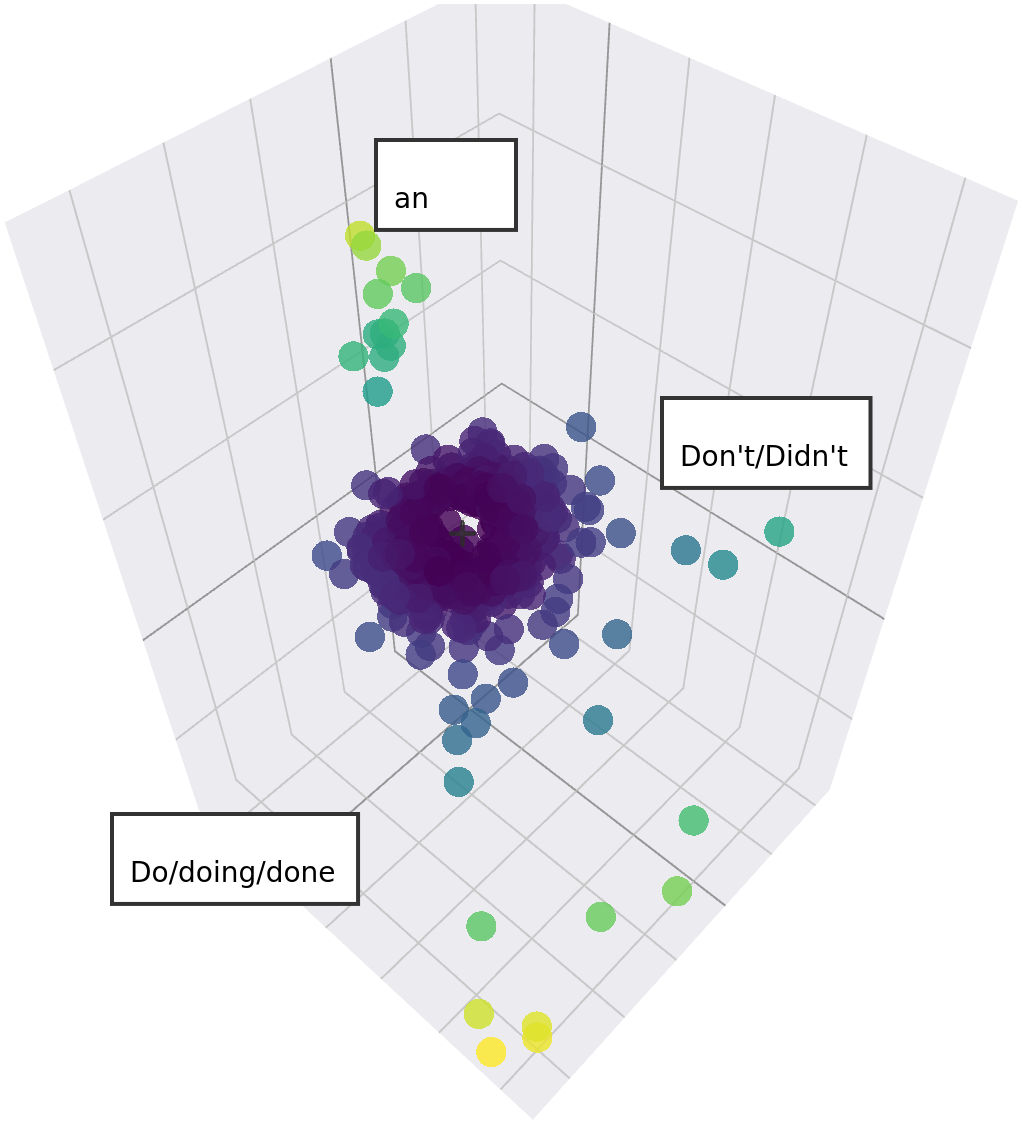}};
  \node[anchor=north west, inner sep=2pt, overlay] at (img.north west) {\small\textbf{(j)}};
\end{tikzpicture}%
\end{minipage}%
\begin{minipage}[c][4.0cm][c]{0.3267\linewidth}%
\centering%
\vspace{0pt}%
\end{minipage}%
\begin{minipage}[c][4.0cm][c]{0.3267\linewidth}%
\centering%
\vspace{0pt}%
\end{minipage}%
\par\vspace{1pt}%
{\footnotesize\textit{Random Experts (Unlabeled)}}%
\end{minipage}%
}
\caption{Gemma 2 9B, Layer 11. \textbf{Random Experts}: (a) Expert 1326. (b) Expert 1527. (c) Expert 1537. (d) Expert 210. (e) Expert 229. (f) Expert 289. (g) Expert 464. (h) Expert 508. (i) Expert 51. (j) Expert 571. Each plot shows the 3-D bottleneck activations of a single SMIXAE expert; points are individual token activations colored by distance from the origin.}
\label{fig:random_gemma_2_9b_l11}
\end{figure*}

%% file: paper/random_gemma_2_9b_l20.tex

\begin{figure*}[t]
\centering
\noindent\makebox[\linewidth][c]{%
\begin{minipage}[t]{\linewidth}%
\vspace{0pt}\centering%
\begin{minipage}[c][4.0cm][c]{0.3267\linewidth}%
\centering%
\begin{tikzpicture}[baseline=(img.base)]
  \node[inner sep=0pt] (img) {\includegraphics[width=\linewidth,height=4.0cm,keepaspectratio]{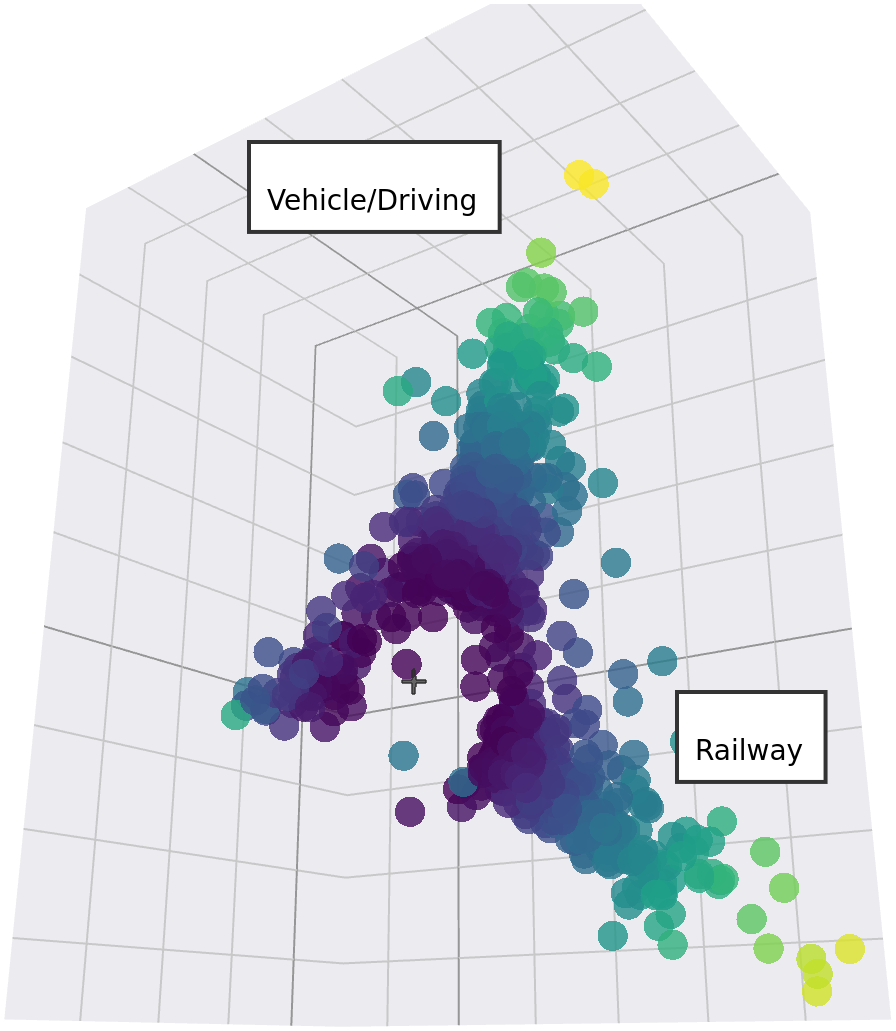}};
  \node[anchor=north west, inner sep=2pt, overlay] at (img.north west) {\small\textbf{(a)}};
\end{tikzpicture}%
\end{minipage}%
\begin{minipage}[c][4.0cm][c]{0.3267\linewidth}%
\centering%
\begin{tikzpicture}[baseline=(img.base)]
  \node[inner sep=0pt] (img) {\includegraphics[width=\linewidth,height=4.0cm,keepaspectratio]{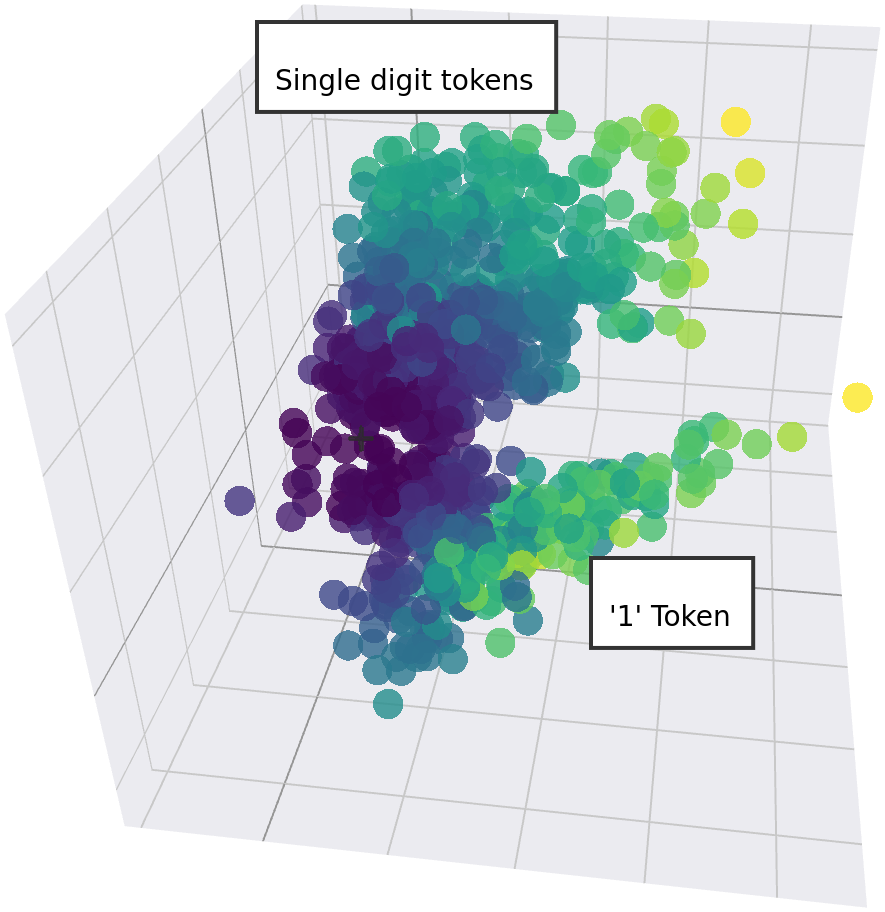}};
  \node[anchor=north west, inner sep=2pt, overlay] at (img.north west) {\small\textbf{(b)}};
\end{tikzpicture}%
\end{minipage}%
\begin{minipage}[c][4.0cm][c]{0.3267\linewidth}%
\centering%
\begin{tikzpicture}[baseline=(img.base)]
  \node[inner sep=0pt] (img) {\includegraphics[width=\linewidth,height=4.0cm,keepaspectratio]{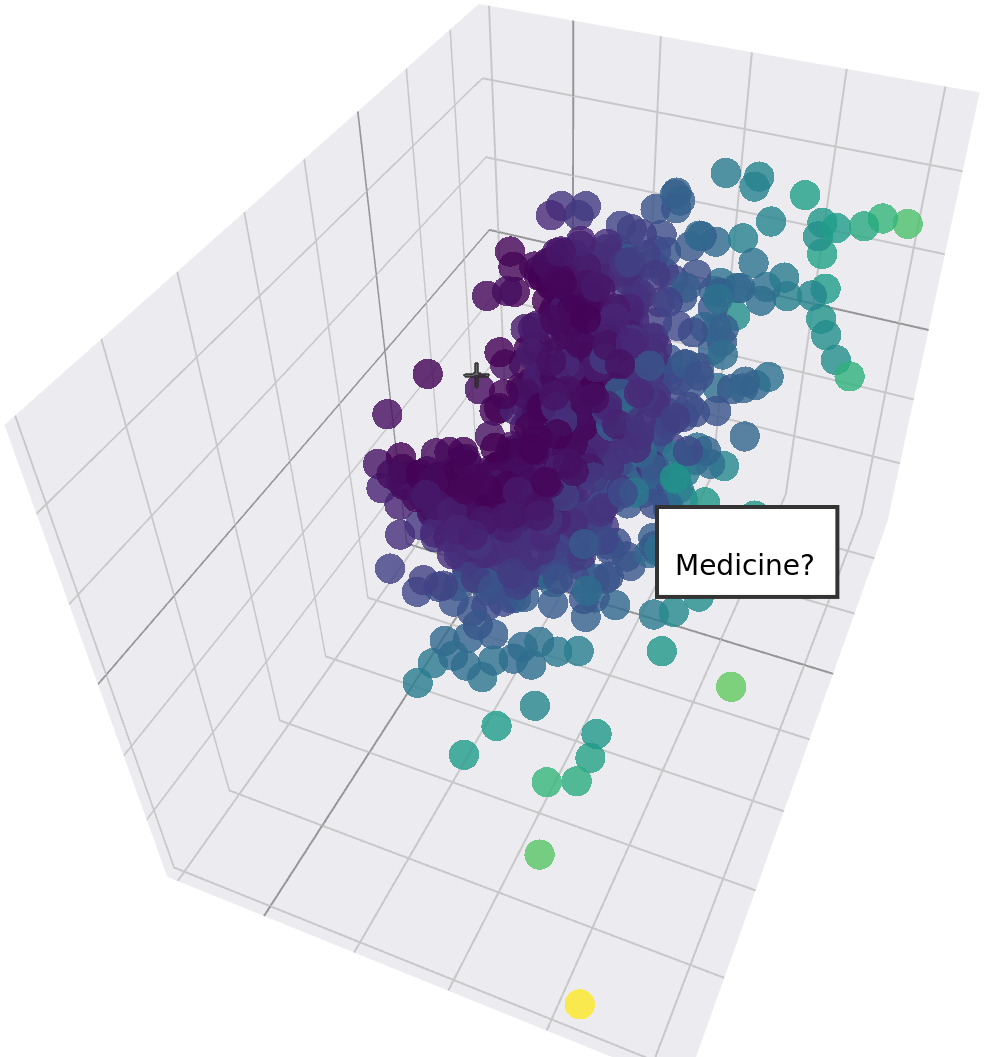}};
  \node[anchor=north west, inner sep=2pt, overlay] at (img.north west) {\small\textbf{(c)}};
\end{tikzpicture}%
\end{minipage}%
\par\vspace{3pt}%
\begin{minipage}[c][4.0cm][c]{0.3267\linewidth}%
\centering%
\begin{tikzpicture}[baseline=(img.base)]
  \node[inner sep=0pt] (img) {\includegraphics[width=\linewidth,height=4.0cm,keepaspectratio]{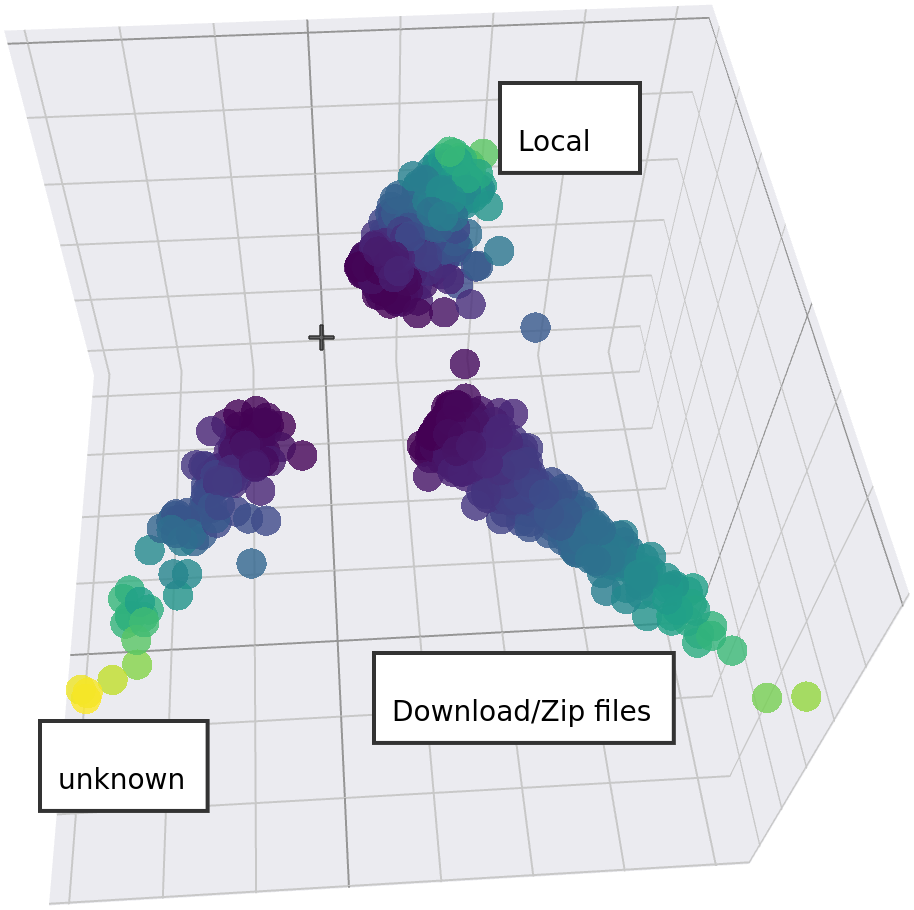}};
  \node[anchor=north west, inner sep=2pt, overlay] at (img.north west) {\small\textbf{(d)}};
\end{tikzpicture}%
\end{minipage}%
\begin{minipage}[c][4.0cm][c]{0.3267\linewidth}%
\centering%
\begin{tikzpicture}[baseline=(img.base)]
  \node[inner sep=0pt] (img) {\includegraphics[width=\linewidth,height=4.0cm,keepaspectratio]{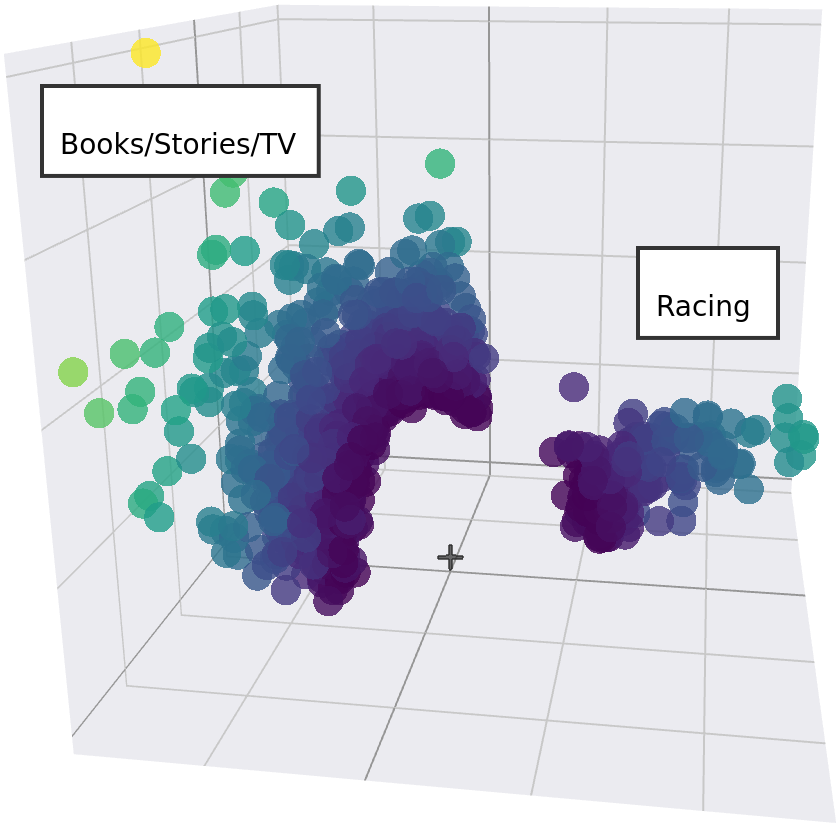}};
  \node[anchor=north west, inner sep=2pt, overlay] at (img.north west) {\small\textbf{(e)}};
\end{tikzpicture}%
\end{minipage}%
\begin{minipage}[c][4.0cm][c]{0.3267\linewidth}%
\centering%
\begin{tikzpicture}[baseline=(img.base)]
  \node[inner sep=0pt] (img) {\includegraphics[width=\linewidth,height=4.0cm,keepaspectratio]{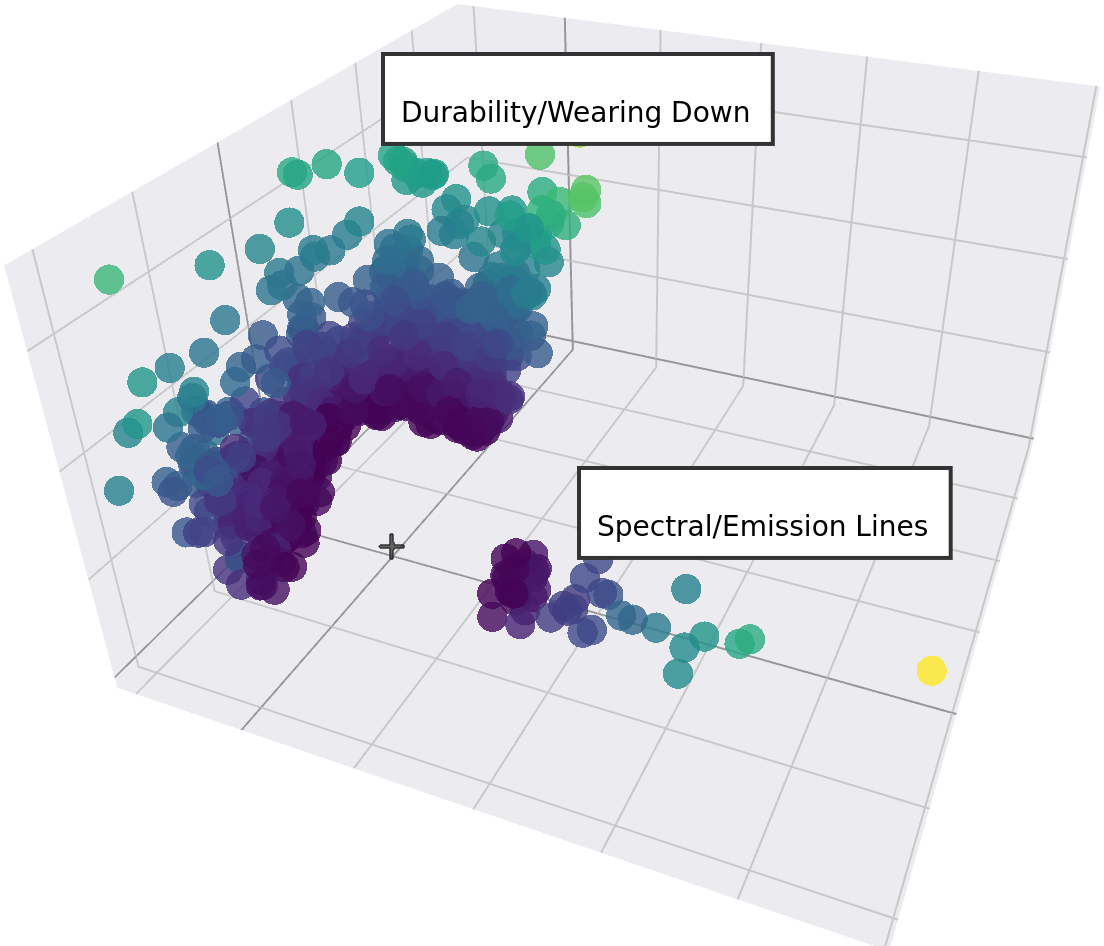}};
  \node[anchor=north west, inner sep=2pt, overlay] at (img.north west) {\small\textbf{(f)}};
\end{tikzpicture}%
\end{minipage}%
\par\vspace{3pt}%
\begin{minipage}[c][4.0cm][c]{0.3267\linewidth}%
\centering%
\begin{tikzpicture}[baseline=(img.base)]
  \node[inner sep=0pt] (img) {\includegraphics[width=\linewidth,height=4.0cm,keepaspectratio]{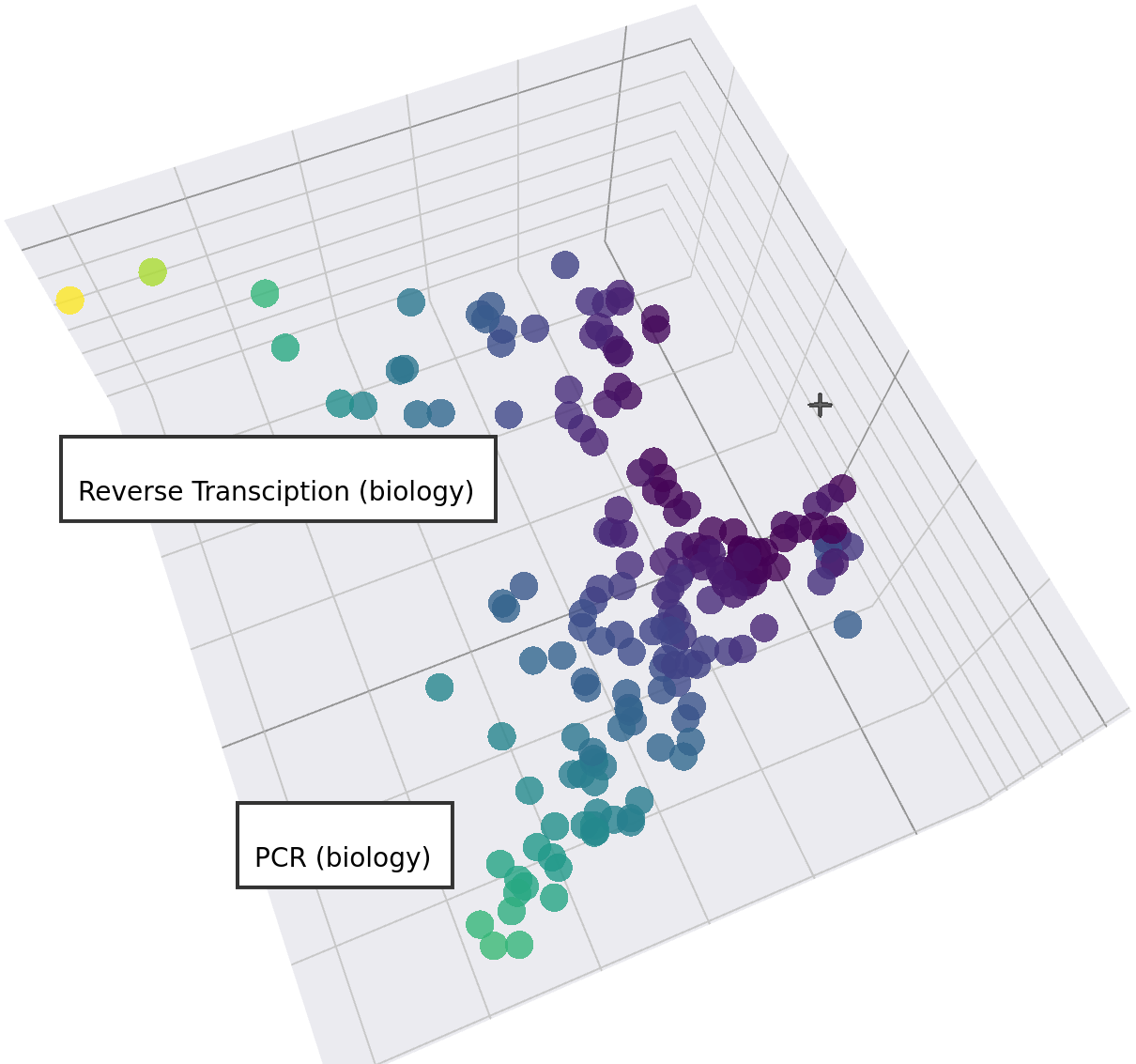}};
  \node[anchor=north west, inner sep=2pt, overlay] at (img.north west) {\small\textbf{(g)}};
\end{tikzpicture}%
\end{minipage}%
\begin{minipage}[c][4.0cm][c]{0.3267\linewidth}%
\centering%
\begin{tikzpicture}[baseline=(img.base)]
  \node[inner sep=0pt] (img) {\includegraphics[width=\linewidth,height=4.0cm,keepaspectratio]{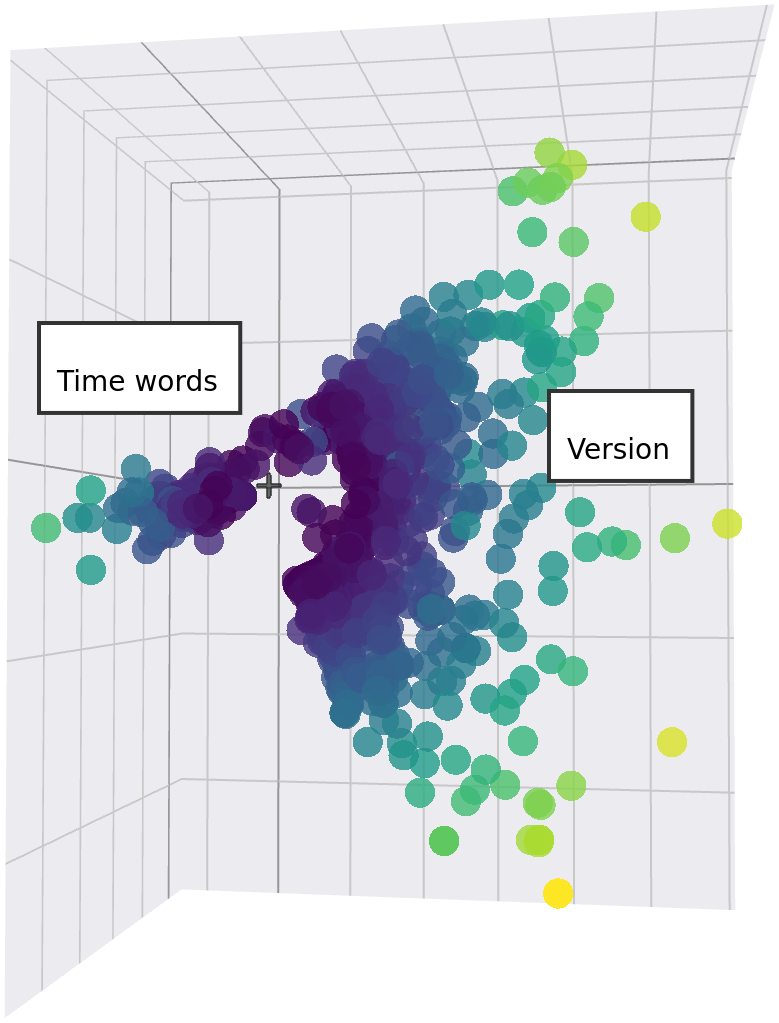}};
  \node[anchor=north west, inner sep=2pt, overlay] at (img.north west) {\small\textbf{(h)}};
\end{tikzpicture}%
\end{minipage}%
\begin{minipage}[c][4.0cm][c]{0.3267\linewidth}%
\centering%
\begin{tikzpicture}[baseline=(img.base)]
  \node[inner sep=0pt] (img) {\includegraphics[width=\linewidth,height=4.0cm,keepaspectratio]{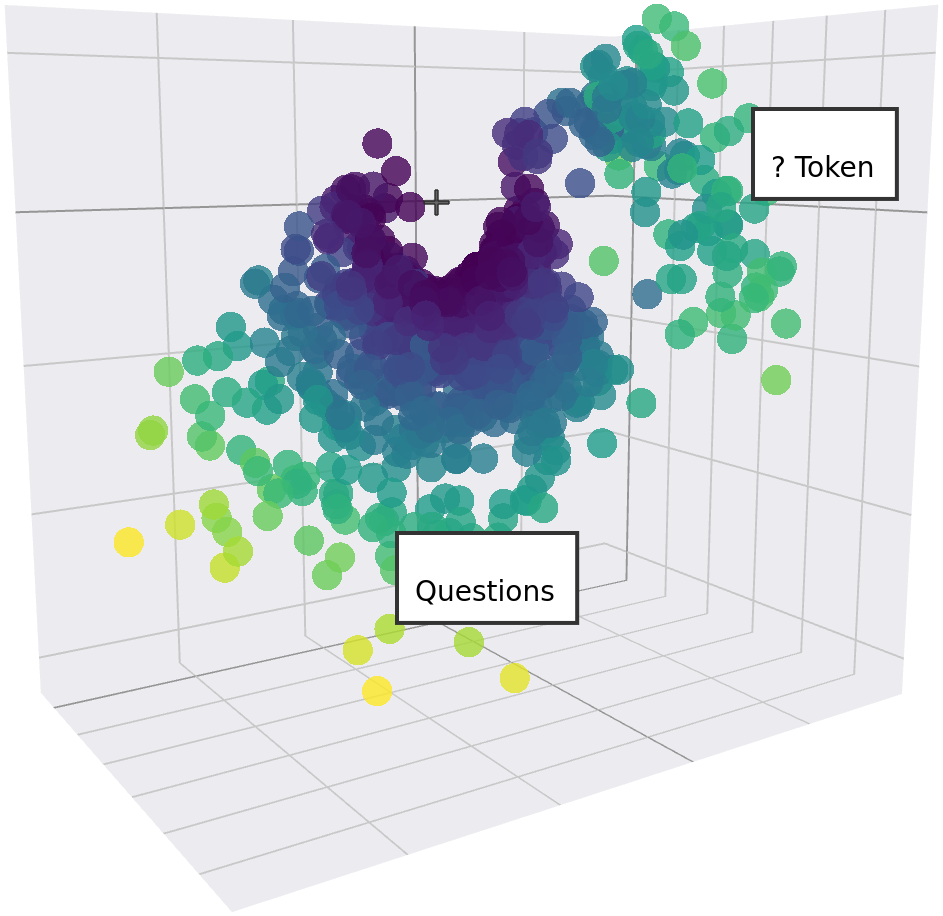}};
  \node[anchor=north west, inner sep=2pt, overlay] at (img.north west) {\small\textbf{(i)}};
\end{tikzpicture}%
\end{minipage}%
\par\vspace{3pt}%
\begin{minipage}[c][4.0cm][c]{0.3267\linewidth}%
\centering%
\begin{tikzpicture}[baseline=(img.base)]
  \node[inner sep=0pt] (img) {\includegraphics[width=\linewidth,height=4.0cm,keepaspectratio]{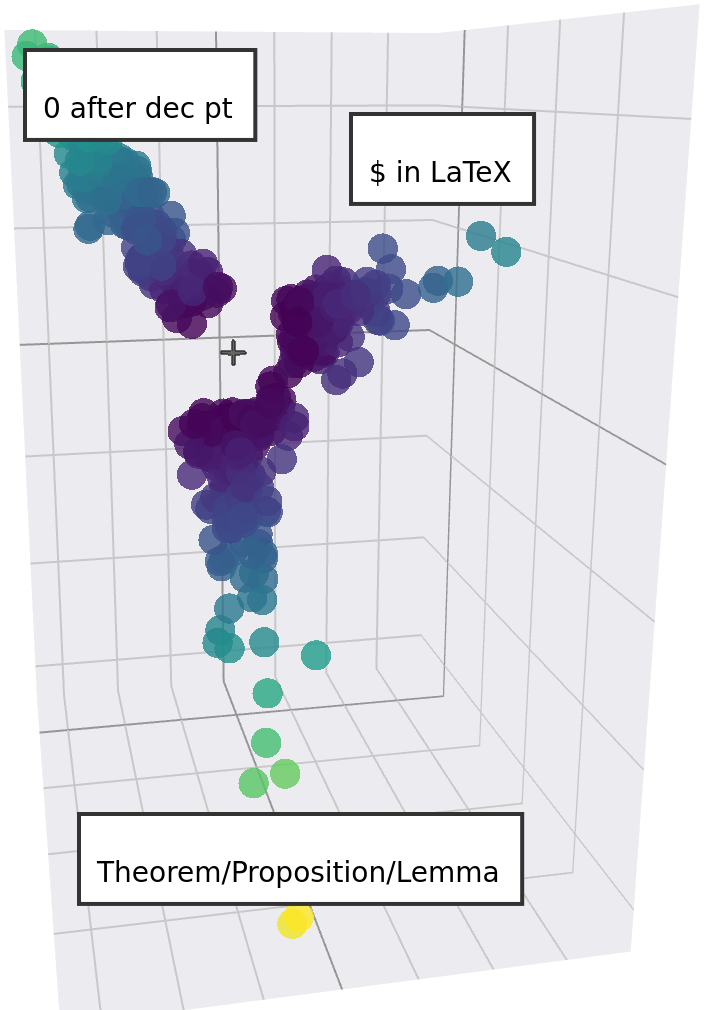}};
  \node[anchor=north west, inner sep=2pt, overlay] at (img.north west) {\small\textbf{(j)}};
\end{tikzpicture}%
\end{minipage}%
\begin{minipage}[c][4.0cm][c]{0.3267\linewidth}%
\centering%
\vspace{0pt}%
\end{minipage}%
\begin{minipage}[c][4.0cm][c]{0.3267\linewidth}%
\centering%
\vspace{0pt}%
\end{minipage}%
\par\vspace{1pt}%
{\footnotesize\textit{Random Experts (Unlabeled)}}%
\end{minipage}%
}
\caption{Gemma 2 9B, Layer 20. \textbf{Random Experts}: (a) Expert 1360. (b) Expert 1567. (c) Expert 1578. (d) Expert 217. (e) Expert 236. (f) Expert 296. (g) Expert 474. (h) Expert 520. (i) Expert 53. (j) Expert 583. Each plot shows the 3-D bottleneck activations of a single SMIXAE expert; points are individual token activations colored by distance from the origin.}
\label{fig:random_gemma_2_9b_l20}
\end{figure*}

%% file: paper/newline_gemma_2_2b_l12.tex

\begin{figure*}[t]
\centering
\noindent\makebox[\linewidth][c]{%
\begin{minipage}[t]{0.6875\linewidth}%
\vspace{0pt}\centering%
\begin{minipage}[c]{0.4255\linewidth}%
\centering%
\begin{tikzpicture}[baseline=(img.base)]
  \node[inner sep=0pt] (img) {\includegraphics[width=\linewidth,height=4.0cm,keepaspectratio]{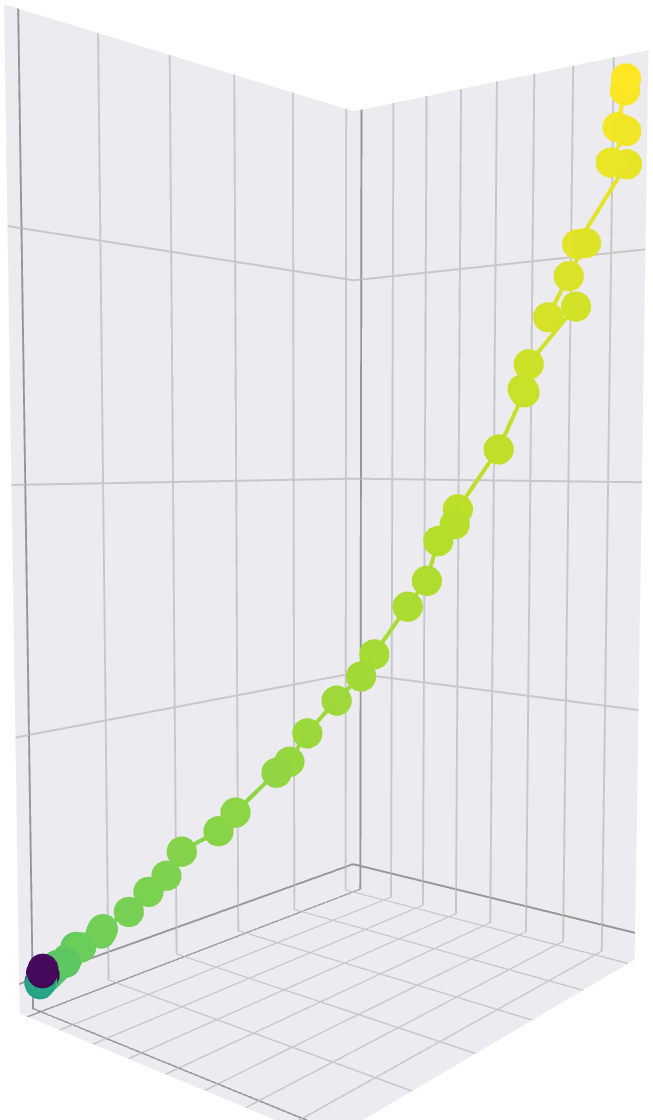}};
  \node[anchor=north west, inner sep=2pt, overlay] at (img.north west) {\small\textbf{(a)}};
\end{tikzpicture}%
\end{minipage}%
\begin{minipage}[c]{0.4255\linewidth}%
\centering%
\begin{tikzpicture}[baseline=(img.base)]
  \node[inner sep=0pt] (img) {\includegraphics[width=\linewidth,height=4.0cm,keepaspectratio]{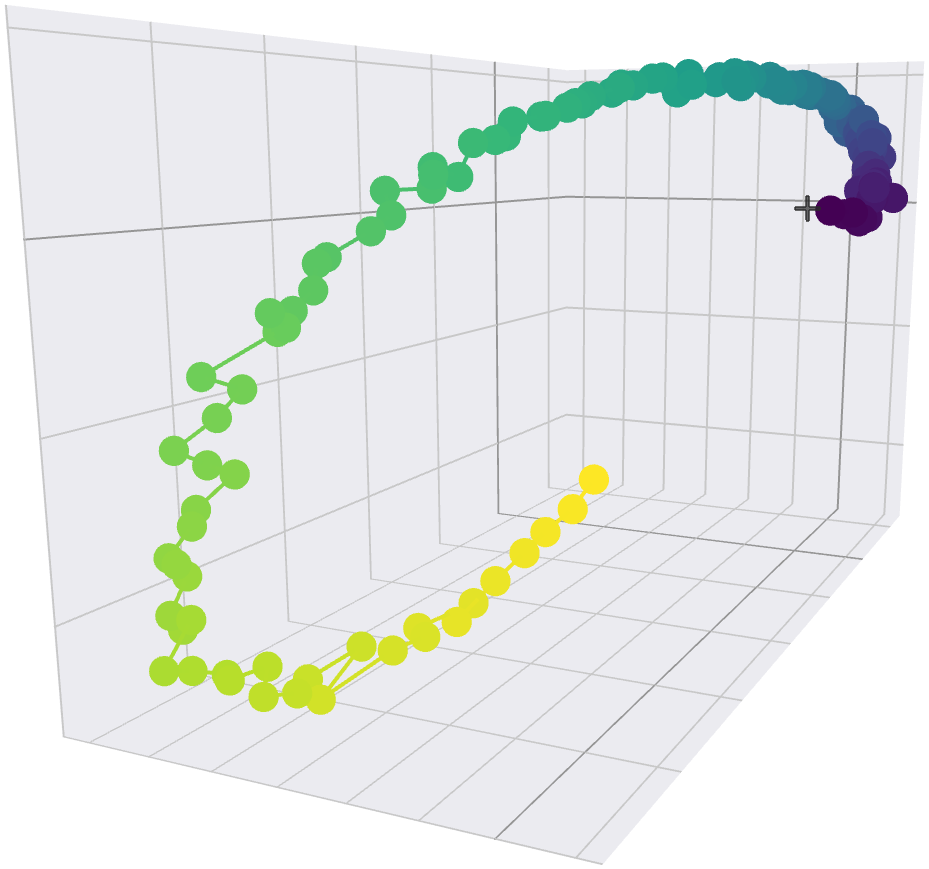}};
  \node[anchor=north west, inner sep=2pt, overlay] at (img.north west) {\small\textbf{(b)}};
\end{tikzpicture}%
\end{minipage}%
\begin{minipage}[c][4.0cm][c]{0.1489\linewidth}%
\centering%
\includegraphics[width=\linewidth,height=4.0cm,keepaspectratio]{paper/legends/legend_pile-uncopyrighted__150.png}%
\end{minipage}%
\par\vspace{1pt}%
{\footnotesize\textit{Newline Position}}%
\end{minipage}%
}
\caption{Newline Position (150 chars) --- Gemma 2 2B, Layer 12. \textbf{Newline Position}: (a) Expert 1895, rank 2, Periodic Gain (Score\,=\,0.406). (b) Expert 647, rank 1, Periodic Gain (Score\,=\,0.225). Points represent per-class mean activations in the bottleneck space, colored by distance since the last newline.}
\label{fig:newline_gemma_2_2b_l12}
\end{figure*}

%% file: paper/newline_gemma_2_9b_l20.tex

\begin{figure*}[t]
\centering
\noindent\makebox[\linewidth][c]{%
\begin{minipage}[t]{0.6875\linewidth}%
\vspace{0pt}\centering%
\begin{minipage}[c][3.3cm][c]{0.4255\linewidth}%
\centering%
\begin{tikzpicture}[baseline=(img.base)]
  \node[inner sep=0pt] (img) {\includegraphics[width=\linewidth,height=3.3cm,keepaspectratio]{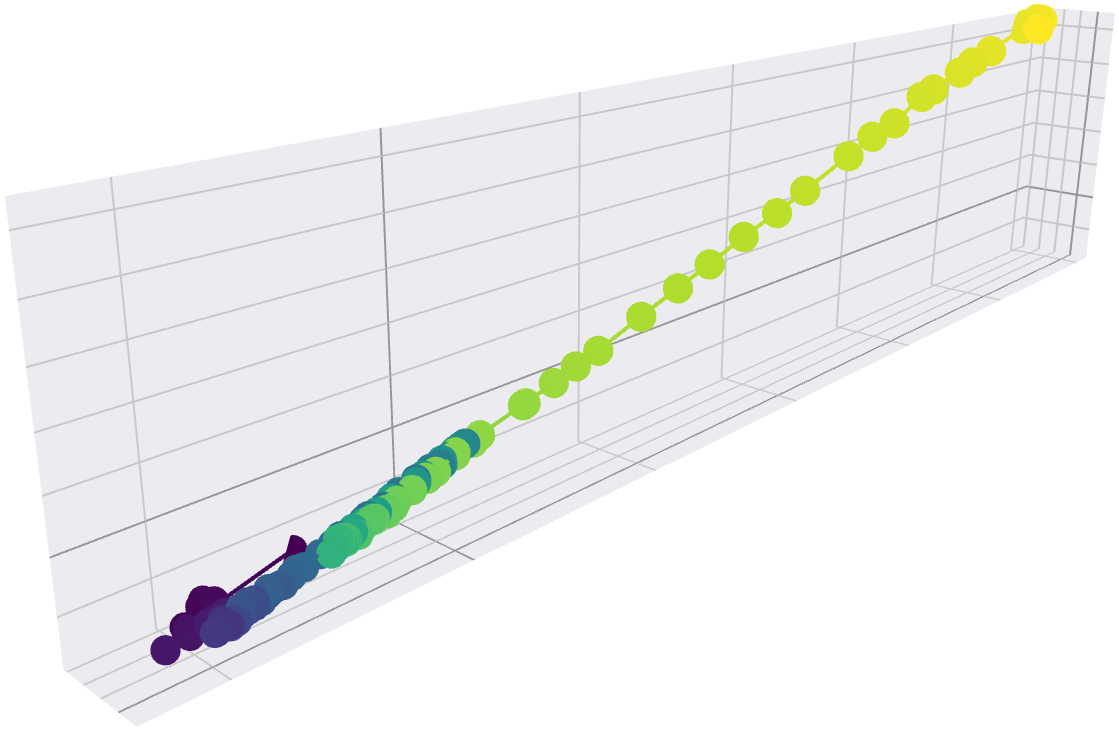}};
  \node[anchor=north west, inner sep=2pt, overlay] at (img.north west) {\small\textbf{(a)}};
\end{tikzpicture}%
\end{minipage}%
\begin{minipage}[c][3.3cm][c]{0.4255\linewidth}%
\centering%
\begin{tikzpicture}[baseline=(img.base)]
  \node[inner sep=0pt] (img) {\includegraphics[width=\linewidth,height=3.3cm,keepaspectratio]{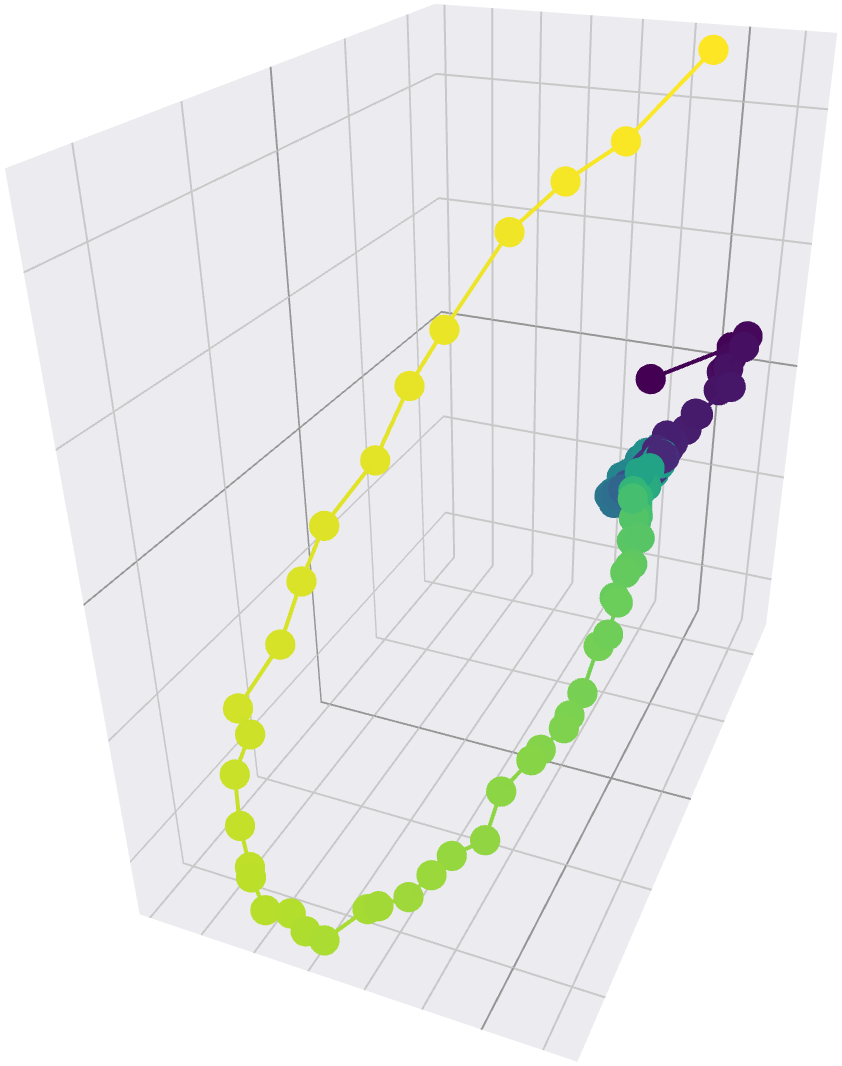}};
  \node[anchor=north west, inner sep=2pt, overlay] at (img.north west) {\small\textbf{(b)}};
\end{tikzpicture}%
\end{minipage}%
\begin{minipage}[c][3.3cm][c]{0.1489\linewidth}%
\centering%
\includegraphics[width=\linewidth,height=3.3cm,keepaspectratio]{paper/legends/legend_pile-uncopyrighted__150.png}%
\end{minipage}%
\par\vspace{1pt}%
{\footnotesize\textit{Newline Position}}%
\end{minipage}%
}
\caption{Newline Position (150 chars) --- Gemma 2 9B, Layer 20. \textbf{Newline Position}: (a) Expert 1749, rank 2, Periodic Gain (Score\,=\,0.320). (b) Expert 1520, rank 1, Periodic Gain (Score\,=\,0.280). Points represent per-class mean activations in the bottleneck space, colored by distance since the last newline.}
\label{fig:newline_gemma_2_9b_l20}
\end{figure*}